%% file: main.tex
\documentclass[10pt,twocolumn,letterpaper]{article}

\usepackage[pagenumbers]{cvpr} 

\input{preamble}
\definecolor{cvprblue}{rgb}{0.21,0.49,0.74}
\usepackage[pagebackref,breaklinks,colorlinks,allcolors=cvprblue]{hyperref}

\def\confName{CVPR}
\def\confYear{2027}

\title{DiRGeo: Learning Directional Geometry of Visual Representations}

\author{
Muhammad Ahtesham, \quad
Xin Zhong \\
Department of Computer Science \\
University of Nebraska Omaha \\
{\tt\small mahtesham@unomaha.edu \quad xzhong@unomaha.edu}
}
\vspace{-1.0em}

\usepackage{microtype}
\usepackage{etoc}
\usepackage{wrapfig}

\begin{document}
\maketitle

\etocdepthtag.toc{main} 

\input{sec/0_abstract}    
\input{sec/1_intro}
\input{sec/2_related}

\input{sec/3_Method}
\input{sec/4_experiment}

\newpage
\section*{LLM Usage Statement}
We used ChatGPT to assist with manuscript restructuring, drafting and editing, and discussions of theoretical formulations and empirical consistency. We also used Claude Code to assist with code development and debugging for experiments. The authors reviewed and verified the resulting content and experimental implementations and take full responsibility for the work. 

{
    \small
    \bibliographystyle{ieeenat_fullname}
    \bibliography{main}
}

\input{sec/X_suppl}

\end{document}

%% file: preamble.tex
\usepackage{float}

%% file: sec/0_abstract.tex
\begin{abstract}
Conventional geometric measures of visual representations are symmetric, assigning equal costs to forward and reverse motion along the same path. We propose DiRGeo, a framework for learning directional geometry over frozen visual representations. Grounded in Randers geometry, DiRGeo learns a state-dependent directional field from ordered image trajectories through self-supervised contrast, without semantic labels. Sphere-tangent correction and attractor-relative decomposition distinguish learned directional structure from normalization artifacts and common convergence. Experiments across 17 visual processes, three encoder families, and multiple representation depths demonstrate consistent directional discrimination and whole-path asymmetry beyond simple attractor-based explanations. The learned fields become increasingly process-specific in deeper representations, while natural-video experiments reveal event-dependent directional preferences. As functional validation, DiRGeo guides restoration selection. These results establish directional geometry as a useful complement to symmetric representation analysis. 
\end{abstract}

%% file: sec/1_intro.tex
\vspace{-2.0em}
\section{Introduction}
\label{sec:introduction}
\vspace{-0.5em}

\begin{figure}[t]
\vspace{-0.5em}
\centering
\includegraphics[width=1.0\columnwidth]{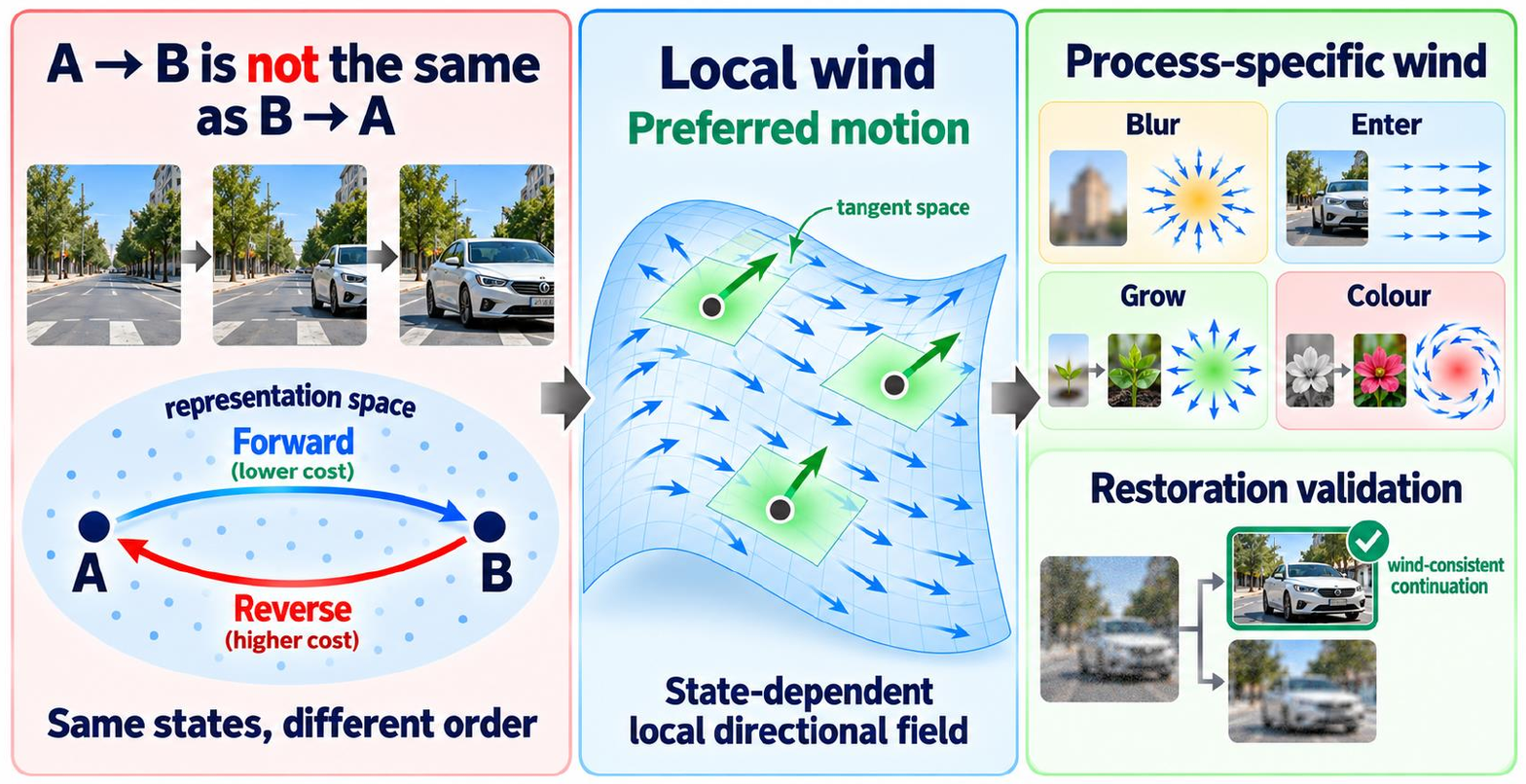}
\vspace{-7.5em}
\caption{\textbf{Directional geometry of visual representations.} 
Left: Forward and reverse visual trajectories can have different costs despite sharing the same states. 
Middle: A state-dependent wind defines local directional preferences. 
Right: The learned ruler distinguishes opposite motions and guides visual restoration.
}
\label{fig:concept}
\vspace{-2.25em}
\end{figure}

Visual representations encode both the content of an image and its response to changes in the scene. When an object gradually appears, an image becomes increasingly occluded, or illumination changes, a visual encoder maps the evolving observations into a trajectory through representation space. Understanding these trajectories requires examining how representations move, beyond where individual images are located. In this work (see Fig.~\ref{fig:concept}), we study the directional geometry of visual representation trajectories: a local geometric structure that assigns different costs to opposite orientations of motion and reveals how these preferences accumulate along an ordered visual process.


Conventional geometric measures characterize representation differences through distances, angles, and local quadratic forms. While capturing displacement magnitude and anisotropy, a Riemannian ruler~\cite{arvanitidis2018latent,shao2018riemannian} assigns equal costs to $v$ and $-v$, and identical lengths to forward and reversed paths. Yet progressive object insertion and its reversal traverse the same states with potentially different directional preferences, while distinct visual processes may connect identical endpoints through different paths. Characterizing such changes therefore requires accounting for both representation state and motion orientation.
However, temporal ordering alone does not establish meaningful directional structure. Predictability may arise from convergence toward a shared endpoint or normalization-induced radial components that spuriously favor inward motion. Moreover, local directional preferences may cancel over a complete trajectory. These challenges motivate our central question: can ordered visual processes reveal reproducible directional geometry beyond normalization artifacts and common-attractor explanations?

Prior work provides complementary foundations for this question. Riemannian analyses characterize the geometry of learned latent spaces~\cite{arvanitidis2018latent}, while studies of perceptual straightening examine the shape of visual trajectories~\cite{henaff2019straightening}. Local invariant-subspace analysis further relates representation geometry to semantic stability~\cite{dasgupta2026inv}. Asymmetric embedding methods preserve directed relations through quasimetrics and Finsler geometry~\cite{wang2022quasimetrics,dages2025finslermds}, and learned directional metrics support trajectory inference and latent interpolation~\cite{zweig2026lineage}. Temporal-order and arrow-of-time learning also demonstrate that sequence orientation provides useful visual supervision~\cite{misra2016shuffle,wei2018arrow}. Building on these foundations, we investigate how a local directional field characterizes visual processes within frozen pretrained representations, and how its measured asymmetry relates to trajectory order, representation normalization, and common convergence.

We learn a local directional ruler over frozen visual representations using a Randers formulation that combines a symmetric base with a state-dependent directional field. Ordered image sequences provide supervision by contrasting observed motion with alternative directions, without semantic class labels. The learned ruler measures both local directional preferences and accumulated forward--reverse asymmetry. Sphere tangent correction and attractor-relative decomposition further distinguish the recovered structure from normalization artifacts and common convergence.

Our contributions are threefold.
\textbf{(1)} We formulate process-specific directional geometry for visual representation trajectories using a non-reversible Randers ruler. The formulation connects state-dependent orientation preferences to accumulated path asymmetry, providing a geometric account of ordered visual change that complements symmetric distances and local subspace descriptions.
\textbf{(2)} We develop a learning framework over frozen visual encoders, together with geometric diagnostics for interpreting the recovered field. Directional contrast learns from sequence order, while sphere tangent correction and attractor-relative decomposition examine whether the measured structure extends beyond normalization-induced radial motion and a shared convergence direction. 
\textbf{(3)} We empirically characterize directional geometry across 17 visual processes, three encoder families, and multiple representation depths. The learned fields exhibit reproducible directional preferences, local structure beyond common convergence, and increasingly distinct process-specific organization at deeper layers. Natural-video experiments reveal event-dependent asymmetry, while restoration selection demonstrates functional utility beyond a fitted attractor for colour correction.


%% file: sec/2_related.tex
\vspace{-1.0em}
\section{Related Work}
\label{sec:related_work}
\vspace{-0.5em}

\noindent\textbf{Geometry of representations and visual trajectories.}
Prior work characterizes representation geometry through Riemannian metrics for latent-space distances and interpolation~\cite{arvanitidis2018latent}, perceptual straightening of natural video trajectories~\cite{henaff2019straightening}, and invariant subspaces that distinguish semantic-preserving from semantic-changing variation~\cite{dasgupta2026inv}. These approaches describe local distances, trajectory curvature, or invariant directions, but do not explicitly characterize orientation-dependent costs along visual trajectories. In contrast, we learn a non-reversible local ruler that distinguishes forward from reverse motion and measures how directional preferences accumulate along a path.

\noindent\textbf{Asymmetric distances and directed embeddings.}
Quasimetric embeddings model asymmetric pairwise relations with structural constraints, supporting applications to directed graphs and goal-reaching problems~\cite{wang2022quasimetrics,wang2022interval}. Finsler multidimensional scaling represents asymmetric dissimilarities in Randers space~\cite{dages2025finslermds}, while Finsler-based spectral embeddings characterize directional transport in directed data~\cite{debaussart2026directed}. Unlike these approaches, DiRGeo learns a state-dependent local ruler over frozen visual representations directly from ordered trajectories, without requiring pairwise asymmetric distance supervision. This enables path-level analysis and decomposition of the learned directional field.

\noindent\textbf{Finsler geometry and learned directional fields.}
Randers geometry defines non-reversible costs and provides an interpretation of navigation under drift~\cite{bao2004zermelo}. Recent work employs Finsler metrics for lineage-guided trajectory inference~\cite{zweig2026lineage} and differentiable distance solvers for learning metric and drift fields from arrival times~\cite{gahtan2026eikonal}. In contrast, DiRGeo learns directional fields from ordered visual trajectories in frozen representations, using self-supervised directional contrast. We characterize process dependence, whole-path asymmetry, and residual structure beyond a common attractor, while controlling for feature-normalization artifacts. 

\noindent\textbf{Temporal order and arrow-of-time learning.}
Temporal ordering provides self-supervision for visual representation learning through frame-order verification~\cite{misra2016shuffle} and forward--backward video discrimination~\cite{pickup2014seeing,wei2018arrow}. Unlike these approaches, DiRGeo learns a local directional ruler over frozen representations rather than training representations to recognize temporal order. It measures orientation-dependent costs and their accumulation along trajectories, without assuming a universal arrow of time across visual processes.

%% file: sec/3_Method.tex
\vspace{-1.0em}
\section{Learning Directional Geometry}
\label{sec:method}
\vspace{-0.5em}

\begin{figure*}[ht]
\vspace{-1.0em}
\centering
\includegraphics[width=1.8\columnwidth]{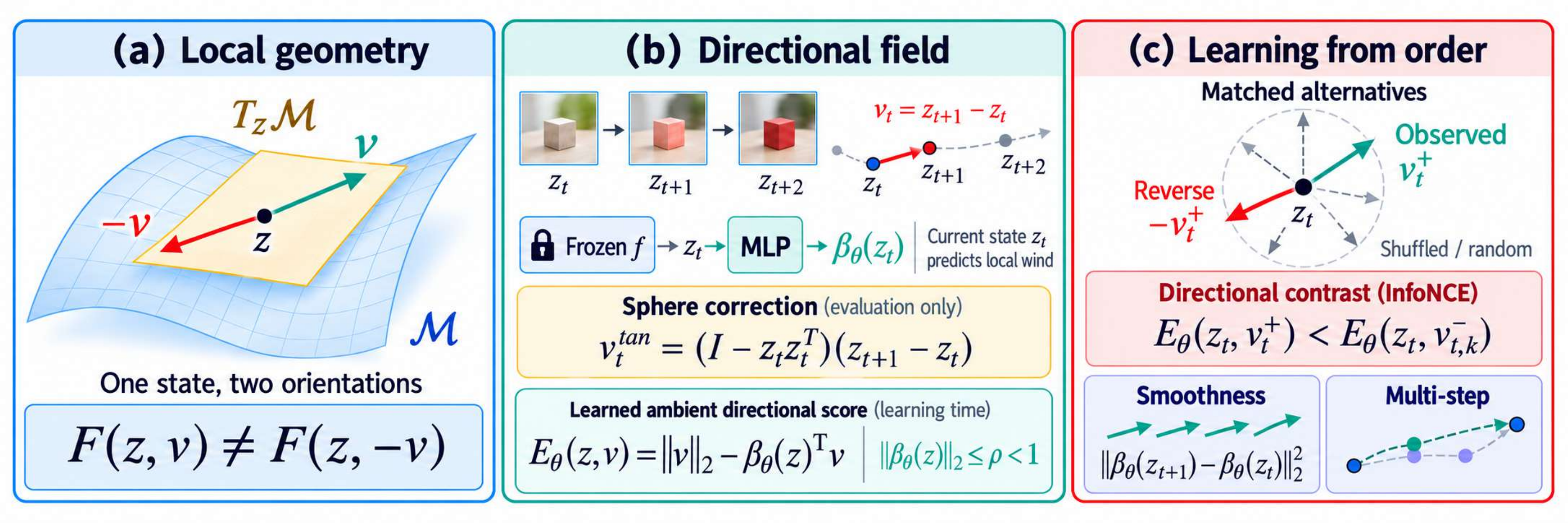}
\vspace{-1.0em}
\caption{\textbf{Learning directional geometry from ordered trajectories.} (a) Opposite motions on a representation manifold can have different costs. (b) A lightweight network learns a bounded Randers wind over frozen features; sphere-tangent correction removes normalization artifacts during evaluation. (c) Directional contrast favors observed motion over reversed, shuffled, and random alternatives, regularized by smoothness and multi-step consistency.
}
\label{fig:method}
\vspace{-2.0em}
\end{figure*}

Figure~\ref{fig:method} provides an overview of DiRGeo, from local directional geometry to field parameterization and self-supervised learning. Section~\ref{sec:formulation} formulates the directional ruler using a non-reversible Randers metric. Section~\ref{sec:parameterization} introduces the bounded state-dependent field and sphere-tangent correction. Section~\ref{sec:learning} describes learning from ordered trajectories, while Section~\ref{sec:characterization} characterizes whole-path asymmetry and decomposition relative to a common attractor.

\vspace{-0.5em} 
\subsection{Directional Geometry of Representation Trajectories}
\label{sec:formulation}
\vspace{-0.5em} 

Let $f:\mathcal{X}\rightarrow\mathbb{R}^{d}$ denote a trained visual encoder and $z=f(x)$ the representation of an image $x$. Following the manifold hypothesis~\cite{fefferman2016testing}, we model the relevant representation states as lying on a smooth manifold $\mathcal{M}\subseteq\mathbb{R}^{d}$. An ordered visual process, such as progressive occlusion or object insertion, induces a representation trajectory $\gamma:[0,1]\rightarrow\mathcal{M}$. 
Wherever $\gamma$ is differentiable, at
\(
    z=\gamma(t),
\)
its velocity
\(
    v=\dot{\gamma}(t)\in T_z\mathcal{M}
\)
describes the oriented local motion through representation space. The tangent space $T_z\mathcal{M}$ provides a first-order description of the surrounding manifold, while $v$ specifies how a particular process moves through that neighborhood. 

\noindent \textbf{From local variation to oriented motion.}
A Riemannian metric $g_z$ measures local motion through
\(
    \alpha(z,v)=\sqrt{g_z(v,v)},
    \space
    \alpha(z,v)=\alpha(z,-v).
\)
Such a metric can be anisotropic: distinct tangent axes may have different costs. Nevertheless, it assigns the same cost to opposite orientations along any given axis. Thus, anisotropy alone cannot express a preference between advancing and reversing a local visual change. For example, progressively occluding an object and revealing it in the reverse order traverse the same representation path with opposite orientations, yet an ordered process distribution may preferentially support one orientation.  
We hypothesize that, for a specified visual process, this preference has reproducible state-dependent structure. The geometric object of interest is therefore a local cost $F(z,v)$ that depends on both the representation state and its oriented motion, allowing $F(z,v)\neq F(z,-v)$. This asymmetry is defined relative to the process and its ordering; it is not implied by the manifold assumption or by temporal order alone. We suppress the process index below for clarity.

\noindent \textbf{A Randers formulation.}
We represent this directional geometry using a Randers metric~\cite{bao2004zermelo},
\begingroup
\setlength{\abovedisplayskip}{3pt}
\setlength{\belowdisplayskip}{3pt}
\setlength{\abovedisplayshortskip}{3pt}
\setlength{\belowdisplayshortskip}{3pt}
\begin{equation}
    F(z,v)=\alpha(z,v)-\beta_z(v),
    \qquad
    \alpha(z,v)=\sqrt{g_z(v,v)},
    \label{eq:randers_formulation}
\end{equation}
\endgroup
where $\beta$ is a smooth differential one-form: at each state $z$, $\beta_z$ is a linear functional on $T_z\mathcal{M}$. In coordinates, $\beta_z(v)=\beta(z)^\top v$. The Riemannian term supplies the symmetric base geometry, while the one-form introduces an orientation-dependent correction. We refer to this correction as a directional wind: under our sign convention, $\beta_z(v)>0$ reduces the cost of motion along $v$. 
To obtain a valid Randers metric, we require
\(
    \|\beta_z\|_{g_z^*}
    :=
    \sup_{v\neq 0}
    \frac{|\beta_z(v)|}{\alpha(z,v)}
    \leq \rho < 1.
    \label{eq:randers_bound}
\)
Together with smoothness and positive definiteness of $g$, this condition ensures positivity and strong convexity away from the zero tangent vector. In particular,
$(1-\rho)\alpha(z,v)\leq F(z,v)\leq
(1+\rho)\alpha(z,v)$.
The resulting cost is positively homogeneous,
$F(z,cv)=cF(z,v)$ for $c>0$, while its response to orientation reversal satisfies 
\(
    F(z,-v)-F(z,v)=2\beta_z(v).
\)
Thus, the directional component is separated explicitly from the symmetric magnitude of local motion. Setting $\beta=0$ recovers the Riemannian geometry.

\noindent \textbf{From local orientation to trajectory asymmetry.}
For a piecewise smooth trajectory $\gamma$, define its directional length 
\begingroup
\setlength{\abovedisplayskip}{3pt}
\setlength{\belowdisplayskip}{3pt}
\setlength{\abovedisplayshortskip}{3pt}
\setlength{\belowdisplayshortskip}{3pt}
\begin{equation}
    L_F(\gamma)
    =
    \int_0^1 F\bigl(\gamma(t),\dot{\gamma}(t)\bigr)\,dt.
    \label{eq:directional_length}
\end{equation}
\endgroup
Positive homogeneity makes this length invariant to orientation-preserving reparameterization. Let
$\bar{\gamma}(t)=\gamma(1-t)$ denote the same path traversed in reverse. By the symmetry of $\alpha$ and linearity of $\beta$,
\(
    L_F(\bar{\gamma})-L_F(\gamma)
    =
    2\int_0^1
    \beta_{\gamma(t)}\bigl(\dot{\gamma}(t)\bigr)\,dt.
\)
The forward--reverse difference therefore measures accumulated alignment with the directional field. Local preferences may reinforce one another or cancel along a trajectory, so local asymmetry need not produce a nonzero whole-path difference. 

This formulation treats directional geometry as a structure associated jointly with the encoder and the observed visual process. It does not require observed trajectories to be geodesics, nor does a nonzero directional cost difference alone establish semantic change or physical irreversibility. The central question is whether ordered visual trajectories support a local directional field that generalizes to unseen instances and captures structure beyond simple global drift or convergence. The following sections develop a learnable realization of this geometry.


\vspace{-0.5em} 
\subsection{Parameterizing the Local Directional Field}
\label{sec:parameterization}
\vspace{-0.5em} 

We instantiate the formulation in Section~\ref{sec:formulation} using a fixed Euclidean base metric and a learned state-dependent directional field. The visual encoder remains frozen throughout: learning changes how representation motion is measured, while preserving the underlying features.

\noindent \textbf{Frozen representation trajectories.}
Given an ordered image sequence $(x_0,\ldots,x_T)$, we extract unit-normalized representations and their consecutive displacements,
\(
    z_t=
    f(x_t)
    \space / \space
    \|f(x_t)\|_2
    ,
    \space
    \Delta z_t=z_{t+1}-z_t.
\)
Here $f$ denotes the selected encoder representation before normalization. Each $\Delta z_t$ is a finite chord between two states on $\mathbb{S}^{d-1}$, rather than an exact tangent vector. This distinction matters because normalization introduces a radial component that can mimic directional structure. Specifically,
\begingroup
\setlength{\abovedisplayskip}{3pt}
\setlength{\belowdisplayskip}{3pt}
\setlength{\abovedisplayshortskip}{3pt}
\setlength{\belowdisplayshortskip}{3pt}
\begin{equation}
    z_t^\top\Delta z_t
    =
    z_t^\top z_{t+1}-1
    =
    -\frac{1}{2}\|\Delta z_t\|_2^2.
    \label{eq:radial_chord}
\end{equation}
\endgroup
Every nonzero chord therefore has an inward component at its starting point. An inward-pointing field proportional to $-z_t$ consequently aligns with every such displacement, even without information about the visual process.

\begin{wrapfigure}{r}{0.5\columnwidth}
    \centering
    \vspace{-0.8em}
    \includegraphics[width=\linewidth]{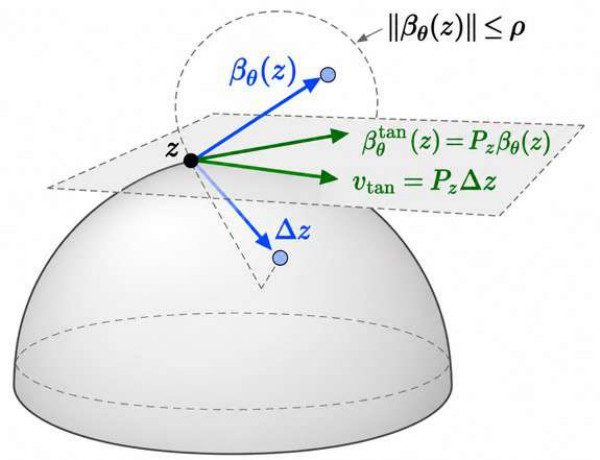}
    \caption{\textbf{Sphere-tangent correction.} Both the observed displacement and the bounded learned wind are projected onto the sphere's tangent space, removing radial components.
    }
    \label{fig:tangent_correction}
    \vspace{-1.0em}
\end{wrapfigure}

\noindent \textbf{Sphere tangent correction.}
To remove this radial contribution, we use the orthogonal projection onto the tangent space of the unit sphere (Fig.~\ref{fig:tangent_correction}). 
For observed motion, the projection is
\(
    P_z=I-zz^\top,
    \space
    v_t^{\mathrm{tan}}
    =
    P_{z_t}\Delta z_t,
    \space
    z_t^\top v_t^{\mathrm{tan}}=0.
\)

For sufficiently close trajectory states, $v_t^{\mathrm{tan}}$ provides a first-order approximation to motion along the sphere. It retains the local orientation and step magnitude to first order, while eliminating the inward component in Eq.~\eqref{eq:radial_chord}.
This correction uses the known geometry induced by feature normalization. It does not estimate the tangent space of the unknown representation manifold. If the normalized representation manifold satisfies $\mathcal{M}\subseteq\mathbb{S}^{d-1}$, then $T_z\mathcal{M}\subseteq T_z\mathbb{S}^{d-1}$, but the two spaces need not coincide. Sphere projection therefore removes a specific geometric artifact without asserting that every remaining direction corresponds to realizable visual variation.

\noindent \textbf{Bounded state-dependent field.}
A lightweight network $b_\theta:\mathbb{R}^{d}\rightarrow\mathbb{R}^{d}$ predicts a directional field from the current representation. We constrain its magnitude through
\(
    \beta_\theta(z)
    =
    (\rho \cdot b_\theta(z) 
    \space / \space
    \max\{1,\|b_\theta(z)\|_2\}),
    \space
    0<\rho<1.
\)
This construction guarantees $\|\beta_\theta(z)\|_2\leq\rho$. Under the Euclidean base metric, the vector $\beta_\theta(z)$ represents the one-form through
$\beta_{\theta,z}(v)=\beta_\theta(z)^\top v$.
Its effective component on the sphere is
\(
    \beta_\theta^{\mathrm{tan}}(z)
    =
    P_z\beta_\theta(z),
    \space
    \space
    \|\beta_\theta^{\mathrm{tan}}(z)\|_2
    \leq
    \|\beta_\theta(z)\|_2
    \leq\rho.
    \label{eq:tangent_wind}
\)
Projection thus preserves the required norm bound without additional rescaling. Moreover, for any $v\in T_z\mathbb{S}^{d-1}$,
$\beta_\theta(z)^\top v
=
\beta_\theta^{\mathrm{tan}}(z)^\top v$;
the radial component of the predicted field has no effect on tangent motion.

With $g_z(u,v)=u^\top v$, the local directional ruler becomes
\(
    F_\theta(z,v)
    =
    \|v\|_2
    -
    \beta_\theta^{\mathrm{tan}}(z)^\top v,
    \space \space
    v\in T_z\mathbb{S}^{d-1}.
\)
The norm bound ensures
$(1-\rho)\|v\|_2\leq F_\theta(z,v)
\leq(1+\rho)\|v\|_2$.
In our primary geometric evaluation, a discrete step is assigned the cost
$F_\theta(z_t,v_t^{\mathrm{tan}})$.
The same sphere correction is applied to learned fields and geometric baselines, ensuring that their directional comparisons exclude the radial artifact.
We implement $b_\theta$ as a three-layer MLP with hidden width $512$ and use $\rho=0.9$ by default. A separate field is learned for each trajectory family and encoder representation. Within each field, the current state $z$ is the sole network input; neither the trajectory index nor the future state is supplied. The following section describes how ordered trajectories supervise this state-dependent field.


\vspace{-0.5em} 
\subsection{Learning from Ordered Visual Trajectories}
\label{sec:learning}
\vspace{-0.5em} 

Ordered visual trajectories provide supervision for the directional field without requiring semantic class labels. At each state, the observed next step supplies a positive direction, while alternative orientations provide contrasting candidates. The learning objective encourages the field to recognize motion supported by the specified process and to maintain coherent directional preferences across neighboring states and temporal scales.

 \noindent \textbf{Directional contrast with matched negatives.}
For a trajectory state $z_t$, let $v_t^+=\Delta z_t=z_{t+1}-z_t$ denote the observed finite displacement. We train using the ambient extension of the Randers cost,
\begingroup
\setlength{\abovedisplayskip}{3pt}
\setlength{\belowdisplayskip}{3pt}
\setlength{\abovedisplayshortskip}{3pt}
\setlength{\belowdisplayshortskip}{3pt}
\begin{equation}
    E_\theta(z,v)
    =
    \|v\|_2-\beta_\theta(z)^\top v,
    \qquad v\in\mathbb{R}^{d}.
    \label{eq:training_energy}
\end{equation}
\endgroup
This score operates on finite representation displacements during learning. The tangent-space evaluation introduced in Section~\ref{sec:parameterization} subsequently tests whether the learned preference persists after removing the radial contribution induced by normalization.

We contrast $v_t^+$ against three types of alternatives: its reversal, a displacement sampled from another trajectory of the same process family, and a random direction. For a nonzero candidate direction $w_{t,k}$, we construct
\(
    v_{t,k}^-=
    \|v_t^+\|_2
    \frac{w_{t,k}}{\|w_{t,k}\|_2},
    \space
    w_{t,k}\in
    \left\{
        -v_t^+,\,
        z_{s+1}^{(j)}-z_s^{(j)},\,
        r
    \right\},
    \space
    r\sim\mathcal{N}(0,I),
    \label{eq:matched_negatives}
\)
where $j$ indexes a different trajectory. All candidates are evaluated at the same anchor $z_t$ and have the same Euclidean magnitude. The reverse candidate tests orientation along the observed axis, while the shuffled candidate tests whether the field associates a direction with the appropriate local state. These alternatives serve as contrastive samples; they are not assumed to be physically impossible transitions.
%
For $K$ negative candidates, the directional contrastive loss is 
\begin{equation}
\mathcal{L}_{\mathrm{dir}}
=
\mathbb{E}_t
\left[
\log\left(
1+\sum_{k=1}^{K}
\exp\frac{E_t^+-E_{t,k}^-}{\tau}
\right)
\right].
\label{eq:directional_contrast}
\end{equation}
where \(E_t^+=E_\theta(z_t,v_t^+)\), \(E_{t,k}^-=E_\theta(z_t,v_{t,k}^-)\), and $\tau>0$ is the temperature and the expectation averages over sampled training steps. Because candidate magnitudes are matched, their Euclidean cost terms cancel within the softmax. Discrimination therefore depends on alignment with $\beta_\theta(z_t)$ rather than on differences in step length. When the field vanishes, all candidates receive identical scores; consistent directional structure in the training data provides a signal for departing from this symmetric solution.

\noindent \textbf{Local field consistency.}
Directional contrast constrains the field at individual sampled states. To encourage coherent variation along a trajectory, we penalize changes between consecutive field predictions,
\begingroup
\setlength{\abovedisplayskip}{3pt}
\setlength{\belowdisplayskip}{3pt}
\setlength{\abovedisplayshortskip}{3pt}
\setlength{\belowdisplayshortskip}{3pt}
\begin{equation}
    \mathcal{L}_{\mathrm{smooth}}
    =
    \mathbb{E}_{t}
    \left[
        \|\beta_\theta(z_{t+1})-\beta_\theta(z_t)\|_2^2
    \right].
    \label{eq:field_smoothness}
\end{equation}
\endgroup
This term regularizes the ambient vector representation of the field at nearby states. It encourages local consistency without requiring a constant direction across the trajectory or across different images.

\noindent\textbf{Multi-step directional consistency.}
A field fitted only to adjacent frames may capture fluctuations that do not persist beyond a single step. We therefore encourage directional consistency over short horizons. For each valid horizon $h\in\mathcal{H}$, let $v_t^{(h)}=z_{t+h}-z_t$. Using the Randers energy difference $E_\theta(z,v)-E_\theta(z,-v)=-2\beta_\theta(z)^\top v$, we define 
\begin{equation}
    \mathcal{L}_{\mathrm{multi}}
    =
    \mathbb{E}_{t,h}
    \left[
    0.1\|v_t^{(h)}\|_2-2\beta_\theta(z_t)^\top v_t^{(h)}
    \right]_+ ,
    \label{eq:multistep_consistency}
\end{equation}
where $[a]_+=\max(a,0)$ and the expectation averages equally over all valid pairs with $h\in\mathcal{H}$ and $t+h\leq T$; the margin is proportional to the step length.
This objective encourages the local field to align with net forward motion over multiple nearby steps. These short-horizon chords provide additional directional supervision; they are not treated as exact manifold tangents or as integrated trajectory lengths.

\noindent \textbf{Overall objective.}
We optimize the field parameters using
\(
    \mathcal{L}
    =
    \mathcal{L}_{\mathrm{dir}}
    +
    \lambda_{\mathrm{smooth}}\mathcal{L}_{\mathrm{smooth}}
    +
    \lambda_{\mathrm{multi}}\mathcal{L}_{\mathrm{multi}},
\)
while keeping the visual encoder fixed. Directional contrast supplies the primary learning signal, and the two regularizers encourage consistency across states and horizons. Supervision comes from the ordering of the selected visual process, rather than semantic annotations. Accordingly, the learned preference characterizes that process distribution; whether it generalizes to unseen images and survives geometric controls is assessed separately on held-out trajectories.


\vspace{-0.5em}
\subsection{Characterizing the Learned Geometry}
\label{sec:characterization}
\vspace{-0.5em}

How can we interpret the learned directional ruler? We examine three complementary properties: whether local preferences accumulate into whole-path asymmetry, whether the field aligns with a common attractor, and whether predictive directional structure remains after removing this alignment.

\noindent\textbf{Whole-path directional asymmetry.}
For a discrete trajectory $\gamma=(z_0,\ldots,z_T)$, we approximate Eq.~\eqref{eq:directional_length} using sphere-projected displacements. Forward and reverse traversal evaluate the field at different departure states:
\(
v_t^{\rightarrow}=P_{z_t}(z_{t+1}-z_t),
\quad
v_t^{\leftarrow}=P_{z_{t+1}}(z_t-z_{t+1}).
\)
The corresponding directional lengths are
\(
\widehat{L}_F(\gamma)=\sum_{t=0}^{T-1}F_\theta(z_t,v_t^{\rightarrow}),
\quad
\widehat{L}_F(\bar{\gamma})=\sum_{t=0}^{T-1}F_\theta(z_{t+1},v_t^{\leftarrow}),
\)
where $\bar{\gamma}=(z_T,\ldots,z_0)$. Unlike local reversal at a fixed state, whole-path reversal accounts for the state dependence of the field. Since
$\|v_t^{\rightarrow}\|_2=\|v_t^{\leftarrow}\|_2=\sqrt{1-(z_t^\top z_{t+1})^2}$,
the symmetric contributions cancel in the difference between directional lengths. We quantify the resulting asymmetry by
\(
R(\gamma)=
\frac{\widehat{L}_F(\bar{\gamma})-\widehat{L}_F(\gamma)}
{\widehat{L}_F(\bar{\gamma})+\widehat{L}_F(\gamma)}.
\)
We set $R=0$ when the denominator vanishes. Positive values indicate lower forward cost, while negative values favor reverse traversal. Near-zero asymmetry does not exclude local preferences, which may cancel along the path. The field bound guarantees $|R|\leq\rho$; we also report $R/\rho$ to normalize by the permitted wind magnitude.

\begin{wrapfigure}{r}{0.40\columnwidth}
    \centering
    \vspace{-0.8em}
    \includegraphics[width=\linewidth]{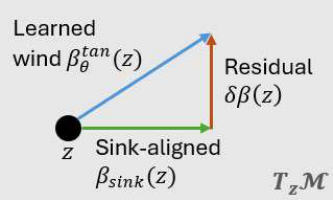}
    \caption{
    \textbf{Sink--residual decomposition.} The learned tangent field separates into a sink-aligned component and an orthogonal residual.
    }
    \label{fig:wind_decomposition}
    \vspace{-1.2em}
\end{wrapfigure}

\noindent\textbf{Decomposition relative to a common attractor.}
Directional preferences may partly reflect convergence toward a shared representation state. 
For example, as a bus progressively enters an image, its representations may move toward a common state corresponding to a fully visible bus. Such convergence could explain the observed directionality through a simple attractor (sink), without requiring a more complex directional field. 
Let $c\in\mathbb{S}^{d-1}$ be a reference attractor estimated from training data for each process family and held fixed during evaluation. 
Its unit spherical sink direction is
\(
u_{\mathrm{sink}}(z;c)=\frac{P_zc}{\|P_zc\|_2},
\)
with $u_{\mathrm{sink}}=0$ when $P_zc=0$. 
This direction provides a reference axis for separating attractor-aligned motion from other local preferences.

We project the learned tangent field onto this axis (Fig.~\ref{fig:wind_decomposition}):
\(
a(z)=\beta_\theta^{\mathrm{tan}}(z)^\top u_{\mathrm{sink}}(z),
\quad
\beta_{\mathrm{sink}}(z)=a(z)u_{\mathrm{sink}}(z),
\quad
\delta\beta(z)=\beta_\theta^{\mathrm{tan}}(z)-\beta_{\mathrm{sink}}(z).
\) 
By construction, $\delta\beta(z)^\top u_{\mathrm{sink}}(z)=0$. The coefficient $a(z)$ retains the signed alignment, distinguishing attraction ($a>0$) from motion away from the attractor ($a<0$). We measure the fraction of field energy along the sink axis by
\(
Q_{\mathrm{sink}}=
\mathbb{E}_z\left[
\frac{a(z)^2}{\|\beta_\theta^{\mathrm{tan}}(z)\|_2^2+\epsilon}
\right],
\)
where $\epsilon>0$ ensures numerical stability. Since this quantity uses squared alignment, it measures axis dependence rather than convergence alone.

\noindent\textbf{Residual directional structure.}
The residual satisfies $\|\delta\beta(z)\|_2\leq\|\beta_\theta^{\mathrm{tan}}(z)\|_2\leq\rho$ and therefore defines a valid bounded directional ruler,
\(
F_{\mathrm{res}}(z,v)=\|v\|_2-\delta\beta(z)^\top v.
\)
We evaluate this residual using the same directional discrimination and path-asymmetry measures. Predictive residual structure reveals preferences beyond the selected common-attractor axis, without ruling out other forms of convergence. Section~\ref{sec:experiments} examines these properties on held-out trajectories.


%% file: sec/4_experiment.tex
\vspace{-1.5em}
\section{Experiments}
\label{sec:experiments}
\vspace{-1.0em}

Section~\ref{sec:experimental_setup} describes the datasets and visual encoders. Section~\ref{sec:directional_existence} establishes empirical evidence of directional structure in visual trajectories. Section~\ref{sec:local_directionality} examines whether the learned field captures local structure beyond global convergence. Section~\ref{sec:process_specificity} characterizes process-dependent directional fields and whole-path asymmetry. Finally, Section~\ref{sec:restoration} evaluates the functional utility of the learned geometry through image restoration selection.

\vspace{-0.5em} 
\subsection{Experimental Setup}
\label{sec:experimental_setup}
\vspace{-0.5em} 

\begin{figure}[!htbp]
    \vspace{-1.0em} 
    \centering
    \includegraphics[width=0.8\columnwidth]{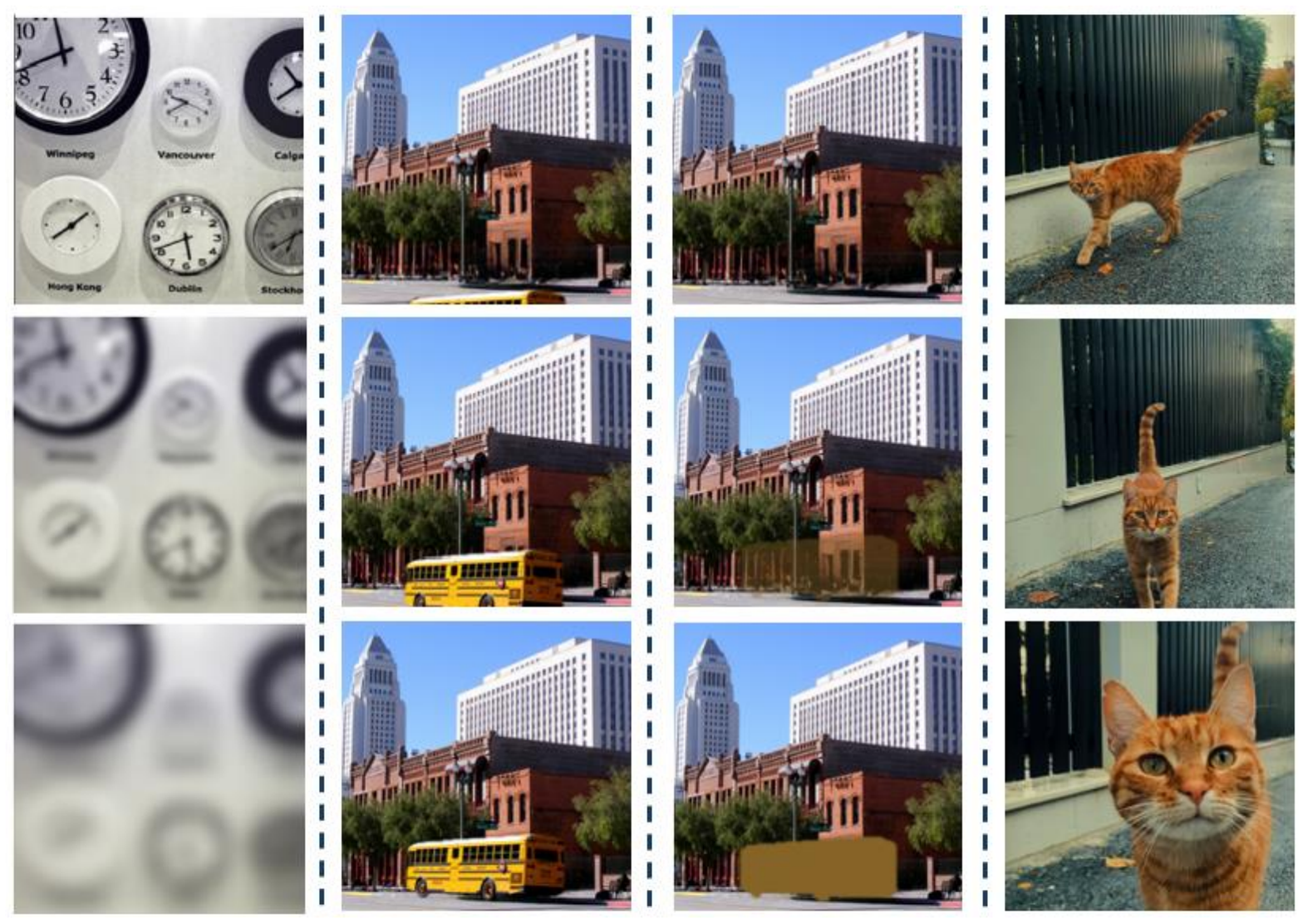}
    \vspace{-2.25em} 
    \caption{\textbf{Representative visual trajectories.} Left to right: progressive Gaussian blur (global appearance), bus entering (object-level), matched blob control, and cat approaching (natural video). Each column shows three ordered states, top to bottom. Controlled trajectories train the geometry; natural video tests transfer without adaptation. Further examples and construction details appear in the appendix.}
    \label{fig:dataset_eg}
    \vspace{-1.2em} 
\end{figure}

\noindent\textbf{Data.}
We evaluate DiRGeo using three groups of controlled visual trajectories and a separate natural-video set (see examples in Fig.~\ref{fig:dataset_eg}). The controlled benchmark comprises 17 families of ordered, 10-frame trajectories constructed from COCO~\cite{lin2014coco} images. Eight global appearance transformations (blur, Gaussian noise, JPEG compression, brightness, and four color tints) probe directional structure under progressively changing image-wide conditions. Four object-level processes (\textsc{Insert}, \textsc{Enter}, \textsc{Grow}, and \textsc{Colour}) examine process-dependent geometry, while five matched controls test whether the observed effects depend on meaningful object content. The global-transformation dataset contains 8,000 training and 2,000 test images; the object-level processes and their controls share 800 training and 200 test images, disjoint from the global-transformation dataset. All splits are by source image. We additionally evaluate five object-category splits and 200 new images for same-endpoint journey comparisons. To examine transfer beyond controlled transformations, we evaluate the photo-trained fields without adaptation on 10 natural videos containing 21 annotated event intervals. Full construction protocols are provided in the appendix.

\noindent \textbf{Representations and training.}
We use three frozen encoders with different pretraining
objectives: CLIP ViT-B/16~\cite{radford2021clip}, DINOv2 ViT-B/14~\cite{oquab2024dinov2}, and a supervised
ViT-B/16~\cite{dosovitskiy2021vit}. We extract CLS features after blocks 4, 8, and 12, together with CLIP's projected image embedding, yielding 10 representations and 170 family--representation combinations.
All features are $\ell_2$-normalized.
For each combination, we train a separate three-layer wind network with hidden width 512 and norm bound $\rho=0.9$ for 30 epochs using three random seeds.
Training uses ordered trajectories without semantic class labels. 
Results are averaged over three seeds, with 95\% confidence intervals obtained by bootstrapping source images (2,000 resamples); full evaluation protocols are provided in the appendix. 


\vspace{-0.5em}
\subsection{Empirical Evidence of Directional Structure}
\label{sec:directional_existence}
\vspace{-0.5em}

We first test whether the learned directional ruler captures reproducible motion preferences on unseen images. We evaluate three complementary measures. Reverse accuracy tests whether $F_\theta(z,v)<F_\theta(z,-v)$, while shuffled accuracy compares the observed displacement with a length-matched displacement from another image of the same process. Both measure the percentage of correctly preferred directions, with $50\%$ as the chance reference. To test whether these local preferences accumulate, we additionally measure whole-path asymmetry $R/\rho$ (Section~\ref{sec:characterization}), where $\rho=0.9$, zero indicates equal forward and reverse costs, and positive values favor forward traversal. All evaluations use sphere-tangent displacements and held-out images.

\noindent\textbf{Directional preferences generalize across all tested processes and accumulate along visual trajectories.} 
Reverse and shuffled accuracy exceed chance in all 170 family--representation combinations, with every pointwise 95\% confidence interval above $50\%$ (Table~\ref{tab:h1_summary}). Global appearance transformations exhibit particularly strong discrimination, while object-level processes show weaker but consistent preferences. Whole-path $R/\rho$ is also positive in all 170 combinations, with every confidence interval excluding zero. These results establish reproducible local and accumulated asymmetry across three encoder families, although directionality also occurs in matched controls and therefore does not, by itself, establish semantic dependence. The smaller $R/\rho$ values for some object-level processes indicate modest relative cost asymmetry despite reliable local directional discrimination, emphasizing that these measures capture complementary properties. 
These findings motivate directional geometry by revealing reproducible orientation-dependent structure that symmetric metrics cannot capture.

\begin{table}[!htbp]
    \vspace{-0.5em}
    \centering
    \small
    \setlength{\tabcolsep}{4pt}
    \begin{tabular}{lccc}
        \hline
        Trajectory group & Reverse (\%) & Shuffled (\%) & $R/\rho$ \\
        \hline
        Global appearance & 84.8--100.0 & 76.0--100.0 & 0.129--0.962 \\
        Object-level & 64.1--92.7 & 56.5--78.2 & 0.014--0.413 \\
        Matched controls & 67.6--99.8 & 59.5--82.7 & 0.036--0.585 \\
        \hline
    \end{tabular}
    \vspace{-0.5em}
    \caption{\textbf{Directional discrimination and whole-path asymmetry.} Ranges over family--representation combinations, averaged across three seeds. In all 170 cases, pointwise 95\% confidence intervals exceed $50\%$ for both accuracies and zero for $R/\rho$.}
    \label{tab:h1_summary}
    \vspace{-1.5em}
\end{table}



\vspace{-0.5em}
\subsection{Beyond a Common Attractor}
\label{sec:local_directionality}
\vspace{-0.5em}

We next ask whether the learned directional preferences can be explained simply by motion toward a common destination in representation space. We examine this explanation through two complementary tests.

First, we fit a spherical sink to approximate the learned field. This baseline uses a single attractor $c$ on the unit sphere, defining a direction $u_{\mathrm{sink}}(z;c)=P_zc/\|P_zc\|_2$ at each representation state. We then remove the learned field's component along this direction, obtaining the orthogonal residual $\delta\beta(z)=\beta_\theta^{\mathrm{tan}}(z)-\beta_{\mathrm{sink}}(z)$, as illustrated in Fig.~\ref{fig:wind_decomposition}. We evaluate whether this residual still assigns lower costs to observed displacements than to length-matched shuffled alternatives. Accuracy above $50\%$ indicates directional information beyond the fitted attractor axis.

Second, we test whether the complete learned field outperforms simpler attractor-based fields. Our primary baseline is the spherical sink fitted to approximate the learned wind. We also consider four alternatives: a spherical sink toward the mean final training representation, a spherical sink trained directly for directional discrimination, and fixed or learned straight-line sinks. 
For a stronger comparison, we select the sink with the highest held-out shuffled accuracy among these five baselines for each process--representation combination. 
We measure the shuffled-accuracy advantage over the primary fitted sink as $\Delta_{\mathrm{local}}=\mathrm{Acc}_{\mathrm{learned}}-\mathrm{Acc}_{\mathrm{fit}}$, in percentage points; positive values favor DiRGeo. All comparisons use held-out images and sphere-tangent displacements.

\noindent\textbf{The learned field retains directional information beyond a common attractor, although simple sinks remain competitive for some processes.}
After removing the sink-aligned component, residual shuffled accuracy remains above chance in all 170 combinations, with every pointwise 95\% confidence interval exceeding $50\%$ (Table~\ref{tab:h2_summary}). For object-level processes, residual accuracy ranges from $54.8\%$ to $78.8\%$, demonstrating predictive information outside the reference sink axis. The full learned field also improves over the field-fitted spherical sink by $1.2$--$45.6$ percentage points across the tested combinations, with significant gains in 169/170 cases. However, the stronger comparison against the best of five sinks yields significant advantages in all 80 global appearance cases but only 24/40 object-level cases; eight object-level comparisons are unresolved and eight favor a sink. Thus, a single fitted attractor cannot fully explain the learned directional structure, although alternative attractor models remain competitive for certain processes. Additional comparisons and synthetic validation are provided in the appendix. 

\begin{table}[!htbp]
    \vspace{-1.0em}
    \centering
    \small
    \setlength{\tabcolsep}{3pt}
    \begin{tabular}{lccc}
        \hline
        Trajectory group & Residual (\%) & $\Delta_{\mathrm{local}}$ (pp) & DiRGeo wins\\
        \hline
        Global appearance & 81.6--99.8 & +7.4 to +45.6 & 80/80 \\
        Object-level & 54.8--78.8 & +1.2 to +27.8 & 24/40 \\
        Matched controls & 59.3--85.8 & +4.4 to +30.3 & 28/50 \\
        \hline
    \end{tabular}
    \vspace{-0.75em}
    \caption{\textbf{Directional structure beyond a common attractor.} Residual accuracy evaluates $\delta\beta(z)$ after removing the sink-aligned component $\beta_{\mathrm{sink}}(z)$. $\Delta_{\mathrm{local}}$ measures shuffled-accuracy gains over the field-fitted spherical sink (pp). Values are ranges across combinations; the final column counts significant DiRGeo wins against the strongest of five sinks, selected by held-out shuffled accuracy. All 170 residual accuracies exceed chance with pointwise 95\% confidence, and 169/170 fitted-sink comparisons significantly favor DiRGeo.}
    \label{tab:h2_summary}
    \vspace{-1.5em}
\end{table}


\vspace{-0.5em}
\subsection{Process-Dependent Directional Geometry}
\label{sec:process_specificity}
\label{sec:process_directionality}
\vspace{-0.5em}

Here, we ask whether different visual processes induce different geometric preferences, and how these preferences accumulate along actual trajectories.

We first compare fields learned separately for different processes, such as \textsc{Insert} and \textsc{Enter}. Both fields are evaluated at the same representation states, so their differences cannot be attributed simply to visiting different locations. We measure their mean cosine similarity, where values near one indicate similar preferred directions and values near zero indicate nearly orthogonal preferences. We repeat this comparison after removing each field's sink-aligned component $\beta_{\mathrm{sink}}(z)$ to determine whether process differences persist in the residual $\delta\beta(z)$.

We then examine how directional preferences accumulate along different visual processes using the whole-path measure $R/\rho$ defined in Section~\ref{sec:characterization}. Unlike field similarity, which compares preferred directions at a fixed state, $R/\rho$ measures the resulting forward--reverse cost difference along an observed trajectory. We evaluate controlled transformations and matched controls, then apply the photo-trained fields to natural-video events without adaptation to examine transfer beyond synthetic trajectories.

\noindent\textbf{Different visual processes induce increasingly distinct local directional fields at greater representation depth.}
The \textsc{Insert}, \textsc{Grow}, and \textsc{Enter} fields are relatively aligned in shallow representations but become more differentiated at depth. For example, \textsc{Insert}--\textsc{Enter} similarity decreases from $0.66$ to $0.25$ in CLIP, $0.81$ to $0.16$ in DINOv2, and $0.74$ to $0.15$ in supervised ViT (Table~\ref{tab:h3_summary}). The same trend persists after removing sink-aligned components, indicating that process differences extend beyond common convergence. In contrast, \textsc{Insert} and \textsc{Colour} are nearly orthogonal even in shallow representations ($-0.04$ to $0.01$). Thus, greater depth does not universally create process differences; rather, it increasingly separates processes whose directional fields initially overlap.

\begin{table}[!htbp]
    \vspace{-0.75em}
    \centering
    \small
    \setlength{\tabcolsep}{3pt}
    \begin{tabular}{lrrrrrr}
        \hline
        & \multicolumn{3}{c}{Full field}
        & \multicolumn{3}{c}{Residual field} \\
        Encoder & L4 & L8 & L12 & L4 & L8 & L12 \\
        \hline
        CLIP
        & 0.66 & 0.48 & 0.25 & 0.48 & 0.37 & 0.23 \\
        DINOv2
        & 0.81 & 0.55 & 0.16 & 0.57 & 0.45 & 0.16 \\
        Supervised ViT
        & 0.74 & 0.42 & 0.15 & 0.53 & 0.36 & 0.15 \\
        \hline
    \end{tabular}
    \vspace{-0.5em}
    \caption{\textbf{Process-dependent directional fields.} Mean cosine similarity between \textsc{Insert} and \textsc{Enter} fields at identical states, before ($\beta_\theta^{\mathrm{tan}}$) and after ($\delta\beta$) sink removal. L4, L8, and L12 denote transformer-block features. Values near one indicate alignment; near zero, orthogonality. Both similarities decrease with depth across encoders.}
    \label{tab:h3_summary}
    \vspace{-1.5em}
\end{table}

\noindent\textbf{Accumulated asymmetry depends on the process and the observed event.}
Global appearance transformations exhibit greater whole-path asymmetry ($R/\rho=0.129$--$0.962$) than object-level processes ($0.014$--$0.413$) within each representation. Matched controls often produce comparable or stronger asymmetry than real-object processes, showing that directional strength alone does not establish semantic dependence. We further test whether these preferences transfer to natural video, where chronological order need not correspond to the direction learned from a particular photo-editing process. Across complete videos, $R/\rho$ remains near zero ($-0.0203$ to $0.0203$). However, separating videos into annotated events reveals a consistent pattern: photo-trained \textsc{Insert}, \textsc{Grow}, and \textsc{Enter} fields favor forward traversal during approach and reverse traversal during recession (Table~\ref{tab:h4_video_summary}). For example, the \textsc{Grow} field yields $+0.0135$ for approach and $-0.0231$ for recession, while ordinary motion and static intervals have much smaller mean asymmetry. These findings suggest that the learned geometry responds to the direction of the observed visual change rather than imposing a universal forward-time preference. Further controls and uncertainty analyses are provided in the appendix.

\begin{table}[!htbp]
    \vspace{-0.5em}
    \centering
    \small
    \setlength{\tabcolsep}{4pt}
    \begin{tabular}{lrrr}
        \hline
        Video event & Insert & Grow & Enter \\
        \hline
        Approach
        & $+0.0100$ & $+0.0135$ & $+0.0126$ \\
        Recede
        & $-0.0158$ & $-0.0231$ & $-0.0217$ \\
        Ordinary motion
        & $-0.0025$ & $-0.0036$ & $-0.0025$ \\
        Static
        & $+0.0032$ & $+0.0015$ & $+0.0013$ \\
        \hline
    \end{tabular}
    \vspace{-0.5em}
    \caption{\textbf{Event-dependent directionality in natural video.} Whole-path $R/\rho$ from photo-trained fields evaluated without adaptation on 10-state video windows (0.2\,s stride), averaged over 10 representations and three seeds. Positive and negative values favor forward and reverse traversal, respectively.}
    \label{tab:h4_video_summary}
    \vspace{-1.0em}
\end{table}

\vspace{-0.5em}
\subsection{Functional Validation via Image Restoration}
\label{sec:restoration}
\vspace{-0.5em}

We ask whether directional geometry can guide a practical decision: selecting a correction for a degraded image without access to its clean reference. This provides functional validation rather than a new restoration benchmark. DiRGeo does not generate restored images; instead, we use its geometric characterization to rank predefined candidate corrections.

\begin{figure}[!htbp]
    \vspace{-1.0em}
    \centering
    \includegraphics[width=0.9\linewidth]{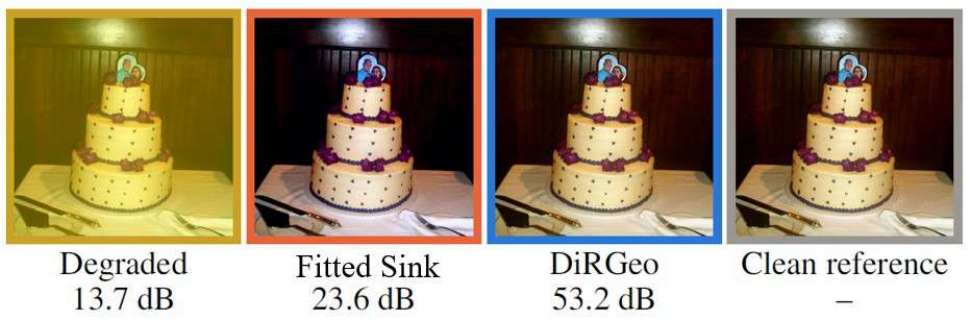}
    \vspace{-1.0em}
    \caption{\textbf{Colour-restoration selection using directional geometry.} A held-out yellow-cast example showing the degraded image, fitted-sink selection, DiRGeo selection, and clean reference. Both geometric scores share other-process penalties. The fitted sink overcorrects (23.6\,dB), while DiRGeo selects the exact inverse (53.2\,dB). Clean references are used only for evaluation; aggregate results appear in Table~\ref{tab:h7_summary}, with further examples in the appendix.}
    \vspace{-0.5em}
    \label{fig:restoration}
    \vspace{-0.5em}
\end{figure}

For a known degradation process $p$, such as a red colour cast, let $\eta$ denote the spherical path from the degraded representation to a candidate correction. We compute $A_p^{\mathrm{learn}}(\eta)=-\int_\eta \beta_{\theta,p}^{\mathrm{tan}}(z)^\top dz$, integrating against the full learned wind. 
We use the directional component rather than the full Randers length to score whether a candidate reverses the learned degradation process. 
A larger score indicates stronger movement against the learned degradation direction. We also compute $A_p^{\mathrm{sink}}(\eta)=-\int_\eta \beta_{\mathrm{fit},p}(z)^\top dz$, using the spherical sink fitted to approximate that wind. 
Here, the fitted sink $\beta_{\mathrm{fit},p}$ points toward the reference attractor used in Section~\ref{sec:characterization}, but replaces the learned field rather than decomposing it. 
%
Moving against one degradation may overshoot into another (e.g., correcting red into cyan). We therefore discourage candidate paths that move toward the fitted sinks of other global appearance processes:
\(
S_p^{\mathrm{learn}}=A_p^{\mathrm{learn}}-P_p,
\quad
P_p=\sum_{q\ne p}\max\{0,-A_q^{\mathrm{sink}}\}.
\)
A negative $A_q^{\mathrm{sink}}$ indicates movement toward degradation $q$ and incurs a penalty. We compare three geometric selection rules: the learned field with these penalties, the fitted sink with identical penalties ($S_p^{\mathrm{sink}}=A_p^{\mathrm{sink}}-P_p$), and the learned field alone ($A_p^{\mathrm{learn}}$). These comparisons reveal how the learned directional field and its simpler attractor-based approximation contribute to restoration selection.
%
We evaluate 200 new images with eight candidate corrections per colour cast. Top-1 accuracy measures whether the highest-scored candidate achieves the best PSNR; clean references are used only for evaluation. We compare against no-reference image-quality estimators MUSIQ~\cite{ke2021musiq}, BRISQUE~\cite{mittal2012brisque}, and NIQE~\cite{mittal2013niqe}, together with nearest-to-degraded and random selection as sanity checks. Fig.~\ref{fig:restoration} presents colour-correction outcome examples. Additional blur experiments, including candidate-ranking analysis, are provided in the appendix.

\noindent\textbf{Directional geometry provides useful guidance for restoration selection.}
Among eight candidate colour corrections, the learned field with other-process sink penalties selects the highest-PSNR correction in $31.8\%$ of cases, compared with $27.8\%$ for the fitted-sink score, $21.4\%$ without penalties, and $12.5\%$ for random selection (Table~\ref{tab:h7_summary}).
The paired improvement over the fitted sink is $3.98$ percentage points (95\% CI: $[2.76,5.14]$), with positive pooled gains across all four colour casts. 
Although the pooled improvement is statistically significant, its magnitude varies across encoder representations.
These results provide initial functional evidence that learned directional geometry can guide restoration selection beyond a fitted attractor. 
Complementary blur experiments, category generalization, and same-endpoint journey comparisons provide further characterization of the learned geometry in the appendix.

\begin{table}[!htbp]
    \vspace{-0.5em}
    \centering
    \small
    \setlength{\tabcolsep}{5pt}
    \begin{tabular}{lr}
        \hline
        Selection score & Colour top-1 (\%) \\
        \hline
        Learned field + penalties & \textbf{31.8} \\
        Fitted sink + penalties & 27.8 \\
        Learned field only & 21.4 \\
        \hline
        MUSIQ & 18.2 \\
        BRISQUE & 9.1 \\
        NIQE & 8.8 \\
        Nearest to degraded & 0.0 \\
        Random & 12.5 \\
        \hline
    \end{tabular}
    \vspace{-0.5em}
    \caption{\textbf{Colour-correction selection on 200 held-out images.} Top-1 accuracy identifies the highest-PSNR candidate across four casts. Geometric scores use the learned field or fitted sink, with or without other-process penalties, averaged over 10 representations and three seeds. MUSIQ, BRISQUE, and NIQE are no-reference quality estimators; nearest-to-degraded and random are sanity checks. Clean references are used only for evaluation.}
    \label{tab:h7_summary}
    \vspace{-1.0em}
\end{table}

%% file: sec/X_suppl.tex
\clearpage
\setcounter{page}{1}
\maketitlesupplementary


\hypersetup{
    hidelinks
}

\appendix
\etocdepthtag.toc{appendix}

\section*{Appendix}

\begingroup
\makeatletter

\renewcommand*\l@section{\@dottedtocline{1}{0em}{1.75em}}

\etocsettagdepth{main}{-2}
\etocsettagdepth{appendix}{1}
\etocsetnexttocdepth{1}
\etocsettocstyle{}{}
\tableofcontents

\makeatother
\endgroup


\section{Extended Experimental Setup and Implementation Details}
\label{app:experimental_setup}

This section extends Section~\ref{sec:experimental_setup} with the datasets, encoder checkpoints, feature extraction, training configurations, and evaluation protocols used throughout DiRGeo. We provide sufficient implementation details to reproduce the shared experimental pipeline, while experiment-specific constructions, metrics, and results are described in the corresponding appendix sections.

\subsection{Trajectory Datasets}
\label{app:trajectory_datasets}

\begin{figure*}[!htp]
    \centering
    \begin{minipage}[t]{0.37\textwidth}
        \vspace{0pt}
        \begin{subfigure}{\linewidth}
            \centering
            \includegraphics[width=\linewidth]{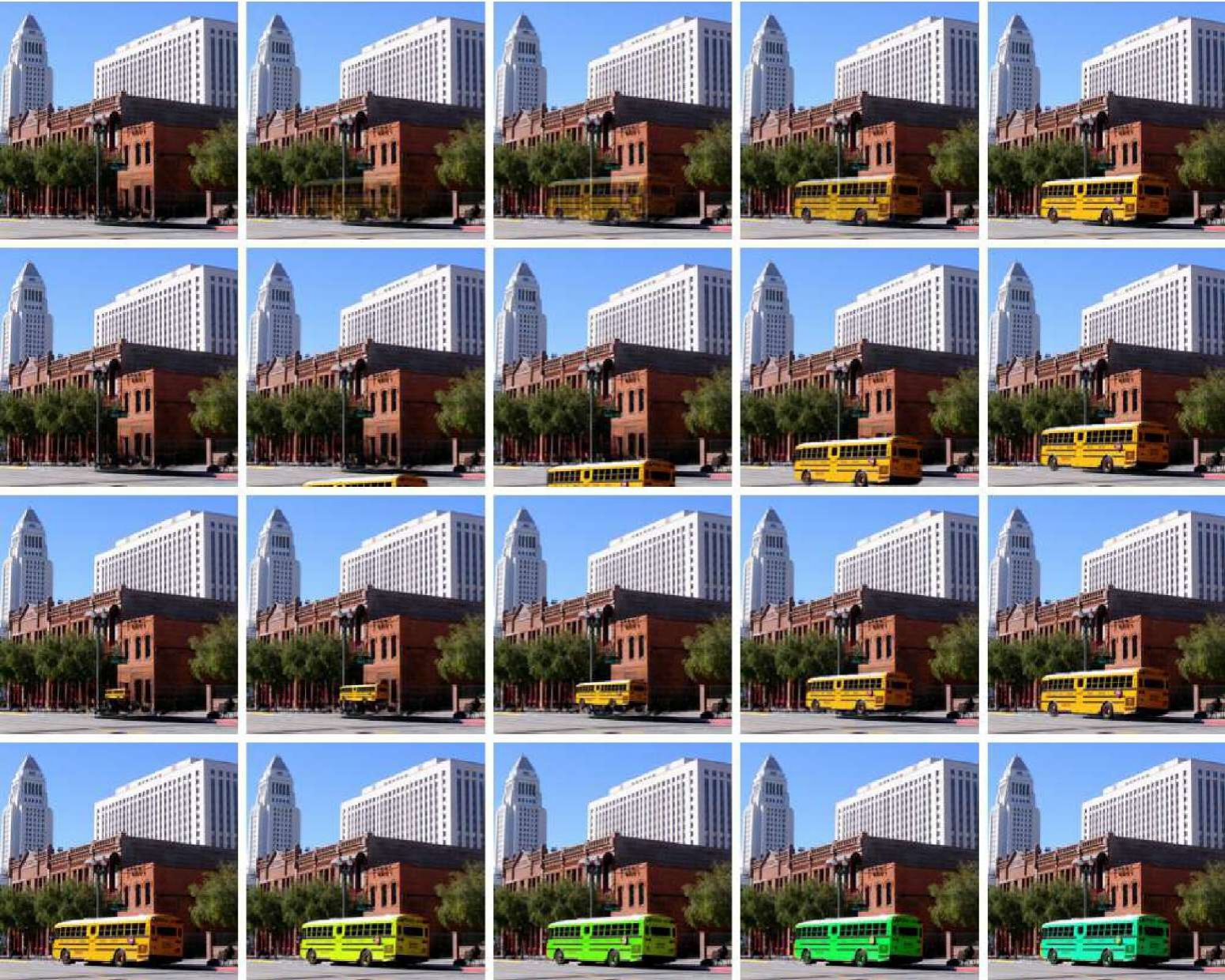}
            \caption{Object-level processes}
            \label{fig:app_object}
        \end{subfigure}\\[3pt]
        \begin{subfigure}{\linewidth}
            \centering
            \includegraphics[width=\linewidth]{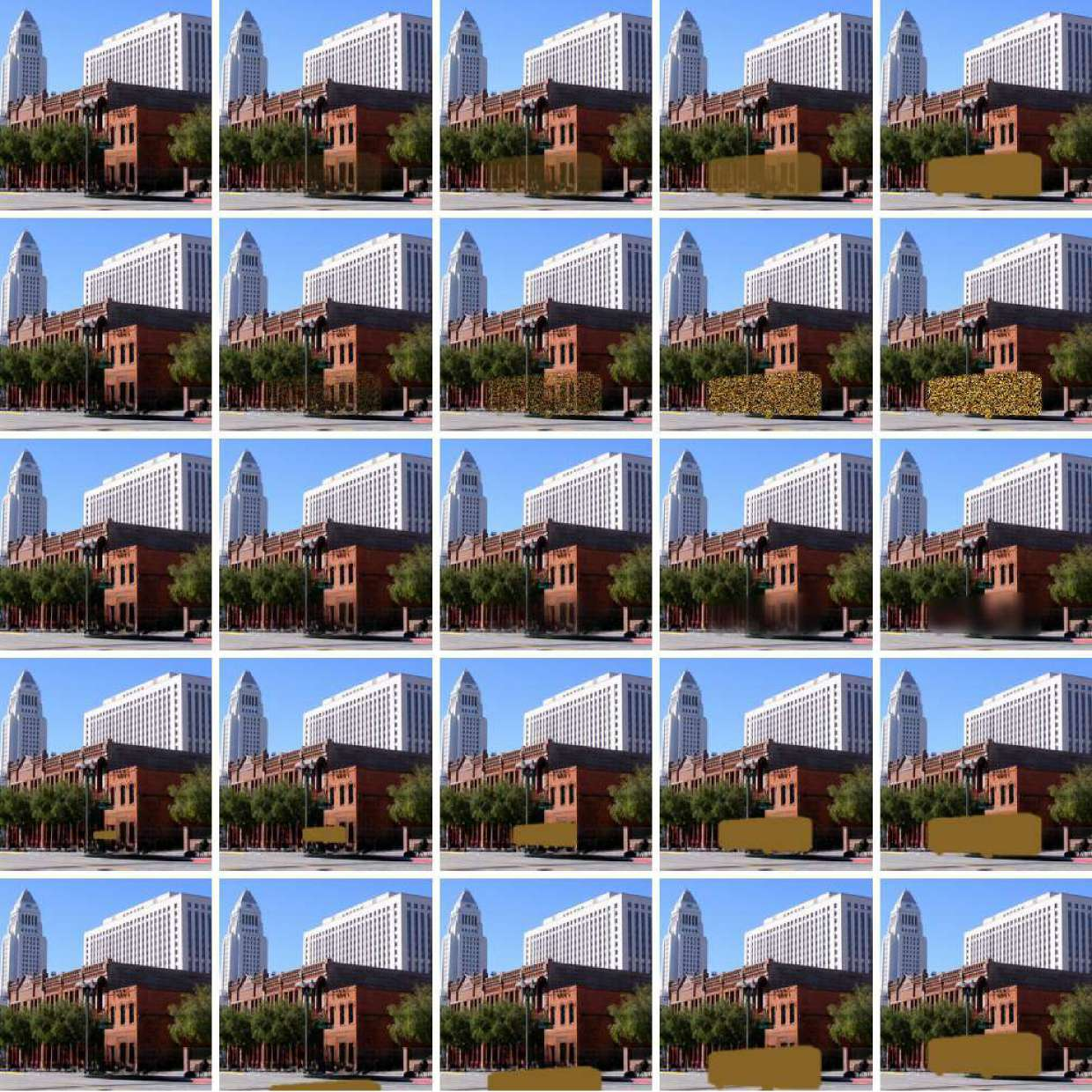}
            \caption{Matched controls}
            \label{fig:app_controls}
        \end{subfigure}\\[3pt]
        \begin{subfigure}{\linewidth}
            \centering
            \includegraphics[width=\linewidth]{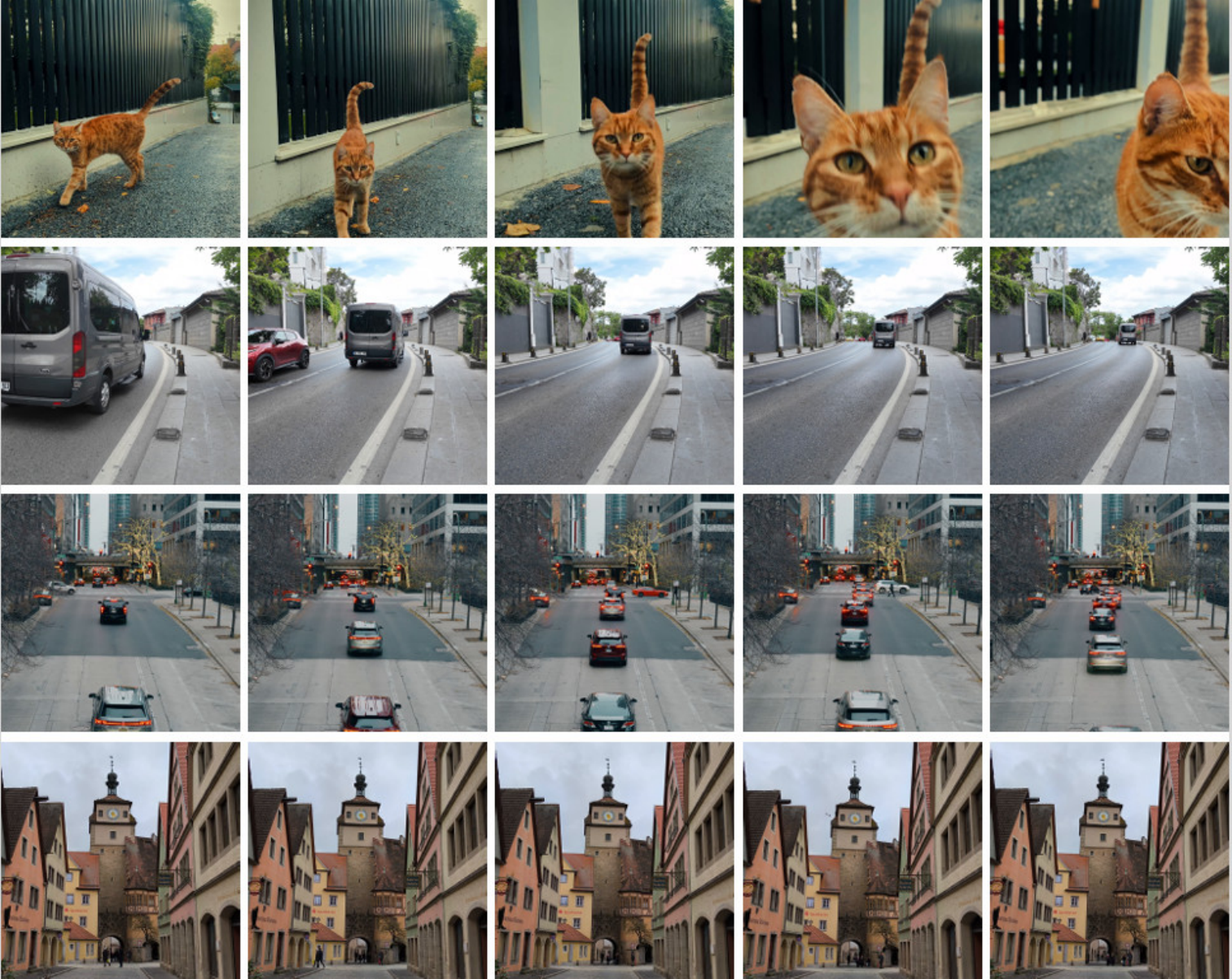}
            \caption{Natural-video event intervals}
            \label{fig:app_video}
        \end{subfigure}
    \end{minipage}\hfill
    \begin{minipage}[t]{0.60\textwidth}
        \vspace{0pt}
        \begin{subfigure}{\linewidth}
            \centering
            \includegraphics[width=\linewidth,height=0.78\textheight,keepaspectratio]{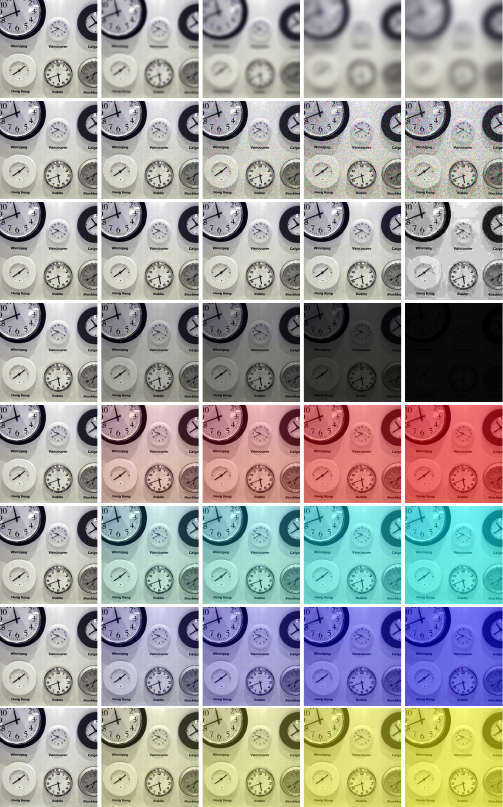}
            \caption{Global appearance transformations}
            \label{fig:app_nuisance}
        \end{subfigure}
    \end{minipage}
    \caption{\textbf{Controlled trajectory families and natural-video examples.} Panels (a), (b), and (d) show states $t=0,2,4,7,9$ from 10-frame controlled trajectories. (a) Object-level processes: \textsc{Insert}, \textsc{Enter}, \textsc{Grow}, and \textsc{Colour}. (b) Matched controls: Insert-blob, Insert-scramble, Insert-blur, Grow-blob, and Enter-blob. (c) Five evenly spaced frames from annotated natural-video intervals: approach (video 12503985), recede (16235000), ordinary motion (12813896), and static (15526828). (d) Eight global appearance transformations applied to one source image with increasing severity: blur, noise, JPEG, brightness, and red, cyan, blue, and yellow tints. Controlled trajectories provide training supervision, while natural videos test transfer to observed events.}
    \label{fig:appendix_trajectory_datasets}
\end{figure*}

We construct controlled visual trajectories to isolate different processes of image change and use natural videos to evaluate transfer to observed events. Figure~\ref{fig:appendix_trajectory_datasets} illustrates the trajectory families, including the intermediate states omitted from the representative examples in the main paper.

\paragraph{Global appearance transformations.}
We construct eight trajectory families from COCO train2017: Gaussian blur, Gaussian noise, JPEG compression, brightness adjustment, and red, cyan, blue, and yellow colour tints. Each family contains 10-frame trajectories with progressively increasing transformation severity. These transformations modify image-wide appearance without requiring object-level editing. The dataset contains 8,000 training and 2,000 test source images, with all trajectories from a source image assigned to the same split. The 10,000 source images are disjoint from the object-level dataset. Figure~\ref{fig:app_nuisance} illustrates the eight processes applied to a common source image, making their different transformation directions and severity progressions directly comparable.

\paragraph{Object-level processes and matched controls.}
The object-level dataset comprises 1,000 COCO images, partitioned into 800 training and 200 test images. For each image, an object is removed using LaMa to obtain a background from which controlled trajectories are constructed. Four processes manipulate object content in different ways: \textsc{Insert} progressively introduces an object, \textsc{Enter} moves it into the scene, \textsc{Grow} increases its apparent size, and \textsc{Colour} changes its appearance. Their implementation names are \texttt{insert}, \texttt{enter}, \texttt{grow}, and \texttt{colour}.

We additionally construct five matched controls: \texttt{insert\_blob}, \texttt{insert\_scramble}, \texttt{grow\_blob}, \texttt{enter\_blob}, and \texttt{insert\_blur}. These controls alter the manipulated content while retaining the associated transformation construction, allowing us to distinguish directional effects associated with visual motion from those requiring intact object content. All four object-level families and five controls use identical source-image partitions and contain 10 frames per trajectory. In particular, \textsc{Insert} and \textsc{Enter} share their initial and final images but follow different intermediate paths, enabling the same-endpoint comparisons in Appendix~\ref{app:same_endpoints}. Figures~\ref{fig:app_object} and~\ref{fig:app_controls} show representative sequences from the two groups.

\paragraph{Preprocessing and split consistency.}
Frames are prepared using a central $256\times256$ view and resized to $224\times224$ for encoder input without an additional crop. The same image sequences, source-image partitions, and experimental random seeds are used across encoder families. All trajectories derived from a source image remain within its assigned partition, preventing overlap between training and test images through different transformations of the same source.

\paragraph{Additional evaluation data.}
For same-endpoint journey evaluation, we use 200 additional COCO train2017 images, disjoint from the 1,000 object-level images and 10,000 global appearance images. These trajectories are generated using the same dataset builder with seed 2027 instead of 42. Additional journey constructions preserve their endpoints while changing intermediate states. Previously trained fields are evaluated on these trajectories without retraining.

Object-category generalization is examined using five category splits, each holding out 40 of the 80 COCO categories. Category-restricted fields are compared with fields trained on all categories using the same evaluation images. This design tests how directional preferences depend on the object categories represented during training while controlling for the evaluation inputs.

\clearpage

\paragraph{Natural-video evaluation.}
We evaluate 10 unique natural videos sampled at $0.1$\,s intervals, yielding 1,626 frames. The 21 manually annotated event intervals comprise seven approach, four recede, five ordinary-motion, and five static intervals. Approach and recede describe motion toward and away from the camera, respectively; ordinary motion does not exhibit a clear approach or recession event. Figure~\ref{fig:app_video} illustrates representative intervals.

We consider both whole-video trajectories and windows restricted to the annotated events. Fields trained on controlled photo trajectories are transferred directly to the video representations without adaptation. As an additional ordering control, separately trained video fields are evaluated using leave-one-video-out testing. The construction of temporal windows, evaluation strides, and video-level uncertainty estimates are detailed in the natural-video experiments.


\subsection{Frozen Encoders and Feature Extraction}
\label{app:encoder_details}

\paragraph{Encoder checkpoints.}
We evaluate three pretrained visual encoders with different learning objectives. CLIP uses the OpenAI weights for \texttt{ViT-B-16-quickgelu} through \texttt{open\_clip}. DINOv2 uses the \texttt{timm} checkpoint \texttt{vit\_base\_patch14\_dinov2.lvd142m}. The supervised ViT uses \texttt{vit\_base\_patch16\_224.augreg\_in21k\_ft\_in1k}, pretrained on ImageNet-21k and fine-tuned on ImageNet-1k. These encoders represent image--text contrastive learning, image-only self-supervision, and supervised classification, respectively. Their different pretraining objectives allow us to examine whether the learned directional geometry depends on a particular representation-learning paradigm.

\paragraph{Representation selection.}
For each encoder, we extract the CLS token after transformer blocks 4, 8, and 12 and apply the encoder's final normalization layer to the selected features. We additionally evaluate CLIP's projected image embedding. This yields four representations from CLIP, three from DINOv2, and three from supervised ViT, for 10 representations in total. Each extracted representation is $\ell_2$-normalized before directional-field training. The encoder parameters remain frozen throughout all training and evaluation procedures; only the directional-field networks and applicable geometric baselines are fitted.

\subsection{Directional-Field Training}
\label{app:wind_training}

\paragraph{Architecture and parameterization.}
The directional field is parameterized by a three-layer MLP $b_\theta:\mathbb{R}^d\rightarrow\mathbb{R}^d$ with hidden width 512. The network consists of two Linear--LayerNorm--GELU stages followed by a linear output layer of dimension $d$. Its output is bounded according to Section~\ref{sec:parameterization}:
\(
\beta_\theta(z)=\rho\,b_\theta(z)/\max\{1,\|b_\theta(z)\|_2\},
\quad \rho=0.9.
\)
This ensures $\|\beta_\theta(z)\|_2\leq\rho<1$, preserving the Randers norm constraint under the fixed Euclidean base metric. The symmetric component of the local cost is not learned.

We train a separate directional field for each of the 17 trajectory families and 10 encoder representations. The resulting 170 family--representation configurations are each trained using three random seeds, yielding 510 trained wind networks. Each network receives only the current normalized representation $z$; future states and semantic category labels are not supplied as network inputs.

\paragraph{Training objective.}
Temporal ordering supplies supervision through three complementary terms. Directional InfoNCE contrasts the observed forward displacement with reversed, shuffled, and random alternatives, using temperature $\tau=0.07$. Shuffled negatives are drawn from other trajectories within the same process family, and candidate directions are matched in magnitude during contrastive learning. The smoothness regularizer penalizes changes in the predicted wind between consecutive states,
\(
\mathcal{L}_{\mathrm{smooth}}
=
\mathbb{E}_t\|\beta_\theta(z_{t+1})-\beta_\theta(z_t)\|_2^2.
\)
This encourages locally coherent field predictions without requiring a constant direction along an entire trajectory.


The multi-step regularizer tests directional consistency beyond adjacent frames. For $v_t^{(h)}=z_{t+h}-z_t$, the implemented hinge term is
\(
    \mathcal{L}_{\mathrm{multi}}
    =
    \mathbb{E}_{t,h}
    \left[
    0.1\|v_t^{(h)}\|_2-2\beta_\theta(z_t)^\top v_t^{(h)}
    \right]_+ ,
\)
where $[a]_+=\max(a,0)$. We use horizons $h\in\{2,3\}$ and omit terms for which $t+h$ exceeds the final frame. All valid $(t,h)$ pairs are weighted equally, and the margin is proportional to the step length. These horizons encourage the learned wind to favor net forward motion over short trajectory segments.

The complete objective is
\(
\mathcal{L}
=
\mathcal{L}_{\mathrm{dir}}
+
0.1\mathcal{L}_{\mathrm{smooth}}
+
0.05\mathcal{L}_{\mathrm{multi}}.
\)
The directional contrast is the primary learning signal, while the two regularizers encourage consistency across nearby representation states and temporal horizons. All training losses use ambient finite displacements; sphere-tangent evaluation is applied subsequently to assess whether the learned preferences persist after removing normalization-induced radial components.

\paragraph{Optimization and training schedule.}
The wind networks are trained for 30 epochs using AdamW with initial learning rate $3\times10^{-4}$ and weight decay $10^{-4}$. We apply cosine learning-rate annealing toward zero over the 30 epochs, updating the scheduler once per epoch. The batch size is 512, with the final incomplete batch dropped. Gradients are clipped to a maximum norm of 1.0.

The constant-field baseline uses the same optimizer and learning-rate schedule, with learning rate $3\times10^{-3}$ and weight decay $10^{-4}$. Learned straight-line and spherical sinks use learning rate $3\times10^{-3}$ without weight decay. The spherical sink fitted to the learned field is optimized using Adam with learning rate $10^{-2}$ for 1,500 steps, sampling 8,192 training states per step. These baselines are evaluated separately from the learned directional field to test whether simpler directional models explain the observed trajectory preferences.

\paragraph{Random seeds and reproducibility.}
All experiments use base seed 42. For model kind $k$, trajectory family $f$, and training run $s\in\{0,1,2\}$, the training seed is
\(
s_{\mathrm{train}}
=
42+100000s+1000k+f.
\)
The seed is applied to Python, NumPy, PyTorch, and CUDA. The implementation uses deterministic cuDNN settings and disables TF32. The test-negative sampling seed is $42+999$, and the bootstrap seed is $42+77$. Keeping evaluation negatives fixed permits paired comparisons between separately trained fields and baselines under the same sampled alternatives.

\subsection{Evaluation and Geometric Controls}
\label{app:shared_evaluation}

\paragraph{Local directional discrimination.}
We evaluate whether the learned field assigns lower directional costs to observed motion than to alternative directions at the same representation state. Reverse accuracy compares the observed step with its negation, measuring whether the field prefers the trajectory's observed orientation. Shuffled accuracy compares the observed displacement with a step sampled from another test image of the same process family, matched to the observed step length. Random-direction accuracy provides an additional control using sampled directions unrelated to the observed trajectories.

All three measurements are reported as the percentage of correctly preferred directions, with $50\%$ as the chance reference. Reverse discrimination tests orientation along the observed axis, while shuffled discrimination is more demanding because the alternative is itself an observed motion from the same visual process.

Primary local evaluations use the orthogonal sphere-tangent projection $P_z=I-zz^\top$ at the current representation state. This removes the radial component introduced by finite displacements between normalized features. It does not estimate the tangent space of the unknown representation manifold and therefore controls a specific normalization artifact rather than all possible geometric confounds. Detailed negative sampling, length matching, and numerical tie handling are provided in Appendix~\ref{app:directional_existence}.

\paragraph{Whole-path asymmetry.}
We evaluate each trajectory in its forward and reversed orders, recomputing the local directional cost at the departure state of every step. Following Section~\ref{sec:characterization}, we summarize the difference between the accumulated directional lengths through
\(
R(\gamma)/\rho
=
\frac{\widehat{L}_F(\bar{\gamma})-\widehat{L}_F(\gamma)}
{\rho\left(\widehat{L}_F(\bar{\gamma})+\widehat{L}_F(\gamma)\right)}.
\)
Positive values indicate lower forward cost, negative values favor the reversed trajectory, and zero indicates equal accumulated costs. The bound $|R|\leq\rho$ guarantees that this normalized measurement lies in $[-1,1]$ whenever the denominator is nonzero. We set the asymmetry to zero when both path lengths vanish.

Sphere-projected steps provide the primary whole-path measurements. Results computed using the original ambient chords are retained as a diagnostic comparison. We report whole-path asymmetry separately from local directional accuracy because individual preferences may cancel when accumulated over an entire trajectory.

\paragraph{Baseline fields and sink--residual decomposition.}
We compare the learned field with a state-independent constant field and five common-attractor baselines. These comprise a spherical sink toward the mean terminal training representation, a spherical sink with a learned attractor, a spherical sink fitted to approximate the learned wind, and fixed- and learned-attractor straight-line sinks. Spherical sinks point along the sphere-tangent direction toward an attractor, whereas straight-line sinks use Euclidean directions toward their reference points. A separate center-of-sphere sink provides a diagnostic for normalization-induced radial artifacts.

The field-fitted spherical sink serves as the primary reference for the decomposition in Section~\ref{sec:characterization}. Its fitted attractor $c$ defines the sink direction $u_{\mathrm{sink}}(z;c)$; projecting the learned tangent field onto this axis gives the sink-aligned component $\beta_{\mathrm{sink}}(z)$, and subtracting that component gives the orthogonal residual $\delta\beta(z)$. The fitted sink is a standalone comparison field, whereas $\beta_{\mathrm{sink}}(z)$ is a projection of the learned wind; they share the reference attractor but are not identical vector fields.

We separately compare the complete learned wind against the field-fitted spherical sink and against the strongest of all five sink baselines. For each family--representation combination, the strongest sink is the one with the highest held-out shuffled accuracy among the five alternatives. This is an evaluation-selected comparison designed to test whether any of the examined attractor-based fields can explain the observed directional discrimination. Additional controls train fields using reversed or randomly ordered frames to test whether the learned preferences depend on the intended trajectory ordering.

\paragraph{Statistical reporting.}
Photo-based results are averaged over three independently trained seeds. We compute pointwise 95\% confidence intervals using 2,000 bootstrap resamples of source images, retaining all dependent steps and trajectories associated with each sampled source. Paired comparisons resample the same source images for both methods, preserving the correspondence between their evaluations.

For natural-video experiments, confidence intervals are obtained by resampling whole videos rather than individual windows, accounting for dependence among windows from the same recording. Same-endpoint confirmation analyses resample their 200 source images while retaining the corresponding representations, training seeds, and journey comparisons.

The 170 family--representation combinations share images and related representations, so pointwise confidence intervals should not be interpreted as a simultaneous guarantee or as 170 independent replications. Experiment-specific aggregation rules, significance criteria, and additional controls are provided alongside their respective results.



\section{Extended Analysis of Directional Structure}
\label{app:directional_existence}

This section extends Section~\ref{sec:directional_existence} by providing the complete evaluation protocol, aggregate results, and per-family measurements supporting: ordered visual processes exhibit reproducible directional preferences in frozen visual representations. We evaluate 17 trajectory families across 10 representations extracted from three pretrained encoder families, yielding 170 family--representation combinations. Each directional field is trained with three random seeds and evaluated on held-out source images. We distinguish local directional discrimination from accumulated whole-path asymmetry because the two measurements characterize complementary properties of the learned geometry.

\subsection{Evaluation Protocol}
\label{app:h1_protocol}

\paragraph{Reverse discrimination.}
For an observed representation displacement $v$ at state $z$, reverse discrimination tests whether the learned directional ruler assigns a lower cost to the observed motion than to its reversal:
\begin{equation}
    F_\theta(z,v)<F_\theta(z,-v).
    \label{eq:h1_reverse_comparison}
\end{equation}
Both candidates are evaluated at the same representation state. This is distinct from reversing an entire trajectory, which changes the departure state at which the field is evaluated.

Under the Euclidean-base Randers formulation, the local cost difference is
\begin{equation}
    F_\theta(z,-v)-F_\theta(z,v)
    =
    2\beta_\theta(z)^\top v.
    \label{eq:h1_reverse_gap}
\end{equation}
Consequently, reverse discrimination tests whether the learned wind aligns positively with the observed displacement. Since the two candidates have identical magnitudes, their symmetric Euclidean contributions cancel. The comparison therefore measures orientation preference rather than displacement length.

\paragraph{Shuffled and random-direction discrimination.}
Reverse discrimination alone is relatively permissive: a field can distinguish $v$ from $-v$ without necessarily preferring the observed displacement over other plausible directions. We therefore evaluate a more demanding comparison using a shuffled alternative sampled from another test trajectory of the same process family. The alternative is taken from a different source image at a randomly sampled time step and rescaled to match the observed displacement length. Both steps are evaluated at the same anchor state $z$.

Shuffled accuracy measures how frequently the field assigns lower cost to the observed local motion than to an alternative motion from the same visual process. Random-direction accuracy provides an additional control using Gaussian directions sampled independently of the observed trajectories. These tests assess different properties: reverse discrimination measures orientation along the observed axis, while shuffled discrimination examines whether the observed direction is preferred over motion occurring elsewhere in the process distribution.

The shuffled comparison should not be confused with a random-order training control. Shuffled test negatives change the alternatives used during evaluation; random-order training changes the temporal supervision used to learn the field.

\paragraph{Sphere-tangent evaluation and length matching.}
The primary results use the sphere-tangent correction introduced in Section~\ref{sec:parameterization}. Given a unit-normalized representation $z$, the projection is $P_z=I-zz^\top$, and an observed displacement $v^+$ is evaluated through $P_zv^+$. This removes the radial component induced by finite chords between normalized representations without assuming that the full tangent space of the underlying representation manifold has been recovered.

Length matching is performed after tangent projection. Specifically, each shuffled or random negative $v^-$ is first projected as $P_zv^-=v^--(z^\top v^-)z$ and then rescaled to have exactly the magnitude $\|P_zv^+\|_2$. The reverse negative is constructed directly as $-P_zv^+$. Thus, every comparison uses alternatives with matched tangent-space magnitudes at the same evaluation state.

For the ambient-chord comparisons retained as geometric controls, shuffled and random negatives are instead rescaled to the magnitude $\|v^+\|_2$ in the original representation space. Keeping these procedures separate avoids conflating orientation preferences with changes in length introduced by tangent projection.

\paragraph{Numerical comparison and tie handling.}
A comparison counts as a correct preference when the observed step has strictly lower directional cost than its negative, $F_\theta(z,v^+)<F_\theta(z,v^-)$. An incorrect preference scores zero, while a tie contributes $0.5$.

For ambient-chord evaluation, only exact equality is treated as a tie. For sphere-tangent evaluation, numerical equality is determined using a relative tolerance. Writing $E^+=F_\theta(z,v^+)$ and $E^-=F_\theta(z,v^-)$ for the costs after the applicable projection and length matching, a tie is declared when
\begin{equation}
    |E^--E^+|
    \leq
    10^{-6}\left(|E^+|+|E^-|+10^{-12}\right).
    \label{eq:h1_tie_tolerance}
\end{equation}
This tolerance prevents small floating-point differences from being interpreted as meaningful directional preferences.

Shuffled negatives are sampled from different trajectories within the same family, while random negatives use Gaussian directions. The negative-sampling seed is reset before each evaluation, ensuring that different learned fields and baselines are compared using identical sampled alternatives for a given test set. Both accuracy measures have a $50\%$ chance reference.

\paragraph{Whole-path directional asymmetry.}
Local reverse discrimination does not establish that a directional preference persists over an entire trajectory. Individual preferences can reinforce or cancel as the path evolves. We therefore separately evaluate each trajectory in its forward and reversed orders, recomputing the directional cost at the corresponding departure state of every step.

The normalized forward--reverse asymmetry is
\begin{equation}
    \frac{R}{\rho}
    =
    \frac{L_F^{\leftarrow}-L_F^{\rightarrow}}
    {\rho\left(L_F^{\leftarrow}+L_F^{\rightarrow}\right)},
    \label{eq:h1_path_asymmetry}
\end{equation}
where $L_F^{\rightarrow}$ and $L_F^{\leftarrow}$ denote the discrete accumulated costs in the forward and reversed directions, respectively. The primary calculation uses sphere-projected displacements, and $\rho=0.9$ is the wind magnitude bound. Positive values indicate lower forward cost, negative values indicate lower reverse cost, and zero indicates equal accumulated costs. The normalization gives $|R/\rho|\leq1$ for the bounded directional ruler; a zero denominator is assigned zero asymmetry.

Unlike local accuracy, $R/\rho$ measures the magnitude and sign of the accumulated cost difference rather than the proportion of correctly preferred steps. A trajectory can therefore exhibit high reverse accuracy but modest whole-path asymmetry when its directional advantages are small relative to its symmetric path length.

\paragraph{Uncertainty and scope of inference.}
All photo-based results are averaged over three training seeds. We compute 95\% confidence intervals using 2,000 bootstrap resamples of source images, preserving the dependence among trajectory steps and transformations derived from the same sampled image. The bootstrap protocol is shared across the encoder families.

The statement that all 170 combinations exceed chance means that each individual family--representation combination has a pointwise 95\% confidence interval above $50\%$. It is not a simultaneous confidence guarantee across all 170 tests. Moreover, the combinations share images and related encoder representations, so they should not be treated as 170 statistically independent replications.

\subsection{Aggregate Directional Discrimination}
\label{app:h1_aggregate}

\paragraph{Consistency across pretraining objectives.}
Table~\ref{tab:h1_aggregate} summarizes the sphere-tangent reverse, shuffled, and random-direction accuracies across the three encoder families. We report global appearance transformations separately from the object-level dataset because the two groups differ in both transformation construction and the resulting strength of directional discrimination.

\begin{table*}[!htbp]
    \centering
    \small
    \setlength{\tabcolsep}{7pt}
    \begin{tabular}{llrrr}
        \hline
        Representation & Trajectory group
        & Reverse & Shuffled & Random \\
        \hline
        CLIP L4  & Global appearance & 100.0 & 99.2 & 100.0 \\
                 & Object dataset & 92.0 & 67.0 & 91.5 \\
        CLIP L8  & Global appearance & 100.0 & 99.7 & 100.0 \\
                 & Object dataset & 88.3 & 71.9 & 87.7 \\
        CLIP L12 & Global appearance & 99.9 & 97.6 & 99.8 \\
                 & Object dataset & 80.7 & 69.0 & 79.5 \\
        CLIP embedding
                 & Global appearance & 99.6 & 96.2 & 99.5 \\
                 & Object dataset & 79.8 & 68.4 & 78.7 \\
        \hline
        DINOv2 L4  & Global appearance & 100.0 & 98.3 & 100.0 \\
                   & Object dataset & 89.1 & 63.8 & 88.7 \\
        DINOv2 L8  & Global appearance & 99.9 & 98.3 & 99.9 \\
                   & Object dataset & 87.3 & 69.5 & 86.8 \\
        DINOv2 L12 & Global appearance & 96.6 & 91.8 & 96.0 \\
                   & Object dataset & 73.7 & 65.8 & 71.5 \\
        \hline
        Supervised ViT L4
                   & Global appearance & 100.0 & 98.4 & 100.0 \\
                   & Object dataset & 91.2 & 65.7 & 91.0 \\
        Supervised ViT L8
                   & Global appearance & 100.0 & 99.7 & 100.0 \\
                   & Object dataset & 84.7 & 71.6 & 84.4 \\
        Supervised ViT L12
                   & Global appearance & 99.9 & 98.0 & 99.8 \\
                   & Object dataset & 76.8 & 67.8 & 75.0 \\
        \hline
    \end{tabular}
    \caption{\textbf{Aggregate sphere-tangent directional accuracy (\%).} Global appearance rows average eight transformation families; object-dataset rows average four object-level processes and five matched controls. Results are averaged over three training seeds. L4, L8, and L12 denote transformer-block features, and the CLIP embedding is its projected image representation. All accuracies use $50\%$ as the chance reference.}
    \label{tab:h1_aggregate}
\end{table*}

The global appearance families exhibit particularly strong directional discrimination across all representations. Their aggregate reverse and random-direction accuracies are generally near saturation, while shuffled accuracy remains high despite using alternative displacements drawn from the same process family. This indicates that these controlled transformations provide highly consistent directional supervision in frozen representation space.

The object-level dataset also remains above chance across all evaluated representations, although shuffled accuracy is substantially lower. For example, CLIP L4 achieves $92.0\%$ reverse accuracy but $67.0\%$ shuffled accuracy when averaged over the object-level dataset and its matched controls. This difference illustrates why reversed motion and shuffled alternatives should be evaluated separately: recognizing the correct orientation of an observed step does not imply equally strong discrimination against other plausible directions.

The aggregate object-dataset rows combine all nine families: four object-level processes and five matched controls. They are not averages over object-level processes alone. We preserve this distinction in the full per-family analysis below to avoid attributing the directional behavior of matched controls exclusively to meaningful object content.

\paragraph{Local discrimination and accumulated asymmetry.}
Both reverse and shuffled accuracy exceed chance in all 170 family--representation combinations, with every pointwise 95\% confidence interval lying entirely above $50\%$. The smallest lower confidence bound across the two metrics is $54.2\%$, attained in the shuffled analysis. For the four object-level processes, reverse accuracy ranges from $64.1\%$ to $92.7\%$; for global appearance transformations, it ranges from $84.8\%$ to $100.0\%$.

The whole-path analysis provides complementary evidence. Sphere-projected $R/\rho$ is positive in all 170 combinations, with every pointwise 95\% confidence interval excluding zero. Thus, the learned preferences are not restricted to individual comparisons: their accumulated directional contributions also favor forward traversal along the tested trajectories.

\begin{figure*}[!htbp]
    \centering
    \includegraphics[width=\textwidth]{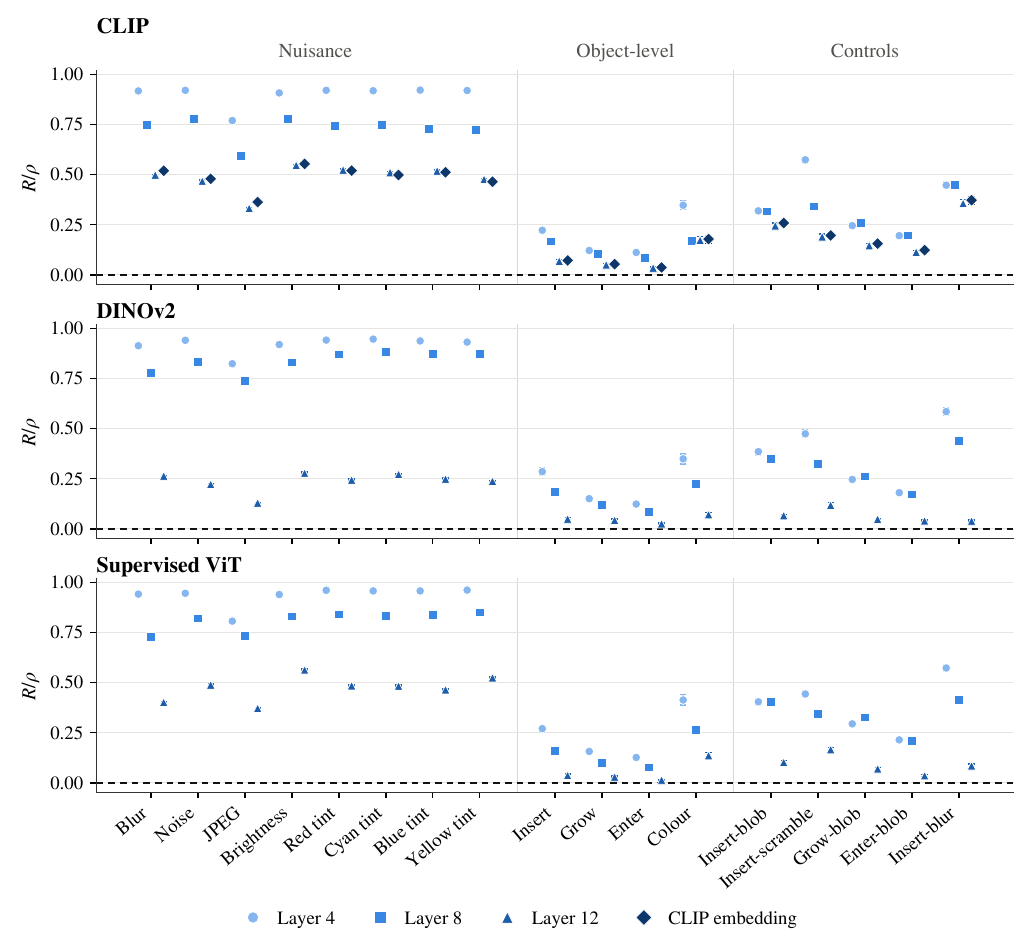}
    \caption{\textbf{Whole-path asymmetry across visual processes and representations.} Sphere-projected $R/\rho$ for all 17 trajectory families and 10 encoder representations, grouped by global appearance transformations, object-level processes, and matched controls. Points are averaged over three training seeds, with pointwise 95\% source-image bootstrap confidence intervals. All intervals lie above zero, indicating lower accumulated forward cost. The magnitude varies substantially across processes and representation depths.}
    \label{fig:h1_path_asymmetry}
\end{figure*}

\paragraph{Effect of sphere-tangent correction.}
To determine whether normalization-induced radial motion accounts for whole-path asymmetry, we recompute $R/\rho$ using sphere-projected displacements and compare the results with the original ambient-chord calculation. Projection changes the sign in none of the 170 combinations, and the maximum absolute change in $R/\rho$ is $0.0306$. Figure~\ref{fig:h1_path_asymmetry} presents the sphere-projected measurements across all trajectory families and encoder representations.

These results show that the detected whole-path asymmetry is not eliminated by removing the known radial artifact of unit-normalized features. They do not imply that sphere projection removes every possible geometric confound, since the true representation manifold may occupy only a subset of the sphere's tangent directions.

\paragraph{Interpretation and scope.}
These findings establish that the tested visual processes contain learnable directional information that generalizes from training images to held-out source images. They empirically motivate an orientation-sensitive geometric description: even when displacement magnitudes are matched, observed motion exhibits reproducible preferences that a symmetric local ruler cannot express.

However, the existence of directional structure does not establish that it is unique to semantic change. Global appearance transformations and matched controls also exhibit strong discrimination and accumulated asymmetry. Nor does the existence alone determine whether the learned field captures genuinely state-dependent preferences beyond a constant drift or common attractor. Those explanations are tested through the sink and residual analyses in Appendix~\ref{app:local_directionality}.

\subsection{Full Per-Family Results}
\label{app:h1_full}

\paragraph{Evaluation across all families.}
Tables~\ref{tab:h1_reverse_full} and~\ref{tab:h1_shuffled_full} provide the complete reverse and shuffled accuracies for all 17 trajectory families across the 10 representations. Each field is evaluated on held-out images from its own process family using sphere-tangent displacements and three independently trained seeds. The tables separate the eight global appearance transformations, four object-level processes, and five matched controls so that their directional behavior can be compared without conflating their experimental roles.

\begin{table*}[!htbp]
    \centering
    \small
    \setlength{\tabcolsep}{3.5pt}
    \begin{tabular}{lrrrrrrrrrr}
        \hline
        & \multicolumn{4}{c}{CLIP}
        & \multicolumn{3}{c}{DINOv2}
        & \multicolumn{3}{c}{Supervised ViT} \\
        Family & L4 & L8 & L12 & Emb.
               & L4 & L8 & L12
               & L4 & L8 & L12 \\
        \hline
        \multicolumn{11}{l}{\textit{Global appearance transformations}} \\
        Blur
        & 100.0 & 100.0 & 100.0 & 99.9
        & 100.0 & 100.0 & 99.2 & 100.0 & 100.0 & 100.0 \\
        Noise
        & 100.0 & 100.0 & 100.0 & 100.0
        & 100.0 & 100.0 & 99.4 & 100.0 & 100.0 & 99.9 \\
        JPEG
        & 100.0 & 100.0 & 99.8 & 99.8
        & 100.0 & 99.7 & 91.5 & 99.9 & 99.8 & 99.4 \\
        Brightness
        & 100.0 & 100.0 & 99.9 & 99.8
        & 100.0 & 99.8 & 84.8 & 100.0 & 100.0 & 99.9 \\
        Red tint
        & 100.0 & 100.0 & 99.9 & 99.4
        & 100.0 & 100.0 & 99.5 & 100.0 & 100.0 & 100.0 \\
        Cyan tint
        & 100.0 & 100.0 & 99.8 & 99.4
        & 100.0 & 100.0 & 99.7 & 100.0 & 100.0 & 99.8 \\
        Blue tint
        & 100.0 & 100.0 & 99.8 & 99.4
        & 100.0 & 100.0 & 99.5 & 100.0 & 100.0 & 99.9 \\
        Yellow tint
        & 100.0 & 100.0 & 99.6 & 99.0
        & 100.0 & 100.0 & 99.4 & 100.0 & 100.0 & 100.0 \\
        \hline
        \multicolumn{11}{l}{\textit{Object-level processes}} \\
        Insert
        & 92.7 & 88.6 & 75.5 & 74.9
        & 90.4 & 88.1 & 73.4 & 91.9 & 82.8 & 70.6 \\
        Grow
        & 85.3 & 82.2 & 75.0 & 72.8
        & 81.6 & 82.5 & 71.0 & 89.1 & 76.8 & 67.9 \\
        Enter
        & 80.9 & 78.2 & 70.9 & 70.4
        & 75.7 & 75.1 & 65.9 & 78.8 & 71.7 & 64.1 \\
        Colour
        & 90.4 & 84.1 & 81.9 & 82.1
        & 88.8 & 85.0 & 74.7 & 86.8 & 85.6 & 85.8 \\
        \hline
        \multicolumn{11}{l}{\textit{Matched controls}} \\
        Insert blob
        & 94.6 & 93.0 & 88.5 & 87.1
        & 95.9 & 93.8 & 77.9 & 94.2 & 90.4 & 79.6 \\
        Insert scramble
        & 99.8 & 96.4 & 89.8 & 88.2
        & 98.1 & 97.4 & 85.4 & 98.1 & 95.9 & 94.4 \\
        Grow blob
        & 96.8 & 92.0 & 79.3 & 78.7
        & 90.8 & 92.3 & 75.2 & 96.9 & 85.8 & 77.8 \\
        Enter blob
        & 88.4 & 86.5 & 78.6 & 77.8
        & 81.9 & 80.6 & 72.5 & 86.6 & 78.3 & 69.3 \\
        Insert blur
        & 98.8 & 94.0 & 86.9 & 86.3
        & 98.7 & 91.3 & 67.6 & 98.6 & 94.9 & 81.7 \\
        \hline
    \end{tabular}
    \caption{\textbf{Per-family sphere-tangent reverse accuracy (\%).} Held-out evaluation averaged over three training seeds. Reverse discrimination compares each observed displacement with its negation at the same state; $50\%$ is the chance reference. All 170 pointwise 95\% confidence intervals lie above chance, with a minimum lower bound of $62.4\%$. Individual intervals are omitted for compactness.}
    \label{tab:h1_reverse_full}
\end{table*}

\begin{table*}[!htbp]
    \centering
    \small
    \setlength{\tabcolsep}{3.5pt}
    \begin{tabular}{lrrrrrrrrrr}
        \hline
        & \multicolumn{4}{c}{CLIP}
        & \multicolumn{3}{c}{DINOv2}
        & \multicolumn{3}{c}{Supervised ViT} \\
        Family & L4 & L8 & L12 & Emb.
               & L4 & L8 & L12
               & L4 & L8 & L12 \\
        \hline
        \multicolumn{11}{l}{\textit{Global appearance transformations}} \\
        Blur
        & 99.7 & 99.6 & 99.2 & 99.0
        & 99.3 & 99.1 & 96.8 & 100.0 & 99.5 & 98.6 \\
        Noise
        & 100.0 & 99.9 & 98.2 & 96.9
        & 100.0 & 99.9 & 96.1 & 100.0 & 100.0 & 97.9 \\
        JPEG
        & 94.4 & 99.8 & 97.5 & 97.3
        & 87.8 & 93.2 & 83.6 & 87.0 & 98.3 & 95.3 \\
        Brightness
        & 99.9 & 99.9 & 98.2 & 97.1
        & 99.8 & 95.1 & 76.0 & 100.0 & 100.0 & 98.6 \\
        Red tint
        & 100.0 & 99.5 & 97.1 & 95.1
        & 99.7 & 99.8 & 95.3 & 100.0 & 99.9 & 98.6 \\
        Cyan tint
        & 100.0 & 99.5 & 97.3 & 95.4
        & 99.9 & 99.7 & 96.0 & 100.0 & 99.9 & 97.5 \\
        Blue tint
        & 100.0 & 99.6 & 96.5 & 94.5
        & 100.0 & 99.9 & 95.3 & 100.0 & 99.9 & 98.4 \\
        Yellow tint
        & 100.0 & 99.5 & 96.4 & 94.1
        & 99.7 & 99.9 & 95.3 & 100.0 & 99.9 & 98.8 \\
        \hline
        \multicolumn{11}{l}{\textit{Object-level processes}} \\
        Insert
        & 70.1 & 71.3 & 68.5 & 68.2
        & 63.1 & 73.3 & 68.7 & 68.5 & 71.0 & 65.3 \\
        Grow
        & 62.1 & 66.5 & 66.0 & 64.6
        & 58.7 & 67.1 & 66.5 & 61.4 & 65.9 & 63.4 \\
        Enter
        & 60.5 & 64.8 & 63.6 & 63.7
        & 56.5 & 64.5 & 62.0 & 59.2 & 60.9 & 60.2 \\
        Colour
        & 74.8 & 69.3 & 73.7 & 73.5
        & 71.6 & 69.5 & 68.5 & 71.7 & 78.2 & 77.6 \\
        \hline
        \multicolumn{11}{l}{\textit{Matched controls}} \\
        Insert blob
        & 69.1 & 76.5 & 71.8 & 70.2
        & 67.0 & 72.9 & 66.2 & 68.7 & 73.9 & 67.5 \\
        Insert scramble
        & 77.1 & 82.3 & 75.4 & 75.1
        & 73.2 & 80.5 & 74.8 & 74.8 & 82.7 & 81.6 \\
        Grow blob
        & 64.2 & 69.6 & 65.6 & 65.4
        & 62.7 & 67.1 & 62.1 & 62.7 & 68.0 & 64.2 \\
        Enter blob
        & 62.0 & 69.2 & 67.6 & 67.1
        & 59.8 & 63.3 & 62.0 & 59.5 & 65.2 & 63.2 \\
        Insert blur
        & 63.4 & 77.1 & 68.8 & 67.6
        & 61.6 & 67.5 & 61.0 & 64.8 & 78.6 & 67.6 \\
        \hline
    \end{tabular}
    \caption{\textbf{Per-family sphere-tangent shuffled accuracy (\%).} Held-out evaluation averaged over three training seeds. Shuffled alternatives are length-matched displacements sampled from other test images of the same process family. All 170 pointwise 95\% confidence intervals lie above the $50\%$ chance reference, with a minimum lower bound of $54.2\%$. Individual intervals are omitted for compactness.}
    \label{tab:h1_shuffled_full}
\end{table*}

\paragraph{Shuffled alternatives provide a more demanding test.}
Shuffled accuracy is no higher than reverse accuracy across the tested combinations, and is strictly lower whenever reverse accuracy is below $100\%$.
For the four object-level processes, reverse accuracy ranges from $64.1\%$ to $92.7\%$, while shuffled accuracy ranges from $56.5\%$ to $78.2\%$. For example, the \textsc{Insert} field at CLIP L4 achieves $92.7\%$ reverse accuracy but $70.1\%$ shuffled accuracy.

This difference is informative. Reverse discrimination asks whether the field prefers an observed direction over its exact opposite, whereas shuffled discrimination asks whether it prefers that direction over a different observed displacement of equal magnitude. The latter provides evidence that directional preferences are associated with the local representation state and process distribution, rather than merely separating forward motion from its negation.

Despite the more demanding comparison, every family--representation combination retains shuffled accuracy above chance, with a minimum pointwise 95\% confidence lower bound of $54.2\%$. Thus, the learned preferences transfer to held-out images even when the competing direction is another observation from the same visual process.

\paragraph{Directional discrimination depends on process and depth.}
Global appearance transformations exhibit particularly strong discrimination, with reverse accuracy ranging from $84.8\%$ to $100.0\%$ and shuffled accuracy from $76.0\%$ to $100.0\%$. Object-level processes generally show weaker discrimination, although their accuracies remain above chance. This difference is consistent with the greater regularity of the controlled image-wide transformations, but does not by itself identify the mechanism responsible for the difference.

For \textsc{Insert}, \textsc{Grow}, and \textsc{Enter}, reverse accuracy at L12 is lower than at L4 in each encoder family. For example, \textsc{Enter} decreases from $80.9\%$ to $70.9\%$ in CLIP, from $75.7\%$ to $65.9\%$ in DINOv2, and from $78.8\%$ to $64.1\%$ in supervised ViT. Shuffled accuracy does not exhibit the same consistent depth trend, indicating that the two discrimination tests respond differently to changes in representation depth.

Matched controls also exhibit reproducible directionality. In several configurations, their accuracies exceed those of the corresponding object-level processes. Consequently, strong directional discrimination should not be interpreted as evidence that a process necessarily depends on meaningful object content. The relationship between process identity, field orientation, and semantic dependence is examined separately in Appendices~\ref{app:process_specificity} and \ref{app:generalization}.

\paragraph{Overall conclusion for the existence evidence.}
The combined evidence supports three conclusions. First, ordered visual processes induce directional preferences that generalize across unseen source images and multiple pretrained representation spaces. Second, these preferences are not explained solely by unequal displacement magnitudes or the known radial artifact of unit normalization. Third, local directional discrimination and whole-path asymmetry provide complementary evidence: the former measures how often the observed orientation is preferred, while the latter measures the accumulated difference in traversal costs.

These findings empirically motivate a non-reversible geometric ruler for visual trajectories. They do not establish semantic specificity or exclude all global convergence mechanisms, which motivates the subsequent attractor, process-specificity, and generalization analyses.



\section{Extended Analysis of Directional Structure Beyond a Common Attractor}
\label{app:local_directionality}

This section extends Section~\ref{sec:local_directionality} by examining whether the directional preferences learned by DiRGeo can be explained by a constant field or convergence toward a common attractor. The distinction matters because a trajectory may have a predictable orientation simply by moving toward a typical final representation, without requiring a more complex state-dependent directional field.

We investigate this explanation through three complementary measurements. First, we compare the complete learned field with simpler constant and attractor-based fields. Second, we remove the learned field's component along a fitted attractor direction and test whether the remaining field is still predictive. Third, we measure the proportion of learned field energy aligned with that direction. We also validate the decomposition on synthetic trajectories where the underlying directional structure is known. Unless stated otherwise, all visual-representation comparisons use held-out source images, sphere-tangent shuffled accuracy, three training seeds, and pointwise 95\% confidence intervals obtained through source-image bootstrap resampling.

\subsection{Reference Fields and Residual Evaluation}
\label{app:h2_protocol}

\paragraph{Constant and common-attractor baselines.}
We first ask whether a simpler directional field can recognize the observed motion as effectively as DiRGeo. A constant field assigns the same preferred direction at every representation state, testing whether a single global drift is sufficient. A common-attractor field instead allows its direction to depend on the current state, but constrains that direction to point toward a single reference destination.

For unit-normalized representations, an attractor $c\in\mathbb{S}^{d-1}$ defines the spherical sink direction
\begin{equation}
    u_{\mathrm{sink}}(z;c)
    =
    \frac{P_zc}{\|P_zc\|_2},
    \qquad
    P_z=I-zz^\top.
    \label{eq:h2_sink_direction}
\end{equation}
When $P_zc=0$, we define the direction to be zero. At other states, this vector points along the sphere toward $c$. A spherical sink therefore provides a simple state-dependent directional model: its preferred direction changes with location, but all states share the same attractor.

We evaluate five common-attractor baselines. The fixed spherical sink points toward the mean terminal representation of the training trajectories. The learned spherical sink optimizes its attractor directly for directional discrimination. The field-fitted spherical sink instead chooses an attractor to approximate the directions of the already learned DiRGeo wind. We additionally evaluate fixed and learned straight-line sinks, which use Euclidean directions toward their attractors rather than the spherical direction construction. All baselines are evaluated with the same sphere-tangent correction and shuffled-discrimination protocol as DiRGeo.

These baselines test different explanations. The fixed terminal sink asks whether movement toward a typical final representation is sufficient. The directly learned sinks test whether an optimized common destination can explain the ordering. The field-fitted sink asks how well the learned directional field itself can be approximated by a single-attractor model. The constant field provides a still simpler state-independent reference.

\paragraph{The fitted sink and the sink-aligned component.}
The field-fitted spherical sink is the primary reference for the decomposition in Section~\ref{sec:characterization}. It determines an attractor $c$, which defines the reference direction $u_{\mathrm{sink}}(z;c)$ at each state. We use this same direction in two distinct ways.

As a standalone baseline, the fitted sink supplies its own directional field and is evaluated independently of the learned wind. For decomposition, however, we project the complete learned tangent field $\beta_\theta^{\mathrm{tan}}(z)$ onto the fitted attractor axis. The resulting sink-aligned component retains the learned field's state-dependent magnitude and signed alignment. Thus, the standalone fitted sink and the projected component $\beta_{\mathrm{sink}}(z)$ share an attractor direction but are not identical vector fields.

\paragraph{Removing the reference sink component.}
Let $u_{\mathrm{sink}}(z)$ denote the direction defined by the fitted attractor. Following Section~\ref{sec:characterization}, the learned tangent field is decomposed as
\begin{equation}
\begin{aligned}
    a(z)
    &=
    \beta_\theta^{\mathrm{tan}}(z)^\top
    u_{\mathrm{sink}}(z),\\
    \beta_{\mathrm{sink}}(z)
    &=
    a(z)u_{\mathrm{sink}}(z),\\
    \delta\beta(z)
    &=
    \beta_\theta^{\mathrm{tan}}(z)
    -
    \beta_{\mathrm{sink}}(z).
\end{aligned}
\label{eq:h2_residual}
\end{equation}
The residual satisfies $\delta\beta(z)^\top u_{\mathrm{sink}}(z)=0$, so it contains no component along the chosen attractor axis. Importantly, the projection removes the entire sink-aligned component, including any variation in its amplitude across representation states. Positive $a(z)$ indicates alignment toward the attractor, while negative values indicate alignment away from it.

We evaluate the residual using
\(
F_{\mathrm{res}}(z,v)=\|v\|_2-\delta\beta(z)^\top v.
\)
No new network is trained: the residual is computed directly from the learned field and its fitted sink direction. Because $\|\delta\beta(z)\|_2\leq\|\beta_\theta^{\mathrm{tan}}(z)\|_2\leq\rho$, the residual also defines a bounded directional ruler.

Residual shuffled accuracy measures whether the remaining field prefers observed motion over length-matched displacements from other test images. Accuracy above $50\%$ indicates predictive directional information outside the reference sink axis. This is a specific geometric statement: the residual cannot be represented solely by a field aligned with the chosen attractor direction. It does not exclude every possible alternative convergence mechanism.

\paragraph{Sink-aligned energy fraction.}
To complement predictive accuracy, we quantify how much of the learned tangent field lies along the reference axis:
\begin{equation}
    Q_{\mathrm{sink}}
    =
    \mathbb{E}_{z}
    \left[
    \frac{
        \bigl(
        \beta_\theta^{\mathrm{tan}}(z)^\top
        u_{\mathrm{sink}}(z)
        \bigr)^2
    }{
        \|\beta_\theta^{\mathrm{tan}}(z)\|_2^2+\epsilon
    }
    \right].
    \label{eq:h2_sink_fraction}
\end{equation}
Here $\epsilon>0$ provides numerical stability, and the expectation is computed over evaluation states. Values near one indicate that most field energy lies along the reference sink axis, whereas values near zero indicate that the learned field is predominantly orthogonal to it.

Because the alignment is squared, $Q_{\mathrm{sink}}$ measures dependence on the attractor axis regardless of orientation. A large value does not necessarily indicate convergence toward the attractor: the signed coefficient $a(z)$ must also be positive. Likewise, the fraction of field energy along an axis does not directly determine its predictive contribution, so we interpret this metric together with residual accuracy.

\paragraph{Paired comparisons and the strongest sink.}
We compare the full learned field against the field-fitted spherical sink through the shuffled-accuracy difference
\(
\Delta_{\mathrm{local}}
=
\mathrm{Acc}_{\mathrm{learned}}
-
\mathrm{Acc}_{\mathrm{fit}},
\)
expressed in percentage points. Positive values indicate better discrimination by the learned wind. Differences are evaluated on the same held-out source images using paired 95\% bootstrap confidence intervals.

The fitted sink is only one common-attractor explanation. To test whether a different attractor construction provides a stronger explanation, we additionally compare against the strongest of the five sink baselines. For each trajectory family and representation, we identify the sink with the highest held-out shuffled accuracy among the five alternatives and compare DiRGeo against that value. This is an evaluation-selected, stringent comparison against the strongest tested alternative, not a baseline chosen independently through validation data. Statistical comparisons are reported using paired, pointwise 95\% confidence intervals.

These two comparisons answer related but distinct questions. The fitted-sink comparison measures whether the learned field outperforms the simple attractor that approximates its own directions. The strongest-sink comparison asks whether it outperforms any of the five examined common-attractor alternatives.

\subsection{Predictive Structure Beyond the Fitted Sink}
\label{app:h2_fitted}

\paragraph{The learned field consistently exceeds its fitted reference.}
Table~\ref{tab:h2_fitted_gap} reports $\Delta_{\mathrm{local}}$ for all 170 family--representation combinations. Every point estimate is positive, and 169/170 paired 95\% confidence intervals lie entirely above zero. The sole unresolved comparison is \textsc{Enter} in DINOv2 L4, where the learned field improves shuffled accuracy by $1.2$ percentage points but the confidence interval $[-0.4,+2.8]$ includes zero.

For global appearance transformations, the gains range from $7.4$ to $45.6$ percentage points. For the four object-level processes, they range from $1.2$ to $27.8$ points. The learned field also significantly outperforms the constant-field baseline in all 170 combinations. Thus, neither a single constant direction nor the spherical sink fitted to approximate the learned wind fully explains the observed directional discrimination.

\begin{table*}[!htbp]
    \centering
    \small
    \setlength{\tabcolsep}{3.5pt}
    \begin{tabular}{lrrrrrrrrrr}
        \hline
        & \multicolumn{4}{c}{CLIP}
        & \multicolumn{3}{c}{DINOv2}
        & \multicolumn{3}{c}{Supervised ViT} \\
        Family & L4 & L8 & L12 & Emb.
               & L4 & L8 & L12 & L4 & L8 & L12 \\
        \hline
        \multicolumn{11}{l}{\textit{Global appearance transformations}} \\
        Blur
        & 32.7 & 23.4 & 9.2 & 7.4 & 30.4 & 33.4 & 31.8 & 37.8 & 28.5 & 28.2 \\
        Noise
        & 18.6 & 31.3 & 37.7 & 30.3 & 24.3 & 21.0 & 40.8 & 24.3 & 13.7 & 38.8 \\
        JPEG
        & 20.9 & 16.2 & 19.6 & 13.5 & 27.1 & 30.2 & 28.9 & 28.6 & 31.1 & 36.0 \\
        Brightness
        & 24.9 & 21.8 & 32.0 & 25.0 & 38.2 & 40.8 & 24.7 & 30.3 & 22.7 & 36.5 \\
        Red tint
        & 23.3 & 32.6 & 40.2 & 33.6 & 37.5 & 39.4 & 44.6 & 29.8 & 30.8 & 44.6 \\
        Cyan tint
        & 22.0 & 29.3 & 40.7 & 33.3 & 36.6 & 37.1 & 45.6 & 28.8 & 31.0 & 43.9 \\
        Blue tint
        & 23.6 & 29.2 & 39.1 & 32.6 & 38.1 & 40.2 & 44.5 & 31.0 & 29.1 & 45.0 \\
        Yellow tint
        & 20.9 & 30.2 & 40.3 & 32.6 & 37.3 & 37.4 & 44.6 & 30.7 & 31.8 & 44.0 \\
        \hline
        \multicolumn{11}{l}{\textit{Object-level processes}} \\
        Insert
        & 16.2 & 18.6 & 18.1 & 17.1 & 6.1 & 20.4 & 19.1 & 12.7 & 20.0 & 13.4 \\
        Grow
        & 9.4 & 13.3 & 13.3 & 12.3 & 4.5 & 13.9 & 17.5 & 5.7 & 16.1 & 12.0 \\
        Enter
        & 4.8 & 6.8 & 8.7 & 7.1 & $1.2^\dagger$ & 8.7 & 12.3 & 2.1 & 8.2 & 3.8 \\
        Colour
        & 21.7 & 16.6 & 21.6 & 22.0 & 20.2 & 15.8 & 17.9 & 21.1 & 22.4 & 27.8 \\
        \hline
        \multicolumn{11}{l}{\textit{Matched controls}} \\
        Insert blob
        & 16.2 & 22.1 & 19.6 & 16.8 & 9.2 & 20.0 & 15.1 & 13.1 & 19.3 & 16.7 \\
        Insert scramble
        & 24.0 & 29.0 & 22.8 & 23.4 & 17.4 & 18.3 & 20.9 & 17.7 & 27.1 & 30.3 \\
        Grow blob
        & 11.0 & 15.9 & 12.3 & 11.4 & 6.1 & 16.0 & 10.9 & 8.2 & 15.0 & 10.8 \\
        Enter blob
        & 9.7 & 13.7 & 13.4 & 10.3 & 4.4 & 12.0 & 8.9 & 4.8 & 8.9 & 8.0 \\
        Insert blur
        & 12.5 & 23.2 & 17.9 & 16.8 & 10.6 & 15.5 & 10.8 & 13.2 & 26.4 & 15.9 \\
        \hline
    \end{tabular}
    \caption{\textbf{Learned field versus the field-fitted spherical sink.} Sphere-tangent shuffled-accuracy differences $\Delta_{\mathrm{local}}$ in percentage points across all 170 configurations, averaged over three training seeds. Positive values favor the learned field. Every paired 95\% confidence interval lies above zero except the entry marked $\dagger$, whose interval is $[-0.4,+2.8]$.}
    \label{tab:h2_fitted_gap}
\end{table*}

\paragraph{The orthogonal residual remains predictive throughout.}
We next test a different question: even when the full field is replaced by its sink-orthogonal residual, does it retain directional information? Table~\ref{tab:h2_residual_full} reports residual shuffled accuracy after removing the fitted sink-aligned component.

The residual exceeds chance in all 170 combinations, with every pointwise 95\% confidence interval above $50\%$. The minimum lower confidence bound is $52.6\%$. Residual accuracy ranges from $81.6\%$ to $99.8\%$ for global appearance transformations and from $54.8\%$ to $78.8\%$ for the four object-level processes. These results establish that the fitted attractor axis does not contain all predictive directional information in the learned field.

Removing a component does not necessarily reduce discrimination accuracy. For example, the \textsc{Colour} residual in supervised ViT L8 achieves $78.8\%$ shuffled accuracy, compared with $78.2\%$ for the complete learned field. This can occur because the removed component may oppose the residual's preferred direction for some observed-versus-shuffled comparisons. Accuracy measures how often a directional preference is correct, not the total magnitude of the vector field.

The residual also retains positive sphere-projected whole-path asymmetry in all 170 combinations, with every pointwise 95\% confidence interval above zero. Its normalized asymmetry $R/\rho$ ranges from $0.011$ to $0.616$. Thus, the residual information is detectable both in local discrimination and in the accumulated costs of forward and reversed trajectories.

\begin{table*}[!htbp]
    \centering
    \small
    \setlength{\tabcolsep}{3.5pt}
    \begin{tabular}{lrrrrrrrrrr}
        \hline
        & \multicolumn{4}{c}{CLIP}
        & \multicolumn{3}{c}{DINOv2}
        & \multicolumn{3}{c}{Supervised ViT} \\
        Family & L4 & L8 & L12 & Emb.
               & L4 & L8 & L12 & L4 & L8 & L12 \\
        \hline
        \multicolumn{11}{l}{\textit{Global appearance transformations}} \\
        Blur
        & 96.1 & 99.0 & 98.5 & 97.6 & 96.8 & 98.3 & 97.2 & 97.0 & 99.2 & 98.8 \\
        Noise
        & 90.0 & 99.5 & 98.8 & 97.2 & 95.8 & 99.7 & 96.6 & 95.9 & 99.8 & 98.7 \\
        JPEG
        & 91.9 & 99.0 & 97.0 & 96.4 & 91.5 & 94.6 & 85.0 & 88.7 & 98.9 & 96.7 \\
        Brightness
        & 95.8 & 99.8 & 98.6 & 97.0 & 98.7 & 97.8 & 81.6 & 97.5 & 99.4 & 98.8 \\
        Red tint
        & 96.5 & 99.2 & 97.8 & 95.4 & 96.6 & 99.7 & 96.3 & 96.5 & 99.7 & 99.0 \\
        Cyan tint
        & 97.8 & 99.2 & 98.3 & 95.8 & 97.4 & 98.7 & 97.1 & 97.2 & 99.6 & 97.9 \\
        Blue tint
        & 96.7 & 99.5 & 97.6 & 95.2 & 98.5 & 99.5 & 96.6 & 97.6 & 99.8 & 98.9 \\
        Yellow tint
        & 97.0 & 99.3 & 97.4 & 94.4 & 97.6 & 99.5 & 96.2 & 95.9 & 99.7 & 99.3 \\
        \hline
        \multicolumn{11}{l}{\textit{Object-level processes}} \\
        Insert
        & 70.1 & 72.2 & 69.4 & 69.4 & 64.9 & 74.8 & 68.8 & 69.9 & 72.1 & 65.4 \\
        Grow
        & 63.6 & 67.4 & 67.0 & 65.2 & 57.8 & 68.5 & 67.4 & 63.1 & 67.7 & 63.5 \\
        Enter
        & 59.7 & 63.8 & 63.0 & 63.3 & 54.8 & 63.3 & 62.2 & 57.6 & 62.6 & 60.0 \\
        Colour
        & 75.9 & 71.7 & 75.6 & 75.4 & 72.3 & 71.1 & 69.0 & 74.8 & 78.8 & 77.7 \\
        \hline
        \multicolumn{11}{l}{\textit{Matched controls}} \\
        Insert blob
        & 70.6 & 81.1 & 78.2 & 76.4 & 72.9 & 79.1 & 68.1 & 72.1 & 78.8 & 68.6 \\
        Insert scramble
        & 81.1 & 85.8 & 79.5 & 79.9 & 78.9 & 81.6 & 77.0 & 78.7 & 84.4 & 84.1 \\
        Grow blob
        & 69.4 & 75.0 & 70.4 & 69.3 & 64.7 & 76.2 & 64.0 & 69.6 & 77.8 & 66.4 \\
        Enter blob
        & 66.5 & 71.6 & 71.1 & 69.5 & 59.3 & 68.6 & 63.1 & 64.4 & 68.6 & 62.2 \\
        Insert blur
        & 75.0 & 84.7 & 73.7 & 72.5 & 74.7 & 75.7 & 60.9 & 73.5 & 81.7 & 68.5 \\
        \hline
    \end{tabular}
    \caption{\textbf{Predictive directional structure after sink removal.} Sphere-tangent shuffled accuracy (\%) of the orthogonal residual $\delta\beta(z)$, evaluated without retraining and averaged over three seeds. All 170 pointwise 95\% confidence intervals lie above the $50\%$ chance reference; the minimum lower bound is $52.6\%$.}
    \label{tab:h2_residual_full}
\end{table*}

\subsection{Dependence on the Choice of Sink}
\label{app:h2_strongest}

\paragraph{Alternative attractors explain some object-level directionality.}
The fitted-sink comparison establishes that a single attractor approximating the learned wind is generally less predictive than the complete field. However, this does not imply that DiRGeo outperforms every possible attractor-based model. A sink trained directly for discrimination, or one pointing toward the mean terminal representation, may perform better than the field-fitted reference.

Table~\ref{tab:h2_strongest_full} therefore compares DiRGeo against the strongest of the five evaluated sinks for each family--representation combination. The entries are shuffled-accuracy differences in percentage points. Unmarked positive entries indicate statistically significant advantages for DiRGeo; $\dagger$ marks comparisons whose paired confidence intervals include zero, and $\ddagger$ marks significant sink advantages.

The learned field significantly outperforms the strongest sink in all 80 global appearance combinations. For the four object-level processes, the corresponding results are 24 significant DiRGeo wins, eight unresolved comparisons, and eight significant sink wins. For matched controls, the counts are 28, 12, and 10, respectively. Overall, DiRGeo significantly outperforms the strongest tested sink in 132/170 combinations.

In all 18 combinations where a sink significantly outperforms DiRGeo, the winning baseline is the spherical sink pointing toward the mean terminal training representation. This provides a concrete interpretation: for some controlled object processes, motion toward a typical final representation is a particularly effective predictor of trajectory order.

\paragraph{Residual information and baseline superiority are different questions.}
The \textsc{Enter} process illustrates why the residual and strongest-sink experiments must be interpreted separately. DiRGeo does not significantly outperform the strongest sink in any of the 10 \textsc{Enter} representations, yet its sink-orthogonal residual remains significantly predictive in all 10.

There is no contradiction. Residual accuracy asks whether the learned wind contains information outside the axis of the field-fitted attractor. The strongest-sink comparison asks whether the complete learned field provides better discrimination than every tested common-attractor alternative. A field may contain additional directional information without using that information more effectively than a competing sink. Consequently, predictive residual structure does not imply universal superiority over attractor-based baselines.

\paragraph{The relative advantage increases at greater depth.}
For the four object-level processes, DiRGeo significantly outperforms every tested sink in only 4/12 L4 combinations, compared with 20/28 combinations at the remaining representation levels. This indicates that simple common convergence is especially competitive in shallow object-level representations, whereas learned directional fields more frequently offer additional predictive benefit at greater depth.

This pattern is consistent with the decreasing sink-aligned energy fractions analyzed below. Nevertheless, it does not establish that every deeper representation favors the learned field, nor does it rule out attractor models outside the five tested constructions.

\begin{table*}[!htbp]
    \centering
    \small
    \setlength{\tabcolsep}{3pt}
    \begin{tabular}{lrrrrrrrrrr}
        \hline
        & \multicolumn{4}{c}{CLIP}
        & \multicolumn{3}{c}{DINOv2}
        & \multicolumn{3}{c}{Supervised ViT} \\
        Family & L4 & L8 & L12 & Emb.
               & L4 & L8 & L12 & L4 & L8 & L12 \\
        \hline
        \multicolumn{11}{l}{\textit{Global appearance transformations}} \\
        Blur
        & 21.0 & 11.9 & 6.7 & 6.2 & 23.0 & 29.0 & 31.1 & 24.8 & 13.4 & 23.8 \\
        Noise
        & 9.1 & 12.6 & 25.9 & 19.2 & 15.7 & 9.4 & 37.7 & 10.9 & 4.8 & 35.3 \\
        JPEG
        & 20.9 & 5.0 & 11.9 & 9.3 & 16.2 & 26.3 & 27.6 & 21.6 & 25.5 & 31.3 \\
        Brightness
        & 24.9 & 18.9 & 28.3 & 21.8 & 38.2 & 37.4 & 24.6 & 30.3 & 21.1 & 36.5 \\
        Red tint
        & 10.7 & 14.9 & 28.9 & 21.8 & 19.1 & 16.3 & 42.2 & 16.6 & 10.3 & 38.2 \\
        Cyan tint
        & 9.2 & 16.0 & 30.5 & 22.8 & 20.4 & 15.2 & 42.6 & 17.3 & 12.8 & 39.3 \\
        Blue tint
        & 12.4 & 13.9 & 27.8 & 20.5 & 24.1 & 19.4 & 41.9 & 17.7 & 10.9 & 39.4 \\
        Yellow tint
        & 8.5 & 15.5 & 29.3 & 21.7 & 19.9 & 17.4 & 42.4 & 17.7 & 15.1 & 39.1 \\
        \hline
        \multicolumn{11}{l}{\textit{Object-level processes}} \\
        Insert
        & 7.8 & 10.5 & 8.0 & 9.6
        & $-4.7^\ddagger$ & 14.8 & 11.9
        & $2.6^\dagger$ & 15.0 & 9.1 \\
        Grow
        & $-2.4^\dagger$ & 3.9 & $2.9^\dagger$ & 3.8
        & $-7.8^\ddagger$ & 8.0 & 9.7
        & $-5.8^\ddagger$ & 8.6 & 6.1 \\
        Enter
        & $-3.8^\ddagger$ & $-2.5^\ddagger$
        & $-2.9^\ddagger$ & $-2.5^\dagger$
        & $-9.6^\ddagger$ & $1.9^\dagger$ & $2.1^\dagger$
        & $-8.0^\ddagger$ & $-2.6^\dagger$ & $-2.1^\dagger$ \\
        Colour
        & 6.0 & 5.7 & 12.7 & 13.4 & 6.9 & 5.1 & 14.1 & 9.0 & 9.2 & 21.5 \\
        \hline
        \multicolumn{11}{l}{\textit{Matched controls}} \\
        Insert blob
        & $0.6^\dagger$ & 6.8 & 10.7 & 7.7
        & $-4.3^\ddagger$ & 6.8 & 12.0
        & $-4.9^\ddagger$ & 4.4 & 12.2 \\
        Insert scramble
        & 7.5 & 15.6 & 11.6 & 11.3
        & $0.6^\dagger$ & 8.4 & 19.5
        & $0.8^\dagger$ & 14.0 & 24.6 \\
        Grow blob
        & $-2.6^\dagger$ & $-1.4^\dagger$ & $0.3^\dagger$
        & $-3.4^\ddagger$ & $-5.1^\ddagger$ & 5.3 & 7.6
        & $-10.1^\ddagger$ & $-3.2^\ddagger$ & 8.5 \\
        Enter blob
        & $-4.8^\ddagger$ & $-0.2^\dagger$ & $1.4^\dagger$
        & $-0.7^\dagger$ & $-9.9^\ddagger$ & $-0.2^\dagger$
        & 4.8 & $-10.8^\ddagger$ & $-5.0^\ddagger$ & 4.5 \\
        Insert blur
        & 4.8 & 10.6 & 12.7 & 8.7
        & $-3.1^\dagger$ & 11.8 & 7.1
        & $1.8^\dagger$ & 13.8 & 14.5 \\
        \hline
    \end{tabular}
    \caption{\textbf{DiRGeo versus the strongest of five common-attractor baselines.} Entries report shuffled-accuracy differences in percentage points, with the strongest sink selected by maximum held-out accuracy for each family--representation combination. Unmarked entries significantly favor DiRGeo; $\dagger$ indicates a paired 95\% confidence interval containing zero, and $\ddagger$ indicates a significant advantage for a sink.}
    \label{tab:h2_strongest_full}
\end{table*}

\subsection{Sink Alignment Across Representation Depth}
\label{app:h2_energy}

\paragraph{The learned field becomes less aligned with the fitted attractor at depth.}
The accuracy experiments establish that the learned field contains predictive information beyond the fitted sink. We now examine how much of the field's magnitude is associated with that attractor direction, independently of its classification accuracy.

Table~\ref{tab:h2_sink_energy} reports $Q_{\mathrm{sink}}$, the mean fraction of tangent-field energy aligned with the reference sink axis. For every trajectory family and encoder, this fraction is smaller in the deepest evaluated representation than at L4.

The decrease is particularly pronounced for object-level processes. In CLIP, their sink-aligned fractions range from $31$--$47\%$ at L4 and fall to $5$--$14\%$ in the projected image embedding. For DINOv2, the corresponding ranges decrease from $38$--$66\%$ at L4 to $1$--$3\%$ at L12. For supervised ViT, they decrease from $40$--$61\%$ to $1$--$6\%$. These values show that deeper learned fields are increasingly dominated, in squared magnitude, by directions orthogonal to the chosen common-attractor axis.

Importantly, this result describes angular organization, not necessarily greater directional discrimination. The object-level reverse accuracies reported in Appendix~\ref{app:directional_existence} often decrease with depth, even as the sink-aligned energy fraction decreases. A field can therefore become less aligned with a common attractor without becoming more accurate at recognizing observed motion.

\paragraph{Energy alignment and predictive contribution are complementary.}
A small sink-aligned component can still affect many binary directional comparisons, particularly when the learned and residual preferences are close. Conversely, a large orthogonal component may contribute little to prediction if it is poorly aligned with the observed steps. We therefore interpret $Q_{\mathrm{sink}}$ alongside the residual shuffled accuracy and baseline comparisons, rather than treating field energy alone as evidence of predictive importance.

The decreasing alignment at depth is also consistent with the process-field comparisons in Appendix~\ref{app:process_specificity}, where separately trained object-level fields become less similar in deeper representations. Together, the measurements suggest that deeper representations support more differentiated directional fields, with less of their energy explained by a shared attractor axis.

\begin{table*}[!htbp]
    \centering
    \small
    \setlength{\tabcolsep}{5pt}
    \begin{tabular}{lrrrrrrrrrr}
        \hline
        & \multicolumn{4}{c}{CLIP}
        & \multicolumn{3}{c}{DINOv2}
        & \multicolumn{3}{c}{Supervised ViT} \\
        Family & L4 & L8 & L12 & Emb.
               & L4 & L8 & L12 & L4 & L8 & L12 \\
        \hline
        \multicolumn{11}{l}{\textit{Global appearance transformations}} \\
        Blur        & 42 & 36 & 29 & 37 & 45 & 29 & 9  & 50 & 35 & 17 \\
        Noise       & 63 & 40 & 22 & 27 & 68 & 46 & 7  & 63 & 42 & 20 \\
        JPEG        & 38 & 27 & 12 & 16 & 62 & 33 & 8  & 52 & 27 & 16 \\
        Brightness  & 46 & 35 & 25 & 29 & 44 & 34 & 12 & 52 & 31 & 13 \\
        Red tint    & 55 & 40 & 23 & 24 & 62 & 47 & 10 & 68 & 28 & 14 \\
        Cyan tint   & 52 & 39 & 23 & 24 & 60 & 44 & 11 & 64 & 27 & 13 \\
        Blue tint   & 53 & 39 & 25 & 27 & 57 & 45 & 11 & 63 & 27 & 14 \\
        Yellow tint & 54 & 38 & 22 & 23 & 61 & 45 & 9  & 66 & 31 & 17 \\
        \hline
        \multicolumn{11}{l}{\textit{Object-level processes}} \\
        Insert & 31 & 19 & 4  & 5  & 54 & 16 & 1 & 43 & 10 & 1 \\
        Grow   & 40 & 25 & 5  & 6  & 61 & 21 & 2 & 53 & 15 & 2 \\
        Enter  & 47 & 25 & 4  & 6  & 66 & 26 & 1 & 61 & 17 & 2 \\
        Colour & 33 & 19 & 13 & 14 & 38 & 24 & 3 & 40 & 16 & 6 \\
        \hline
        \multicolumn{11}{l}{\textit{Matched controls}} \\
        Insert blob     & 47 & 31 & 22 & 26 & 63 & 39 & 8  & 61 & 31 & 8 \\
        Insert scramble & 59 & 26 & 15 & 16 & 62 & 34 & 7  & 58 & 25 & 10 \\
        Grow blob       & 61 & 44 & 30 & 32 & 68 & 50 & 11 & 72 & 47 & 12 \\
        Enter blob      & 57 & 31 & 26 & 27 & 67 & 43 & 16 & 70 & 34 & 13 \\
        Insert blur     & 71 & 37 & 24 & 26 & 79 & 49 & 4  & 74 & 27 & 6 \\
        \hline
    \end{tabular}
    \caption{\textbf{Sink-aligned energy across representations.} Mean $Q_{\mathrm{sink}}$ (\%) over held-out states, measuring the squared projection of $\beta_\theta^{\mathrm{tan}}(z)$ onto the fitted sink axis relative to total field energy. The quantity is unsigned: it measures axis alignment rather than convergence toward the attractor. L4, L8, and L12 denote transformer-block features; Emb. is the projected CLIP image embedding.}
    \label{tab:h2_sink_energy}
\end{table*}

\subsection{Synthetic Validation of the Decomposition}
\label{app:h2_synthetic}

\paragraph{Why synthetic validation is needed.}
The visual-representation experiments show that the residual remains predictive after removing a fitted sink direction. However, the true generative geometry of these representations is unknown, so we cannot directly determine whether this behavior corresponds to an additional directional mechanism or to an imperfectly specified attractor model.

We therefore construct synthetic trajectories with known directional structure. The purpose is not to reproduce natural images, but to test whether the proposed decomposition behaves correctly when convergence is the only directional mechanism and when an additional state-dependent field is present.

\paragraph{Controlled convergence and local-field conditions.}
The synthetic experiment uses 3,000 trajectories of 10 frames on a 64-dimensional unit sphere, with an 80/20 training--test split. Under the pure-convergence condition, each trajectory moves toward a common reference center with added noise. The directional preference is therefore generated by a single spherical attractor.

Under the convergence-plus-local-field condition, we introduce an additional state-dependent rotation field. The trajectories retain an attractor component, but their motion also follows directional structure that cannot be represented solely by pointing toward the common center. This creates two conditions with a known difference: one contains only convergence, while the other contains convergence and an additional local directional preference.

We train and evaluate the learned MLP field and the corresponding sink baselines under both conditions. We then remove different reference sink directions and evaluate the remaining residuals using sphere-tangent reverse and shuffled accuracy. An appropriate decomposition should produce chance-level residual discrimination under pure convergence while retaining predictive information when the additional local field is present.

\paragraph{The spherical residual distinguishes the two conditions.}
Table~\ref{tab:h2_synthetic} reports the results. Under pure convergence, the learned MLP reaches $100.0\%$ reverse accuracy and $60.3\%$ shuffled accuracy, while the fitted spherical sink reaches $100.0\%$ and $62.3\%$, respectively. Thus, the original trajectories contain directional information that both learned and attractor-based fields can detect.

After removing the fitted spherical sink direction, residual reverse and shuffled accuracies fall to $49.8\%$ and $50.0\%$, respectively. The learned-spherical-sink residual gives the same result. This is the expected behavior under the pure-convergence condition: once the known attractor direction is removed, no meaningful directional preference remains.

In contrast, when the additional rotation field is present, the residual retains $100.0\%$ reverse and shuffled accuracy after spherical sink removal. The decomposition therefore preserves the additional directional signal rather than eliminating all motion preferences indiscriminately.

\paragraph{Why the reference geometry matters.}
The experiment also exposes a potential failure mode. Under pure convergence, removing a straight-line sink leaves $79.4\%$ reverse accuracy and $52.6\%$ shuffled accuracy. The residual appears directional even though the synthetic process contains no additional local-field mechanism.

This occurs because a straight-line reference does not match the spherical geometry of the generated trajectories. An incorrectly specified reference direction can leave part of the convergence signal in the residual, producing a misleading indication of additional structure.

The synthetic results therefore support the use of a spherical sink for analyzing normalized visual representations. They validate the behavior of the decomposition under these controlled conditions, but do not prove that the fitted spherical sink captures every convergence mechanism in real pretrained representations. The interpretation of the visual experiments remains relative to the chosen reference attractor.

\begin{table*}[!htbp]
    \centering
    \small
    \setlength{\tabcolsep}{7pt}
    \begin{tabular}{lrrrr}
        \hline
        & \multicolumn{2}{c}{Pure convergence}
        & \multicolumn{2}{c}{Convergence + local field} \\
        Field & Reverse & Shuffled & Reverse & Shuffled \\
        \hline
        Learned MLP field
        & 100.0 & 60.3 & 100.0 & 100.0 \\
        Fitted spherical sink
        & 100.0 & 62.3 & 100.0 & 59.4 \\
        Learned straight-line sink
        & 100.0 & 62.3 & 100.0 & 59.4 \\
        Center-of-sphere sink
        & 50.0 & 50.0 & 50.0 & 50.0 \\
        \hline
        Residual: fitted spherical sink removed
        & 49.8 & 50.0 & 100.0 & 100.0 \\
        Residual: learned spherical sink removed
        & 49.8 & 50.0 & 100.0 & 100.0 \\
        Residual: straight-line sink removed
        & 79.4 & 52.6 & 100.0 & 100.0 \\
        \hline
    \end{tabular}
    \caption{\textbf{Synthetic validation of sink--residual decomposition.} Sphere-tangent reverse and shuffled accuracy (\%), averaged over three training seeds, under known pure-convergence and convergence-plus-local-field conditions. Residual rows evaluate orthogonal components after removing the specified reference direction. Spherical sink removal reduces residual discrimination to chance under pure convergence while preserving the additional directional signal; straight-line sink removal leaves a spurious residual preference.}
    \label{tab:h2_synthetic}
\end{table*}

\paragraph{Overall interpretation.}
The visual and synthetic experiments provide complementary evidence. The full learned field consistently outperforms its field-fitted spherical sink, and its orthogonal residual remains predictive across every tested family--representation combination. At the same time, comparisons against alternative sinks show that a common terminal attractor can be highly competitive, particularly for object-level processes in shallow representations.

The decreasing sink-aligned energy fractions further indicate that deeper representations contain directional fields whose orientations are less dominated by the selected common-attractor axis. Synthetic controls support the decomposition's ability to separate convergence from an added local field when the reference geometry is correctly specified.

Together, these findings establish directional information beyond the tested common-attractor explanation, without requiring the stronger claim that all global convergence mechanisms have been excluded. This motivates the following analyses of how the remaining directional structure differs across visual processes.



\section{Extended Analysis of Process-Dependent Directional Geometry}
\label{app:process_specificity}
\label{app:process_directionality}

This section extends Section~\ref{sec:process_specificity} by examining two complementary properties of the learned directional geometry. First, we ask whether different visual processes induce different local directional fields, even when the fields are evaluated at identical representation states. Second, we ask how these preferences accumulate along actual trajectories, including controlled image transformations and natural-video events.

These questions are related but require different measurements. Field cosine similarity compares the orientations preferred by two separately trained fields at the same representation state. Whole-path asymmetry $R/\rho$ measures the accumulated difference between forward and reverse traversal costs along a particular trajectory. A pair of fields can become less aligned without necessarily producing stronger path asymmetry, because field orientation and accumulated cost magnitude are distinct geometric properties.

We first analyze process-trained fields and their sink-removed residuals, then compare whole-path asymmetry across controlled transformations and natural videos. All photo-based results use frozen encoder representations and fields trained on ordered image trajectories. Unless stated otherwise, results are averaged over three training seeds, and confidence intervals are estimated by resampling source images or, for video experiments, whole recordings.

\subsection{Two Complementary Measures of Process Dependence}
\label{app:h3_protocol}

\paragraph{Comparing directional fields at identical states.}
For process families $p$ and $q$, let $\beta_{\theta,p}^{\mathrm{tan}}(z)$ and $\beta_{\theta,q}^{\mathrm{tan}}(z)$ denote their separately trained tangent fields within the same encoder representation. We compare their preferred directions through
\begin{equation}
    C_{p,q}
    =
    \mathbb{E}_{z}
    \left[
        \cos\left(
            \beta_{\theta,p}^{\mathrm{tan}}(z),
            \beta_{\theta,q}^{\mathrm{tan}}(z)
        \right)
    \right].
    \label{eq:h3_field_similarity}
\end{equation}
Both fields are evaluated at the same representation states from the object-level evaluation dataset. This is important because two processes may naturally visit different parts of representation space. Comparing fields along their separate trajectories would confound differences in preferred directions with differences in the states visited. Evaluating both fields at identical states instead directly measures whether the processes assign different directional preferences to the same representation.

Cosine similarity near one indicates aligned preferences, values near zero indicate approximately orthogonal directions, and negative values indicate opposing directions. This metric characterizes field orientation rather than magnitude. It is also distinct from wind--velocity alignment: $C_{p,q}$ compares two learned fields with each other, while wind--velocity alignment compares a field with an observed displacement.

We repeat the comparisons after removing each field's component along its own fitted spherical sink direction. Writing $\delta\beta_p(z)$ and $\delta\beta_q(z)$ for the corresponding residual fields, we compute their mean cosine similarity at the same states. This tests whether process differences remain after excluding the chosen common-attractor components.

All vector comparisons are performed within a single encoder representation space. We compare scalar similarity measurements across layers and encoders, but do not directly compare vectors belonging to different representation spaces.

\paragraph{Measuring accumulated path asymmetry.}
Field similarity does not reveal how strongly a process favors forward traversal along its observed trajectory. For this purpose, we evaluate the normalized whole-path asymmetry defined in Section~\ref{sec:characterization}:
\begin{equation}
    \frac{R(\gamma)}{\rho}
    =
    \frac{
        \widehat{L}_F(\bar{\gamma})
        -
        \widehat{L}_F(\gamma)
    }{
        \rho\left(
            \widehat{L}_F(\bar{\gamma})
            +
            \widehat{L}_F(\gamma)
        \right)
    }.
    \label{eq:h4_path_asymmetry}
\end{equation}
Here $\gamma$ denotes an observed representation trajectory, $\bar{\gamma}$ denotes the same trajectory traversed backward, and $\rho=0.9$ bounds the directional field. The primary measurement uses sphere-tangent displacements, with the field evaluated at the departure state of each forward or backward step.

Positive values indicate lower forward cost, negative values indicate lower backward cost, and zero indicates equal accumulated costs. The magnitude measures relative cost asymmetry, not directional classification accuracy. A field may consistently favor forward motion in local comparisons while yielding modest accumulated asymmetry because its directional contributions are small relative to the total path length or partially cancel along the path.

For controlled image trajectories, we evaluate the field trained for the corresponding transformation family. For natural videos, we apply fields trained on controlled photo trajectories without adaptation. This provides a test of whether learned process preferences transfer to new, observed visual events.

\subsection{Process-Specific Directional Fields}
\label{app:h3_full_fields}

\paragraph{Initially aligned process fields become more distinct with depth.}
Table~\ref{tab:h3_full_similarity} reports pairwise cosine similarities between separately trained fields. At L4, the \textsc{Insert}, \textsc{Grow}, and \textsc{Enter} fields have substantial mutual alignment, with similarities ranging from $0.65$ to $0.88$. For all three process pairs, similarity decreases from L4 to L8 to L12 within each encoder family.

At L12, pairwise similarities among these processes range from $0.25$ to $0.29$ in CLIP, $0.16$ to $0.18$ in DINOv2, and $0.15$ to $0.18$ in supervised ViT. For example, \textsc{Insert}--\textsc{Enter} similarity decreases from $0.66$ to $0.25$ in CLIP, from $0.81$ to $0.16$ in DINOv2, and from $0.74$ to $0.15$ in supervised ViT.

These measurements indicate that directions preferred by the three processes are relatively similar in shallow representations but increasingly differentiated in deeper representations. Because all fields are evaluated at identical states, the observed changes cannot be attributed merely to the processes visiting different regions of representation space.

CLIP's projected image embedding also exhibits greater differentiation than L4, although its similarities ($0.30$--$0.34$) are slightly higher than those at L12. Thus, the consistent decrease holds across the sampled transformer blocks without requiring the final projection to produce further separation.

\paragraph{Some process distinctions appear even in shallow representations.}
The depth trend does not imply that every process pair begins with a shared directional preference. The \textsc{Insert}--\textsc{Colour} similarity remains between $-0.04$ and $0.01$ across all evaluated representations. These fields are therefore nearly orthogonal even at L4.

The increasing differentiation with depth is most evident among \textsc{Insert}, \textsc{Grow}, and \textsc{Enter}, whose shallow fields are initially aligned. This distinction is important: representation depth appears to separate some previously overlapping process preferences rather than universally creating differentiation among all processes.

\begin{table*}[!htbp]
    \centering
    \small
    \setlength{\tabcolsep}{4pt}
    \begin{tabular}{lrrrrrrr}
        \hline
        Representation
        & Insert--Enter
        & Insert--Grow
        & Grow--Enter
        & Insert--Colour
        & Insert--Blob
        & Grow--Blob
        & Enter--Blob \\
        \hline
        CLIP L4
        & 0.66 & 0.65 & 0.78 & -0.03 & 0.29 & 0.29 & 0.30 \\
        CLIP L8
        & 0.48 & 0.54 & 0.59 & -0.04 & 0.20 & 0.23 & 0.22 \\
        CLIP L12
        & 0.25 & 0.26 & 0.29 & -0.03 & 0.09 & 0.03 & 0.03 \\
        CLIP embedding
        & 0.30 & 0.32 & 0.34 & -0.04 & 0.09 & 0.03 & 0.05 \\
        \hline
        DINOv2 L4
        & 0.81 & 0.82 & 0.88 & 0.01 & 0.26 & 0.18 & 0.26 \\
        DINOv2 L8
        & 0.55 & 0.56 & 0.64 & -0.04 & 0.24 & 0.26 & 0.25 \\
        DINOv2 L12
        & 0.16 & 0.18 & 0.18 & -0.04 & 0.08 & 0.07 & 0.05 \\
        \hline
        Supervised ViT L4
        & 0.74 & 0.75 & 0.85 & -0.01 & 0.33 & 0.34 & 0.37 \\
        Supervised ViT L8
        & 0.42 & 0.44 & 0.55 & -0.02 & 0.17 & 0.20 & 0.20 \\
        Supervised ViT L12
        & 0.15 & 0.18 & 0.17 & -0.03 & 0.09 & 0.09 & 0.09 \\
        \hline
    \end{tabular}
    \caption{\textbf{Similarity between process-trained directional fields.} Mean cosine similarity between full tangent fields evaluated at identical representation states. Insert--Blob, Grow--Blob, and Enter--Blob compare each object-level process with its matched blob-control field. Values near one indicate aligned directions, while values near zero indicate approximately orthogonal preferences. L4, L8, and L12 denote transformer-block features.}
    \label{tab:h3_full_similarity}
\end{table*}

\subsection{Directional Fields After Sink Removal}
\label{app:h3_residual_fields}

\paragraph{Process differentiation remains beyond fitted attractor directions.}
We next examine whether the process differences reported above can be explained by the fields sharing or differing in their common-attractor components. Following Appendix~\ref{app:local_directionality}, we remove from each field its projection along the corresponding fitted sink axis and compare the resulting residuals at identical representation states.

Table~\ref{tab:h3_residual_similarity} reports these comparisons. Among \textsc{Insert}, \textsc{Grow}, and \textsc{Enter}, residual similarities range from $0.48$ to $0.69$ at L4 and from $0.15$ to $0.26$ at L12. Each process pair again becomes less aligned from L4 to L8 to L12 across all three encoders.

For \textsc{Insert}--\textsc{Enter}, residual similarity decreases from $0.48$ to $0.23$ in CLIP, from $0.57$ to $0.16$ in DINOv2, and from $0.53$ to $0.15$ in supervised ViT. These trends parallel the full-field comparisons, demonstrating that process differentiation is not confined to the components aligned with the fitted attractors.

\paragraph{The contribution of the sink component changes with depth.}
At L4, removing the sink-aligned components generally reduces similarity among \textsc{Insert}, \textsc{Grow}, and \textsc{Enter}. This is consistent with some of their shallow field alignment arising from common convergence directions.

At L12, the full-field and residual similarities become closer. This agrees with the sink-energy analysis in Appendix~\ref{app:h2_energy}, where deeper representations have smaller fractions of field energy aligned with the fitted attractor. Thus, the deeper process differences persist largely outside the selected sink axes.

The residual \textsc{Insert}--\textsc{Colour} similarity remains near zero, ranging from $-0.09$ to $-0.03$. As in the full-field analysis, this pair is already differentiated in shallow representations rather than becoming distinct only at depth.

\begin{table*}[!htbp]
    \centering
    \small
    \setlength{\tabcolsep}{4pt}
    \begin{tabular}{lrrrrrrr}
        \hline
        Representation
        & Insert--Enter
        & Insert--Grow
        & Grow--Enter
        & Insert--Colour
        & Insert--Blob
        & Grow--Blob
        & Enter--Blob \\
        \hline
        CLIP L4
        & 0.48 & 0.50 & 0.63 & -0.05 & 0.27 & 0.30 & 0.28 \\
        CLIP L8
        & 0.37 & 0.44 & 0.48 & -0.04 & 0.18 & 0.23 & 0.22 \\
        CLIP L12
        & 0.23 & 0.24 & 0.26 & -0.03 & 0.10 & 0.07 & 0.07 \\
        CLIP embedding
        & 0.28 & 0.29 & 0.31 & -0.03 & 0.11 & 0.07 & 0.08 \\
        \hline
        DINOv2 L4
        & 0.57 & 0.60 & 0.68 & -0.09 & 0.31 & 0.27 & 0.27 \\
        DINOv2 L8
        & 0.45 & 0.48 & 0.55 & -0.08 & 0.20 & 0.24 & 0.23 \\
        DINOv2 L12
        & 0.16 & 0.18 & 0.17 & -0.04 & 0.08 & 0.07 & 0.06 \\
        \hline
        Supervised ViT L4
        & 0.53 & 0.56 & 0.69 & -0.05 & 0.28 & 0.28 & 0.29 \\
        Supervised ViT L8
        & 0.36 & 0.38 & 0.48 & -0.03 & 0.15 & 0.18 & 0.17 \\
        Supervised ViT L12
        & 0.15 & 0.18 & 0.16 & -0.03 & 0.08 & 0.08 & 0.09 \\
        \hline
    \end{tabular}
    \caption{\textbf{Process-field similarity after sink removal.} Mean cosine similarity between residual fields $\delta\beta_p(z)$ and $\delta\beta_q(z)$ evaluated at identical states. Each residual is obtained by removing its own fitted sink-aligned component. Process pairs and evaluation states match Table~\ref{tab:h3_full_similarity}. Increasing angular differentiation with depth persists after sink removal.}
    \label{tab:h3_residual_similarity}
\end{table*}

\subsection{Matched Controls and Field Interpretation}
\label{app:h3_controls}

\paragraph{Strong directionality does not imply identical fields.}
The matched blob controls provide another comparison. These controls retain the associated transformation construction while replacing meaningful object content, allowing us to examine whether the learned field depends on the manipulated visual content.

The full-field similarities between \textsc{Insert}, \textsc{Grow}, and \textsc{Enter} and their corresponding blob controls range from $0.18$ to $0.37$ at L4, but only $0.03$--$0.09$ in the deepest representations. Residual-field comparisons exhibit a similar pattern, with similarities of $0.27$--$0.31$ at L4 and $0.06$--$0.11$ at depth.

These fields therefore become increasingly differentiated even when the paired processes retain similar transformation constructions. Importantly, the blob controls themselves exhibit strong directional discrimination. Their separation from the corresponding real-object fields cannot simply be interpreted as one process having directional geometry and the other lacking it.

\paragraph{Field orientation and directional strength are distinct properties.}
Two fields may both predict their respective process directions accurately while preferring different orientations at a common representation state. Conversely, two fields can be closely aligned while differing in directional magnitude or discrimination accuracy.

The cosine comparisons establish angular differentiation among the tested fields. They do not directly establish that one field is more predictive, nor that the differences are exclusively semantic. The relationship between learned geometry and object categories is examined separately in Appendix~\ref{app:generalization}.

Figure~\ref{fig:h3_depth} summarizes the full-field and residual comparisons for the three process pairs whose initially aligned fields become increasingly distinct across representation depth.

\begin{figure}[!htbp]
    \centering
    \includegraphics[width=\columnwidth]{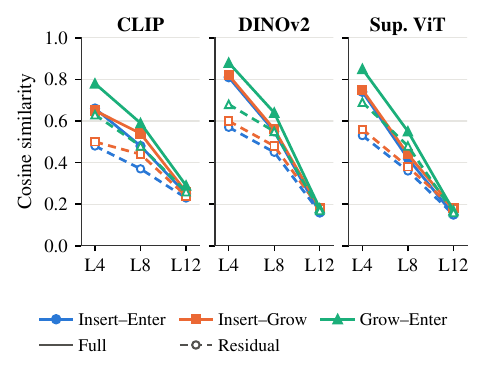}
    \caption{\textbf{Process-specific fields become less aligned with depth.} Mean cosine similarities among \textsc{Insert}, \textsc{Grow}, and \textsc{Enter} fields across L4, L8, and L12 in CLIP, DINOv2, and supervised ViT. Solid lines show full fields; dashed lines show sink-removed residuals. All comparisons use identical evaluation states and average three seeds. Both full and residual field similarities decrease with depth.}
    \label{fig:h3_depth}
\end{figure}

\subsection{Accumulated Asymmetry Across Controlled Visual Processes}
\label{app:h4_image_processes}

\paragraph{Why path asymmetry is a separate test.}
The preceding analyses compare which directions different visual processes prefer at a fixed representation state. We now examine what happens when these preferences are accumulated along the trajectories generated by those processes.

For each controlled family, we evaluate its own learned field on held-out image trajectories and compute the sphere-projected whole-path asymmetry $R/\rho$. As described in Section~\ref{sec:characterization}, forward and reverse traversal use their respective departure states. This is necessary because reversing an entire trajectory changes not only displacement orientation but also where the field is evaluated.

\paragraph{Global appearance transformations exhibit stronger accumulated asymmetry.}
Table~\ref{tab:h4_photo_paths} reports sphere-projected $R/\rho$ for all 170 family--representation combinations. Within each representation, every global appearance transformation has greater accumulated asymmetry than every object-level process.

Across all representations, global appearance values range from $0.129$ to $0.962$, whereas the four object-level processes range from $0.014$ to $0.413$. The pooled ranges overlap because they combine different representations, but the within-representation ordering is consistent.

The greater asymmetry of global appearance transformations is compatible with their strong local directional discrimination. However, these metrics are not interchangeable. Local accuracy measures the proportion of comparisons whose preferred orientation is correct, while $R/\rho$ measures the relative accumulated difference in traversal costs.

\paragraph{Object-level processes also exhibit distinct path behavior.}
Even among controlled object-level processes, accumulated asymmetry differs. For example, \textsc{Insert} has greater $R/\rho$ than \textsc{Enter} in all 10 representations, although the two trajectory constructions share identical endpoints.

The paired confidence interval for their asymmetry difference lies above zero in every representation under both the original ambient-chord calculation and the sphere-projected recalculation. This demonstrates that the two process-trained fields assign different relative asymmetries to their corresponding trajectories.

This comparison should not be confused with a reciprocal path-preference test. Here, each process is evaluated under its own learned field. Whether a field distinguishes alternative journeys between identical endpoints is examined separately in Appendix~\ref{app:same_endpoints}.

\paragraph{Field differentiation and path-asymmetry magnitude have different depth trends.}
For every tested trajectory family and encoder, $R/\rho$ is lower in the deepest representation than at L4. This contrasts with the increasing angular differentiation of process-trained fields reported in the preceding subsections.

The two findings are compatible. A deeper representation can contain more distinct directional fields for different visual processes while exhibiting smaller relative forward--reverse costs along each process trajectory. Thus, process specificity concerns differences between fields, while path asymmetry concerns accumulated directional effects within a trajectory.

\begin{table*}[!htbp]
    \centering
    \small
    \setlength{\tabcolsep}{3pt}
    \begin{tabular}{lrrrrrrrrrr}
        \hline
        & \multicolumn{4}{c}{CLIP}
        & \multicolumn{3}{c}{DINOv2}
        & \multicolumn{3}{c}{Supervised ViT} \\
        Family & L4 & L8 & L12 & Emb.
               & L4 & L8 & L12 & L4 & L8 & L12 \\
        \hline
        \multicolumn{11}{l}{\textit{Global appearance transformations}} \\
        Blur
        & .918 & .748 & .499 & .519 & .914 & .780 & .264 & .942 & .727 & .402 \\
        Noise
        & .921 & .778 & .468 & .479 & .941 & .833 & .222 & .946 & .820 & .489 \\
        JPEG
        & .770 & .593 & .332 & .363 & .824 & .737 & .129 & .807 & .734 & .372 \\
        Brightness
        & .908 & .780 & .547 & .554 & .920 & .830 & .278 & .940 & .831 & .564 \\
        Red tint
        & .921 & .743 & .524 & .520 & .942 & .870 & .244 & .961 & .841 & .485 \\
        Cyan tint
        & .919 & .750 & .511 & .498 & .947 & .882 & .272 & .958 & .834 & .483 \\
        Blue tint
        & .922 & .730 & .519 & .512 & .938 & .873 & .249 & .958 & .839 & .465 \\
        Yellow tint
        & .920 & .723 & .477 & .464 & .932 & .874 & .238 & .962 & .850 & .525 \\
        \hline
        \multicolumn{11}{l}{\textit{Object-level processes}} \\
        Insert
        & .222 & .166 & .068 & .071 & .285 & .182 & .049 & .270 & .159 & .037 \\
        Grow
        & .121 & .102 & .050 & .053 & .150 & .118 & .043 & .156 & .100 & .029 \\
        Enter
        & .111 & .084 & .034 & .036 & .123 & .084 & .025 & .126 & .076 & .014 \\
        Colour
        & .347 & .167 & .173 & .178 & .349 & .225 & .071 & .413 & .264 & .136 \\
        \hline
        \multicolumn{11}{l}{\textit{Matched controls}} \\
        Insert blob
        & .319 & .316 & .245 & .258 & .385 & .348 & .066 & .404 & .402 & .102 \\
        Insert scramble
        & .574 & .341 & .190 & .196 & .474 & .323 & .118 & .443 & .343 & .166 \\
        Grow blob
        & .245 & .259 & .146 & .155 & .246 & .261 & .048 & .294 & .326 & .070 \\
        Enter blob
        & .195 & .196 & .113 & .123 & .180 & .171 & .040 & .214 & .207 & .036 \\
        Insert blur
        & .447 & .447 & .357 & .372 & .585 & .438 & .039 & .573 & .414 & .085 \\
        \hline
    \end{tabular}
    \caption{\textbf{Whole-path asymmetry across controlled visual processes.} Sphere-projected $R/\rho$ for all 17 trajectory families and 10 representations, averaged over three seeds. All values are positive, with every pointwise 95\% confidence interval above zero. Larger values indicate greater relative preference for forward traversal.}
    \label{tab:h4_photo_paths}
\end{table*}

\subsection{Object-Level Processes and Matched Controls}
\label{app:h4_matched_controls}

\paragraph{Strong accumulated directionality does not require intact object content.}
We evaluate five matched pairs: \textsc{Insert} against its blob, scrambled, and blurred controls, \textsc{Grow} against its blob control, and \textsc{Enter} against its blob control. These comparisons retain the corresponding transformation construction while modifying the manipulated content.

Table~\ref{tab:h4_control_comparisons} summarizes paired differences across the five process--control pairs and 10 representations. The controls have significantly higher reverse accuracy in 49/50 comparisons and significantly higher original chord-based $R/\rho$ in 48/50 comparisons. Shuffled accuracy is more mixed: controls win in 23 comparisons, object-level processes win in five, and 22 comparisons are unresolved.

The different outcomes show that the strength of directional discrimination depends on the chosen metric. Reverse accuracy and accumulated path asymmetry can be particularly high for synthetic controls, even when shuffled discrimination does not consistently favor those controls.

The paired significance counts for path asymmetry use the original ambient-chord analysis, for which paired confidence intervals are available. The sphere-projected values in Table~\ref{tab:h4_photo_paths} retain the broad pattern of strong control directionality, but the original and projected significance analyses are not treated as interchangeable.

\begin{table}[!htbp]
    \centering
    \small
    \setlength{\tabcolsep}{4pt}
    \begin{tabular}{lrrr}
        \hline
        Metric & \shortstack{Control\\higher} & \shortstack{Object\\higher} & Unresolved \\
        \hline
        Reverse accuracy & 49 & 1 & 0 \\
        Shuffled accuracy & 23 & 5 & 22 \\
        Chord-based $R/\rho$ & 48 & 1 & 1 \\
        \hline
    \end{tabular}
    \caption{\textbf{Paired object--control comparisons.} Counts across five matched process--control pairs and 10 representations. Higher indicates a paired 95\% confidence interval excluding zero in the corresponding direction. Local accuracies use sphere-tangent displacements; path comparisons use the original chord-based $R/\rho$.}
    \label{tab:h4_control_comparisons}
\end{table}

\paragraph{Interpretation.}
These results show that directional strength alone does not establish semantic dependence. Controlled changes in visibility, shape, or appearance may produce highly reproducible orientation preferences without requiring intact object content.

This does not contradict the field-similarity analysis. Two processes may both exhibit substantial path asymmetry while inducing nearly orthogonal directional fields at the same states. The former characterizes the strength of accumulated orientation preferences; the latter characterizes the organization of the learned fields. Object-category dependence is evaluated separately in the generalization experiments.

\subsection{Natural-Video Evaluation Protocol}
\label{app:h4_video_protocol}

\paragraph{Data and annotated events.}
We next examine whether fields learned from controlled photo trajectories exhibit meaningful directional preferences on natural video. This is a transfer evaluation: the fields are not adapted to the video data.

We evaluate 10 unique videos containing 1,626 frames sampled at $0.1$\,s intervals. All video frames use the same preprocessing and frozen feature extraction as the controlled photo trajectories.

The videos contain 21 manually annotated event intervals: seven approach, four recede, five ordinary-motion, and five static intervals. Approach and recede denote motion toward and away from the camera, respectively. Ordinary motion contains visual movement without a clear approach or recession event, while static intervals have comparatively little observed motion.

These annotations allow us to distinguish chronological ordering from process-specific visual change. A photo-trained \textsc{Grow} field, for example, may favor forward traversal during approach because the observed object becomes larger, but favor reverse traversal during recession because the object becomes smaller. A universal forward-time field would not exhibit this distinction.

\paragraph{Temporal windows and evaluation strides.}
Each evaluation window contains 10 representation states. Whole-video evaluation uses non-overlapping windows with temporal strides of $0.2$, $0.5$, and $1.0$\,s. Event-conditioned evaluation uses overlapping windows contained entirely within the corresponding annotated interval.

The primary event-conditioned comparisons use $0.1$ and $0.2$\,s strides, corresponding to window durations of $0.9$ and $1.8$\,s. At a $0.5$\,s stride, the approach class contributes only 14 windows from a single video. This setting is therefore excluded from the primary event-conditioned conclusions.

Table~\ref{tab:h4_video_counts} reports the number of windows and contributing videos for each event type. Because windows from the same recording are statistically dependent, uncertainty is estimated by resampling whole videos rather than individual windows.

\begin{table}[!htbp]
    \centering
    \small
    \setlength{\tabcolsep}{5pt}
    \begin{tabular}{lrrrr}
        \hline
        & \multicolumn{2}{c}{$0.1$\,s stride}
        & \multicolumn{2}{c}{$0.2$\,s stride} \\
        Event & Windows & Videos & Windows & Videos \\
        \hline
        Approach & 273 & 6 & 107 & 6 \\
        Recede & 150 & 3 & 57 & 3 \\
        Motion & 450 & 5 & 203 & 5 \\
        Static & 408 & 4 & 187 & 3 \\
        \hline
    \end{tabular}
    \caption{\textbf{Natural-video event-window counts.} Numbers of 10-state evaluation windows and contributing videos at two temporal strides. Event windows overlap within annotated intervals, so whole recordings rather than individual windows are resampled for uncertainty estimation.}
    \label{tab:h4_video_counts}
\end{table}

\subsection{Whole-Video and Event-Conditioned Results}
\label{app:h4_video_results}

\paragraph{Whole videos exhibit little consistent net directional preference.}
We first compute whole-path asymmetry across video windows without conditioning on event type. For all photo-trained fields, representations, and whole-video strides, the sphere-projected $R/\rho$ values lie between $-0.0203$ and $0.0203$.

The corresponding original chord-based measurements have a similar range. Across 510 family--representation--stride combinations, 38 pointwise 95\% confidence intervals exclude zero in the positive direction and 39 exclude zero in the negative direction (Table~\ref{tab:h4_whole_video}). These results indicate no consistent orientation favored across the complete video evaluations.

This does not mean that every video is time-symmetric or that no local directional effects exist. Rather, whole-video trajectories can contain different events whose directional contributions have opposite signs and partially cancel.

\begin{table*}[!htbp]
    \centering
    \small
    \setlength{\tabcolsep}{8pt}
    \begin{tabular}{lrrrr}
        \hline
        Field group & Cases & Chord-based $R/\rho$ range
        & CI above zero & CI below zero \\
        \hline
        Global appearance & 240 & $[-0.0204,+0.0160]$ & 15 & 13 \\
        Object-level & 120 & $[-0.0146,+0.0083]$ & 11 & 5 \\
        Matched controls & 150 & $[-0.0166,+0.0204]$ & 12 & 21 \\
        \hline
        All groups & 510 & $[-0.0204,+0.0204]$ & 38 & 39 \\
        \hline
    \end{tabular}
    \caption{\textbf{Whole-video asymmetry of photo-trained fields.} Results across family--representation--stride combinations, averaged over three field-training seeds. The ranges and pointwise 95\% video-bootstrap significance counts use the original chord-based analysis. The sphere-projected range across all combinations is $[-0.0203,+0.0203]$.}
    \label{tab:h4_whole_video}
\end{table*}

\paragraph{Event conditioning reveals opposite directional preferences.}
We next restrict the analysis to annotated events. Table~\ref{tab:h4_event_paths} reports mean sphere-projected $R/\rho$ for the photo-trained \textsc{Insert}, \textsc{Grow}, and \textsc{Enter} fields.

Across all three fields, approach events have positive asymmetry, whereas recede events have negative asymmetry. At a $0.2$\,s stride, the \textsc{Grow} field yields $+0.0135$ for approach and $-0.0231$ for recede, compared with $-0.0036$ for ordinary motion and $+0.0015$ for static intervals.

This pattern has a process-specific interpretation. The \textsc{Grow} field is trained on ordered photo sequences in which an object's apparent size increases. In natural video, it favors forward traversal when an object approaches the camera and becomes larger, but favors the reversed order when an object recedes and becomes smaller. Thus, its preference follows the orientation of the observed visual change rather than uniformly favoring chronological time.

The same sign pattern occurs for the \textsc{Insert} and \textsc{Enter} fields. However, the values are small relative to the stronger asymmetry measured on controlled photo trajectories. This distinction is expected from the experimental setting: natural videos contain additional changes beyond the specific controlled processes used for field training.

\begin{table*}[!htbp]
    \centering
    \small
    \setlength{\tabcolsep}{10pt}
    \begin{tabular}{llrrr}
        \hline
        Stride & Event & Insert & Grow & Enter \\
        \hline
        $0.1$\,s & Approach
        & $+0.0067$ & $+0.0089$ & $+0.0083$ \\
        & Recede
        & $-0.0106$ & $-0.0161$ & $-0.0150$ \\
        & Ordinary motion
        & $-0.0024$ & $-0.0028$ & $-0.0021$ \\
        & Static
        & $+0.0023$ & $+0.0011$ & $+0.0010$ \\
        \hline
        $0.2$\,s & Approach
        & $+0.0100$ & $+0.0135$ & $+0.0126$ \\
        & Recede
        & $-0.0158$ & $-0.0231$ & $-0.0217$ \\
        & Ordinary motion
        & $-0.0025$ & $-0.0036$ & $-0.0025$ \\
        & Static
        & $+0.0032$ & $+0.0015$ & $+0.0013$ \\
        \hline
    \end{tabular}
    \caption{\textbf{Event-conditioned directional asymmetry in natural video.} Mean sphere-projected $R/\rho$ across 10 representations and three training seeds. Fields are learned from controlled photo edits and applied to natural video without adaptation. Positive values favor forward traversal; negative values favor reverse traversal.}
    \label{tab:h4_event_paths}
\end{table*}

\paragraph{Sign consistency and statistical uncertainty.}
The expected sign holds in all 60 approach and all 60 recede combinations formed from three process-trained fields, 10 representations, and two evaluation strides. This demonstrates consistent orientation at the level of reported point estimates.

Sign consistency does not imply that every confidence interval excludes zero. At a $0.2$\,s stride, the video-bootstrap intervals exclude zero in the expected direction for 5/10, 7/10, and 8/10 approach representations for \textsc{Insert}, \textsc{Grow}, and \textsc{Enter}, respectively. The corresponding recede counts are 6/10, 8/10, and 7/10.

These uncertainty results qualify the conclusion. The video experiments provide evidence for a consistent event-conditioned directional pattern across the evaluated fields, but do not establish equally strong statistical support for every individual representation. The small number of contributing recordings, especially for recede events, also limits precision.

\paragraph{The event pattern is visible across contributing videos.}
Table~\ref{tab:h4_per_video} reports sphere-projected event asymmetry separately for contributing videos at a $0.2$\,s stride. All three process-trained fields produce positive values for approach in each of six contributing videos and negative values for recede in each of three videos.

The magnitudes vary considerably across recordings. For example, the \textsc{Grow} approach score is $+0.0020$ in video 16011564 but $+0.0444$ in video 16235000. Ordinary-motion and static groups exhibit values of both signs. Some larger deviations arise from event groups with few windows, including two static windows in video 16235000 and four ordinary-motion windows in video 16905315.

Thus, the aggregate event pattern should not be interpreted as a uniform effect of identical strength across all recordings. It indicates a reproducible orientation trend within the sampled events, with substantial variation in magnitude.

\begin{table*}[!htbp]
    \centering
    \small
    \setlength{\tabcolsep}{9pt}
    \begin{tabular}{llrrrr}
        \hline
        Event & Video ID & Insert & Grow & Enter & Windows \\
        \hline
        Approach & 12503985 & $+.0227$ & $+.0373$ & $+.0219$ & 9 \\
                 & 14241144 & $+.0167$ & $+.0205$ & $+.0185$ & 4 \\
                 & 16011564 & $+.0008$ & $+.0020$ & $+.0035$ & 59 \\
                 & 16235000 & $+.0358$ & $+.0444$ & $+.0429$ & 9 \\
                 & 16359676 & $+.0088$ & $+.0116$ & $+.0112$ & 14 \\
                 & 16905315 & $+.0256$ & $+.0287$ & $+.0277$ & 12 \\
        \hline
        Recede & 16235000 & $-.0330$ & $-.0464$ & $-.0443$ & 18 \\
               & 16905315 & $-.0088$ & $-.0235$ & $-.0147$ & 5 \\
               & 16905676 & $-.0077$ & $-.0107$ & $-.0108$ & 34 \\
        \hline
        Motion & 12813896 & $+.0014$ & $+.0009$ & $+.0017$ & 42 \\
               & 14411398 & $-.0027$ & $-.0036$ & $-.0036$ & 110 \\
               & 16235000 & $-.0085$ & $-.0114$ & $-.0072$ & 30 \\
               & 16359676 & $-.0026$ & $-.0056$ & $-.0029$ & 17 \\
               & 16905315 & $+.0056$ & $+.0173$ & $+.0198$ & 4 \\
        \hline
        Static & 15526828 & $+.0052$ & $+.0046$ & $+.0053$ & 69 \\
               & 16235000 & $-.0077$ & $-.0189$ & $-.0175$ & 2 \\
               & 16905676 & $+.0023$ & $+.0001$ & $-.0007$ & 116 \\
        \hline
    \end{tabular}
    \caption{\textbf{Per-video event asymmetry.} Sphere-projected $R/\rho$ at a $0.2$\,s stride, averaged over 10 representations and three seeds. Multiple annotated intervals of the same event within a video are grouped together. The number of contributing windows is reported to indicate the support underlying each mean.}
    \label{tab:h4_per_video}
\end{table*}

\subsection{Temporal-Ordering Controls and Scope of the Video Evidence}
\label{app:h4_video_controls}

\paragraph{The event-conditioned signal depends on frame order.}
An important question is whether the measured preferences depend on the actual ordering of visual changes or merely on the image states appearing in a video.

Reversing the state order changes the sign of the forward--reverse asymmetry by construction. More informatively, randomly shuffling the frames substantially reduces the event-conditioned effect. Table~\ref{tab:h4_order_controls} reports this comparison for the photo-trained \textsc{Grow} field.

At a $0.2$\,s stride, the original chord-based approach and recede scores change from $+0.0135$ and $-0.0232$ under observed ordering to $+0.0007$ and $-0.0010$ under shuffled ordering. The same pattern is visible at a $0.1$\,s stride.

Thus, the event-conditioned preference is associated with the ordered progression of visual states rather than a directional effect that remains equally strong under arbitrary ordering.

\begin{table}[!htbp]
    \centering
    \small
    \setlength{\tabcolsep}{5pt}
    \begin{tabular}{llrr}
        \hline
        Stride & Order & Approach & Recede \\
        \hline
        $0.1$\,s & Observed & $+0.0090$ & $-0.0161$ \\
                 & Shuffled & $-0.0002$ & $-0.0004$ \\
        $0.2$\,s & Observed & $+0.0135$ & $-0.0232$ \\
                 & Shuffled & $+0.0007$ & $-0.0010$ \\
        \hline
    \end{tabular}
    \caption{\textbf{Frame-order control for the photo-trained \textsc{Grow} field.} Original chord-based $R/\rho$ on approach and recede events, averaged across representations and seeds. Randomly shuffling the frames substantially reduces the directional effect.}
    \label{tab:h4_order_controls}
\end{table}

\paragraph{Local discrimination alone does not establish an arrow of time.}
We additionally train directional fields directly on natural-video trajectories using leave-one-video-out evaluation. This provides a separate test of whether local directional discrimination necessarily captures a persistent temporal orientation.

We compare fields trained on observed, reversed, and randomly ordered frames. Table~\ref{tab:h4_video_training} reports the resulting mean local reverse accuracy and whole-path asymmetry.

All three training conditions yield local reverse accuracy above $50\%$. In particular, randomly ordered training reaches $65.6\%$ mean reverse accuracy, compared with $63.3\%$ under observed ordering and $62.3\%$ under reversed ordering. However, the corresponding whole-path asymmetry remains small, and only a minority of confidence intervals exclude zero.

This control demonstrates that local directional discrimination does not, by itself, establish a persistent chronological arrow. The learned ruler must be interpreted relative to the visual process, training order, and trajectory distribution. This is also why we report accumulated asymmetry separately from local accuracy.

\begin{table*}[!htbp]
    \centering
    \small
    \setlength{\tabcolsep}{8pt}
    \begin{tabular}{lrrr}
        \hline
        Video training order
        & Mean reverse accuracy (\%)
        & Chord-based $R/\rho$ range
        & CI excludes zero \\
        \hline
        Observed
        & 63.3 & $[-0.0103,\,+0.0064]$ & 4/30 \\
        Reversed
        & 62.3 & $[-0.0040,\,+0.0059]$ & 0/30 \\
        Random
        & 65.6 & $[-0.0113,\,+0.0044]$ & 2/30 \\
        \hline
    \end{tabular}
    \caption{\textbf{Natural-video training-order controls.} Leave-one-video-out evaluation of fields trained on observed, reversed, or randomly ordered video frames. The 30 cases comprise 10 representations and three whole-video strides. Reverse accuracy uses sphere-tangent comparisons; path ranges and confidence counts use the original chord-based evaluation.}
    \label{tab:h4_video_training}
\end{table*}

\paragraph{Event-dependent preferences are not exclusively semantic.}
The matched-control experiments suggest that visual processes can exhibit directional structure without requiring intact semantic object content. The natural-video experiments provide a related observation.

Blob-trained fields retain part of the recede signal. At a $0.1$\,s stride, the original chord-based \textsc{Grow}-blob field produces $R/\rho=-0.0085$ on recede events, compared with $-0.0161$ for the corresponding real-object \textsc{Grow} field. Changes in apparent size or visibility can therefore contribute to the event-conditioned preference even when the training process does not preserve meaningful object appearance.

Static intervals also need not produce exactly zero asymmetry. For example, the sphere-projected \textsc{Insert} field has mean $R/\rho=+0.0032$ on static intervals at a $0.2$\,s stride. These effects may reflect small visual changes or representation fluctuations and should not be interpreted as categorical event recognition.

Together, these findings support an event-sensitive directional preference, but not an exclusively semantic detector of temporal direction.

\begin{figure*}[!htbp]
    \centering
    \includegraphics[width=\textwidth]{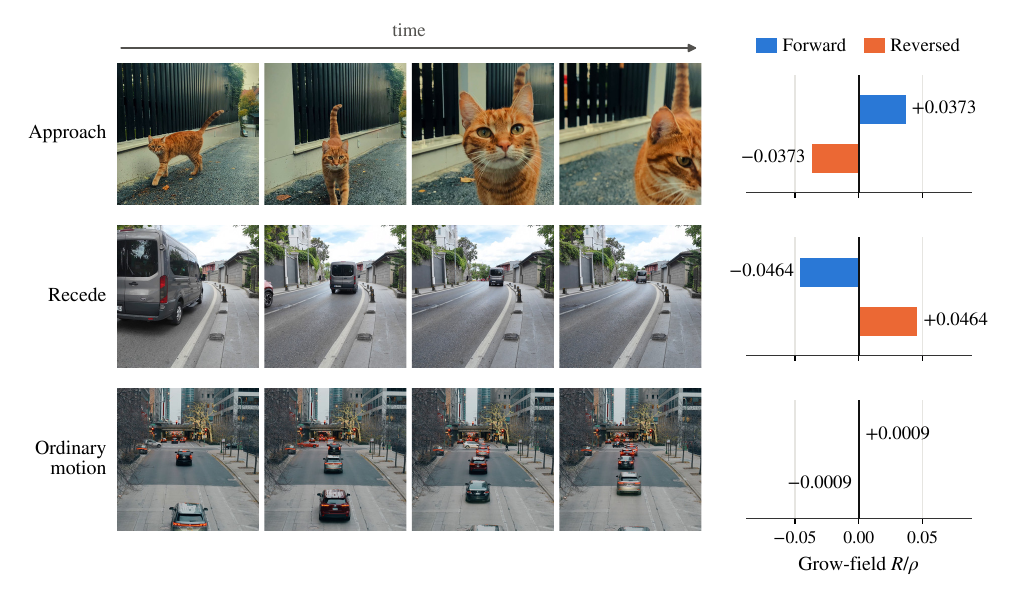}
    \caption{\textbf{Event-dependent directional preferences in natural video.} Rows show an approaching cat (video 12503985, 0.0--3.6\,s), a receding van (video 16235000, 10.0--14.5\,s), and ordinary traffic (video 12813896, 0.0--10.2\,s), with four evenly spaced frames per interval. Bars show sphere-projected $R/\rho$ of the photo-trained \textsc{Grow} field, pooled over marked intervals of the corresponding event within each video at a $0.2$\,s stride and averaged across 10 representations. Forward and reversed traversal have opposite asymmetry signs. Examples are selected for clarity of the annotated events.}
    \label{fig:h4_video_examples}
\end{figure*}

\paragraph{Representative event examples.}
Figure~\ref{fig:h4_video_examples} illustrates an approaching cat, a receding van, and ordinary street traffic. The corresponding \textsc{Grow}-field asymmetry is reported under forward and reversed traversal.

The examples are selected for visual clarity of the annotated events, not for unusually large geometric scores. They illustrate the process dependence observed in the aggregate evaluation, where all six contributing approach videos have positive mean \textsc{Grow}-field asymmetry and all three contributing recede videos have negative mean asymmetry.

\paragraph{Scope and limitations of the natural-video evidence.}
The video evaluation contains 10 recordings and 21 annotated intervals marked by one annotator. Although event windows provide many local observations, they are correlated within recordings. In particular, recede events are drawn from only three videos. Resampling whole videos accounts for within-recording dependence but does not eliminate the uncertainty associated with the small number of recordings.

The supported conclusion is therefore specific: photo-trained directional fields exhibit small but consistently oriented preferences on the tested approach and recede events, with effects that depend on visual event type and frame ordering. The results do not establish a universal forward-time preference, the absence of temporal asymmetry in natural video generally, or a broadly validated detector of semantic events.

\paragraph{Overall interpretation.}
The controlled and natural-video analyses answer two complementary questions about process-dependent geometry.

At the field level, different visual processes induce distinct local directional preferences, and initially aligned fields become increasingly differentiated in deeper representations. This pattern persists after removing the fitted sink components and is also observed in comparisons between object-level processes and matched controls.

At the trajectory level, the strength and orientation of accumulated asymmetry depend on the visual process. Global appearance transformations produce stronger relative asymmetry than the tested object-level processes, while natural-video events reveal opposite preferences for approach and recession under photo-trained object-process fields.

These findings establish process dependence in both the learned local field and its accumulated effect along observed paths. They do not imply that all directional structure is semantic, that stronger field differentiation produces stronger path asymmetry, or that every visual trajectory has a universal preferred temporal direction.



\section{Extended Functional Validation via Restoration Selection}
\label{app:restoration}

This section extends Section~\ref{sec:restoration} by providing the complete experimental protocol, candidate constructions, geometric scoring functions, statistical analyses, and additional results for restoration selection. The objective is to determine whether directional geometry learned from ordered visual trajectories can guide a practical decision on previously unseen images.

We consider a known degradation process and a predefined set of candidate corrections. Given a degraded image, a scorer ranks these candidates without access to the clean reference. The clean image is used only after selection to evaluate restoration quality. DiRGeo does not train an image restoration network or generate corrected images; its role is to evaluate candidate transformations using the directional fields learned in the preceding experiments.

The main paper emphasizes colour correction, where the learned field provides a measurable improvement over a matched fitted-sink reference. Here we additionally examine deblurring, alternative scoring rules, exploratory development experiments, and the statistical limits of the final confirmation. These analyses clarify when the learned directional geometry contributes useful information and when a simpler attractor-based explanation remains competitive.

\subsection{Data and Candidate Construction}
\label{app:h7_data}

\paragraph{Evaluation cohorts and experimental independence.}
The restoration experiments involve four image-based evaluation runs, each using 200 COCO train2017 source photographs. Their sampling seeds are 2029, 2030, 2031, and 2032, respectively. All cohorts exclude the images used in the preceding non-restoration experiments, ensuring that the restoration evaluations do not reuse the images employed for directional-field training or the primary geometric tests.

The experiments were conducted in successive stages. Step 1 examines pairwise restoration discrimination, while Step 2 evaluates ranking among multiple candidates. Phase A reanalyzes saved results to interpret the roles of the sink and residual fields. Phase B repeats ranking experiments on a new image cohort, and Phase C reuses those results to compare alternative cross-process penalties. Finally, Phase D evaluates the selected scoring rule on a separate confirmation cohort.

Phase B excludes the images used in Steps 1 and 2, and Phase D excludes all earlier restoration cohorts. Step 2 uses a different sampling seed from Step 1 but does not explicitly exclude its images, so a small overlap between those exploratory cohorts is possible. Phases A and C do not introduce additional source-image cohorts because they analyze previously computed results.

All restoration experiments use the same 10 frozen encoder representations and three trained seeds per directional field as the geometric characterization experiments. Eight global appearance fields are available for each representation: blur, noise, JPEG compression, brightness, and red, cyan, blue, and yellow colour casts. These fields are trained on their corresponding ordered degradation trajectories and are not retrained for restoration selection.

\paragraph{Degradations and candidate restorations.}
We evaluate two types of restoration decisions: correcting a known colour cast and reducing Gaussian blur. The degraded images are generated using the same transformation constructions employed for directional-field training, but from previously unseen source images. Candidate corrections are generated independently of the learned fields.

For blur, we consider 12 classical candidates: Richardson--Lucy deconvolution with 3, 10, 30, or 100 iterations; Wiener filtering with parameters 0.3, 0.1, 0.03, or 0.01; and unsharp masking at 50, 150, 300, or 600\%. These candidates represent different restoration mechanisms and correction strengths.

For each colour cast, we consider eight candidates. Five apply inverse colour correction with strength multipliers $0.5$, $0.75$, $1.0$, $1.25$, and $1.5$, while three use gray-world correction, white-patch correction, and contrast stretching. The multiplier $1.0$ corresponds to the exact inverse of the known synthetic cast. Table~\ref{tab:h7_candidates} summarizes these candidate sets.

\begin{table*}[!htbp]
    \centering
    \small
    \setlength{\tabcolsep}{7pt}
    \begin{tabular}{llr}
        \hline
        Degradation & Candidate constructions & Count \\
        \hline
        Blur
        & Richardson--Lucy: 3, 10, 30, 100 iterations & 4 \\
        & Wiener: 0.3, 0.1, 0.03, 0.01 & 4 \\
        & Unsharp mask: 50, 150, 300, 600\% & 4 \\
        \hline
        Colour cast
        & Inverse correction: $\times0.5$, 0.75, 1.0, 1.25, 1.5 & 5 \\
        & Gray world, white patch, contrast stretch & 3 \\
        \hline
    \end{tabular}
    \caption{\textbf{Restoration candidate constructions.} Each degraded image has 12 blur candidates or eight colour-correction candidates. The colour set includes the exact inverse transformation at multiplier $1.0$. The candidate sets are fixed independently of the learned directional fields.}
    \label{tab:h7_candidates}
\end{table*}

\paragraph{Degradation severity and final confirmation.}
The final confirmation uses blur severity levels 2 and 3 and colour-cast severity levels 3 and 6. The blur levels were fixed before Phase D because the available classical restoration methods provided limited improvements at the strongest degradation levels. At levels 2 and 3, the best available blur candidate improves PSNR over the degraded image by median values of $1.63$ and $1.28$\,dB, respectively. These levels therefore provide nontrivial but practically distinguishable candidate rankings.

For controlled colour correction, the exact inverse is the highest-PSNR candidate for every source image in the final confirmation cohort. This makes the top-1 selection criterion especially interpretable: it measures whether a scorer identifies the known inverse correction among the eight alternatives.

The exact inverse can produce very large PSNR improvements because the degradation is synthetically generated. PSNR is capped at 100\,dB in the evaluation. Accordingly, we emphasize candidate-selection accuracy and ranking metrics rather than interpreting large colour-correction PSNR gains as evidence of comparable improvements in general restoration quality.

\subsection{Geometric Restoration Scores}
\label{app:h7_scores}

\paragraph{Scoring motion against a degradation process.}
Let $p$ denote the known degradation family, such as a red colour cast, and let $z_d$ and $z_c$ be the normalized representations of the degraded image and a candidate correction. The field $\beta_{\theta,p}$ was learned from trajectories in which degradation severity increased. A successful correction should therefore move the representation against the preferred degradation direction.

This leads to a geometric selection principle: among candidate corrections, prefer those whose representation paths exhibit stronger motion opposite to the learned degradation field. The scoring function does not require the clean representation, only the degraded and candidate representations and the previously trained field.

\paragraph{Local direction score.}
As a simple scoring rule, we evaluate the learned field at the degraded representation alone. Let $v=P_{z_d}(z_c-z_d)$ be the sphere-tangent displacement from the degraded state toward the candidate. The local direction score is
\begin{equation}
    S_p^{\mathrm{dir}}(z_d,z_c)
    =
    -\frac{
        \beta_{\theta,p}^{\mathrm{tan}}(z_d)^\top v
    }{
        \|v\|_2
    }.
    \label{eq:h7_direction}
\end{equation}
A larger score indicates that the candidate displacement points more strongly against the learned degradation direction. This score characterizes the initial direction of correction, but does not measure how the field changes along the path or how far the candidate moves.

\paragraph{Integrated restoration score.}
To capture directional preference along the entire candidate path, we consider the great-circle trajectory $\gamma_{d,c}$ connecting $z_d$ and $z_c$ on the unit sphere. The own-process learned-field score is
\begin{equation}
    A_p^{\mathrm{learn}}(z_d,z_c)
    =
    -\int_{\gamma_{d,c}}
    \beta_{\theta,p}^{\mathrm{tan}}(z)^\top dz.
    \label{eq:h7_path}
\end{equation}
The negative sign rewards movement against the degradation field. Thus, a larger value indicates a stronger accumulated directional preference for the candidate correction.

This score uses only the directional component of the Randers ruler, rather than the complete path length. The objective is not to prefer the shortest candidate path, but to quantify how strongly the candidate reverses the learned degradation progression. It also differs in sign from the forward wind-alignment score used in the same-endpoint journey experiments.

We approximate the integral using 16 midpoint steps along the spherical path and perform the computation in double precision. Unlike the local direction score, integration evaluates the field at intermediate representation states and accounts for the extent of candidate motion.

\paragraph{Fitted-sink reference score.}
For a matched geometric baseline, we replace the learned wind with the spherical sink fitted to approximate that field. Let $c_p$ denote the fitted attractor for degradation process $p$. Under the sink convention used in the implementation, the integrated reference score has the closed form
\begin{equation}
    A_p^{\mathrm{sink}}(z_d,z_c)
    =
    \rho\left[
        \operatorname{angle}(z_c,c_p)
        -
        \operatorname{angle}(z_d,c_p)
    \right].
    \label{eq:h7_sink_path}
\end{equation}
Here $\operatorname{angle}(z,c)=\arccos(z^\top c)$ is the spherical angular distance between normalized representations. A positive value indicates that the candidate moves farther from the degradation attractor than the original degraded state.

The sink score is particularly important because the fitted attractor already captures part of the learned degradation direction. Comparing the learned and sink scores therefore tests whether the state-dependent wind contributes information beyond this simpler geometric reference.

\paragraph{Penalizing movement toward other degradation attractors.}
A candidate that moves against one degradation field is not necessarily a good correction. For example, removing a red cast too aggressively can introduce a cyan cast. Such a candidate may strongly oppose the red degradation direction while moving toward another undesirable appearance state.

To address this ambiguity, we use the fitted sinks of the other global appearance processes as additional constraints. For known degradation $p$, the penalty is
\begin{equation}
    P_p(z_d,z_c)
    =
    \sum_{q\neq p}
    \max\left\{
        0,
        -A_q^{\mathrm{sink}}(z_d,z_c)
    \right\},
    \label{eq:h7_penalty}
\end{equation}
where the sum covers the other seven global appearance processes. A negative $A_q^{\mathrm{sink}}$ indicates movement toward the attractor associated with degradation $q$, which incurs a positive penalty.

We evaluate the learned-field score and its matched sink reference using identical penalties:
\begin{align}
    S_p^{\mathrm{learn}}
    &=
    A_p^{\mathrm{learn}}-\lambda P_p, \\
    S_p^{\mathrm{sink}}
    &=
    A_p^{\mathrm{sink}}-\lambda P_p.
    \label{eq:h7_matched_scores}
\end{align}
The penalty weight is fixed at $\lambda=1$ before the final confirmation experiment.

This construction separates two effects. The common penalty discourages candidate paths that resemble other degradation processes, while the own-process term distinguishes the learned directional field from its fitted-sink approximation. Because the two scores share the same penalty, their comparison specifically tests the additional benefit of the learned own-process geometry.

\paragraph{Alternative scoring rules and baselines.}
We additionally evaluate the learned and sink path scores without cross-process penalties, a variant that uses other learned fields rather than fitted sinks for the penalties, and the local direction score. These alternatives help determine whether performance gains arise from path integration, cross-process constraints, or the learned field itself.

Non-geometric references include selection of the candidate nearest to the degraded image in feature space and uniform random selection. We also evaluate three no-reference image-quality estimators: NIQE, BRISQUE, and MUSIQ, implemented using \texttt{pyiqa} version \texttt{0.1.15.post2}. Their score directions are normalized so that higher values indicate better predicted image quality.

The no-reference quality estimators do not use the known degradation field, whereas the geometric scorers explicitly condition on the degradation family. The comparisons therefore evaluate practical candidate-ranking behavior under the specified task, rather than claiming that all methods use identical prior information.

\subsection{Evaluation Metrics and Statistical Protocol}
\label{app:h7_evaluation}

\paragraph{Top-1 candidate-selection accuracy.}
The primary metric is top-1 accuracy: the percentage of degraded-image instances for which the highest-scored candidate is also the candidate with the highest PSNR relative to the clean reference. The clean image is used only to establish this evaluation target and is not provided to the scorer.

With eight colour candidates, uniform random selection has expected top-1 accuracy $12.5\%$. With 12 blur candidates, the corresponding reference is approximately $8.3\%$. Top-1 accuracy is stringent because selecting the second-best candidate counts as incorrect even if its restoration quality is close to the optimum.

\paragraph{Ranking metrics.}
To complement exact selection, we report the percentile of the selected candidate within the PSNR ranking. This measures the fraction of alternative candidates with lower PSNR than the selected candidate, with $0.5$ as the random-selection reference.

We additionally report Spearman rank correlation between the geometric score ordering and PSNR ordering across candidates. Unlike top-1 accuracy, Spearman correlation considers the complete ordering and can distinguish scorers that provide useful overall rankings even when they do not select the exact optimum.

SSIM and LPIPS provide secondary restoration-quality assessments. Their purpose is to check whether conclusions based on PSNR are compatible with other measures of image similarity, rather than replacing the primary candidate-selection criterion.

\paragraph{Aggregation and uncertainty.}
Results are aggregated within source images over degradation severity levels and training seeds, and across representations for pooled summaries. Confidence intervals are estimated using 2,000 source-image bootstrap resamples with bootstrap seed 119. Resampling source images retains the dependence among their severity variants, candidate evaluations, representations, and training seeds.

Paired differences between scorers are computed using the same source images. This provides a more informative uncertainty estimate than comparing two independently bootstrapped marginal accuracies.

We distinguish pooled performance from representation-specific confirmation. A pooled difference can be statistically significant even when the improvement is not consistently detectable in every encoder representation. Accordingly, the final primary criterion requires the paired 95\% confidence interval for the learned-versus-sink top-1 improvement to lie above zero in at least eight of the 10 representations.

\paragraph{Final confirmation criterion.}
Phase D evaluates whether the learned-field score with cross-process sink penalties improves colour-correction top-1 accuracy over the fitted-sink score with the same penalties. The predefined criterion requires a significant positive improvement in at least 8/10 representations.

This is intentionally stronger than requiring only a positive pooled mean. The criterion evaluates whether the benefit of the learned own-process field generalizes across representation choices rather than depending on a small subset of favorable encoders or layers.

\paragraph{Correctness and numerical checks.}
The experimental pipeline checks that the degradation construction is reproduced, saved encoder features are recovered consistently, and the previously trained field evaluations agree with the saved results. Path-score computations additionally check the field decomposition and agreement between numerical and analytic sink integration.

In Phase D, the maximum relative discrepancy between the numerical sink integral and its closed-form expression is $4.2\times10^{-4}$. The comparison between 16-point and 64-point integration gives a discrepancy of $5.5\times10^{-4}$. Both are below the specified tolerance of $10^{-2}$, supporting the numerical stability of the path-score implementation.

\subsection{Final Confirmation: Colour Correction}
\label{app:h7_colour}

\paragraph{Learned directional geometry improves pooled colour selection.}
Table~\ref{tab:h7_colour_full} reports the final colour-correction results. The proposed learned-field score with other-process sink penalties achieves $31.8\%$ top-1 accuracy, with a 95\% confidence interval of $[29.8,33.8]$. The matched sink-only score achieves $27.8\%$ $[25.6,30.1]$, while the learned own-process path without penalties reaches $21.4\%$ $[19.7,22.9]$.

The paired improvement over the matched sink is $3.98$ percentage points, with a 95\% confidence interval of $[2.76,5.14]$. Thus, the pooled evaluation supports a positive contribution from the learned own-process directional field beyond the fitted attractor reference.

The ranking metrics are consistent with this conclusion. The learned score improves selected-candidate percentile from $0.693$ to $0.716$ and Spearman correlation from $0.587$ to $0.602$ relative to the matched sink. These improvements are smaller than the top-1 difference, indicating that the most visible benefit is more frequent selection of the exact best correction rather than a large change in the entire candidate ranking.

\begin{table*}[!htbp]
    \centering
    \small
    \setlength{\tabcolsep}{9pt}
    \begin{tabular}{lrrr}
        \hline
        Scorer & Top-1 (\%) & Percentile & Spearman PSNR \\
        \hline
        Learned + sink penalties
        & $31.8\ [29.8,33.8]$ & .716 & .602 \\
        Sink-only + sink penalties
        & $27.8\ [25.6,30.1]$ & .693 & .587 \\
        Own learned path only
        & $21.4\ [19.7,22.9]$ & .605 & .467 \\
        Own sink path only
        & 18.7 & .588 & .469 \\
        Learned + learned-field penalties
        & 23.6 & .578 & .331 \\
        \hline
        MUSIQ
        & $18.2\ [14.7,21.6]$ & .563 & .365 \\
        BRISQUE
        & 9.1 & .406 & .041 \\
        NIQE
        & 8.8 & .437 & .072 \\
        Nearest to degraded
        & 0.0 & .400 & -.048 \\
        Random
        & 12.5 & .500 & .000 \\
        \hline
    \end{tabular}
    \caption{\textbf{Final colour-correction confirmation.} Four colour casts with eight candidate corrections each, evaluated on 200 held-out images. Results aggregate source images, severity levels, representations, and training seeds where applicable. Brackets give available source-image bootstrap 95\% confidence intervals. Larger values are better for all three metrics.}
    \label{tab:h7_colour_full}
\end{table*}

\paragraph{The improvement is positive across all four colour casts.}
Table~\ref{tab:h7_casts} separates the red, cyan, blue, and yellow correction tasks. DiRGeo improves top-1 selection over the matched fitted-sink score for every cast. The gains are $4.14$ percentage points for red, $4.98$ for cyan, $2.93$ for blue, and $3.87$ for yellow, with all four pooled paired confidence intervals above zero.

The largest improvement occurs for cyan, where accuracy increases from $30.1\%$ to $35.1\%$. Blue exhibits the smallest gain, increasing from $24.6\%$ to $27.6\%$. These differences show that the contribution of the learned field varies with the degradation process, even within the controlled colour-correction setting.

\begin{figure*}[!htb]
    \centering
    \includegraphics[width=0.90\textwidth]{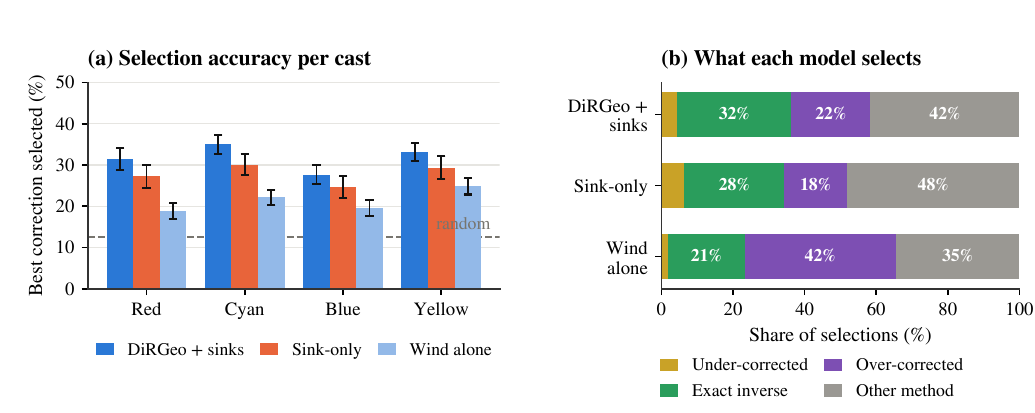}
    \caption{\textbf{Colour-correction selection accuracy and decision behavior.} (a) Top-1 accuracy for red, cyan, blue, and yellow casts under the learned field with sink penalties, matched sink-only score, and learned field without penalties. The dashed line marks the $12.5\%$ random-selection reference for eight candidates. (b) Distribution of selected correction types across colour casts, grouped into undercorrection, exact inverse, overcorrection, and other methods. Adding other-process sink penalties substantially reduces overcorrection, while the learned own-process term improves exact-inverse selection over the matched sink reference. Percentages in the selection-distribution panel are rounded.}
    \label{fig:h7_appendix_stats}
\end{figure*}

Figure~\ref{fig:h7_appendix_stats}(a) visualizes the per-cast top-1 results and the random-selection reference. The figure makes two observations easier to interpret than aggregate accuracy alone: all four cast types benefit from the learned score, but none approaches perfect inverse selection. Directional geometry provides useful ranking information rather than fully determining the correct restoration.

\begin{table*}[!htbp]
    \centering
    \small
    \setlength{\tabcolsep}{8pt}
    \begin{tabular}{lrrrr}
        \hline
        Cast & Learned (\%) & Sink-only (\%)
        & Paired gain [95\% CI] & Representations with CI $>0$ \\
        \hline
        Red    & 31.4 & 27.3 & $4.14\ [2.51,5.73]$ & 5/10 \\
        Cyan   & 35.1 & 30.1 & $4.98\ [3.25,6.66]$ & 7/10 \\
        Blue   & 27.6 & 24.6 & $2.93\ [1.05,4.73]$ & 5/10 \\
        Yellow & 33.2 & 29.3 & $3.87\ [2.01,5.70]$ & 4/10 \\
        \hline
        All casts & 31.8 & 27.8 & $3.98\ [2.76,5.14]$ & 6/10 \\
        \hline
    \end{tabular}
    \caption{\textbf{Per-cast top-1 selection in Phase D.} Learned and fitted-sink scores use identical other-process penalties. Paired gains are expressed in percentage points and computed before rounding displayed accuracies. The final column reports the number of representations with paired 95\% confidence intervals above zero.}
    \label{tab:h7_casts}
\end{table*}

\paragraph{Penalties change which corrections are selected.}
The performance of the own learned path without penalties provides an important control. Its top-1 accuracy is only $21.4\%$, compared with $31.8\%$ after adding other-process sink penalties. The penalty therefore contributes substantially to effective colour-correction selection.

The role of the penalty is more interpretable when examining which candidates each scorer prefers. An own-process wind can reward strong movement against the degradation direction, including excessive correction. For example, correcting a red cast too aggressively may introduce a cyan cast. Other-process sink penalties discourage such candidates by detecting movement toward alternative degradation attractors.

Figure~\ref{fig:h7_appendix_stats}(b) groups the selected candidates into undercorrection, exact inverse, overcorrection, and other restoration methods. With the learned field alone, overcorrected inverse candidates account for approximately $42\%$ of selections. Adding sink penalties reduces this share to approximately $22\%$, while increasing exact-inverse selection from approximately $21\%$ to $32\%$. The matched sink-only score selects the exact inverse in approximately $28\%$ of cases and other restoration methods more frequently.

These aggregated selection patterns support the interpretation that other-process penalties help suppress excessive corrections, while the learned own-process field provides an additional preference among the remaining candidates. The figure reports rounded proportions across colour-correction selections; the exact top-1 rates are provided in Table~\ref{tab:h7_colour_full}.

\paragraph{Red-cast decisions illustrate the correction tradeoff.}
Table~\ref{tab:h7_red_choices} reports the distribution of selected candidates for the red cast. The strongest inverse correction, with multiplier $1.5$, is selected in $31.6\%$ of cases by the own learned path without penalties, but only $6.0\%$ after adding the other-process penalties. This is consistent with the penalty discouraging excessive correction toward the complementary colour.

Relative to the matched sink-only score, DiRGeo increases exact-inverse selection from $27.3\%$ to $31.4\%$ and reduces contrast-stretch selection from $47.4\%$ to $41.0\%$. Contrast stretching nevertheless remains the most frequently selected candidate for both scores. Thus, the learned field improves the decision distribution but does not reliably recover the correct inverse transformation in every case.

\begin{table*}[!htbp]
    \centering
    \small
    \setlength{\tabcolsep}{8pt}
    \begin{tabular}{lrrrrr}
        \hline
        Candidate & Best by PSNR & Learned + penalties
        & Sink-only + penalties & Own learned path & MUSIQ \\
        \hline
        Inverse $\times0.5$  & 0 & 0.2 & 0.5 & 0.0 & 27.8 \\
        Inverse $\times0.75$ & 0 & 2.1 & 3.0 & 0.6 & 16.2 \\
        Inverse $\times1.0$  & 100 & 31.4 & 27.3 & 18.7 & 10.5 \\
        Inverse $\times1.25$ & 0 & 16.1 & 13.6 & 14.9 & 7.8 \\
        Inverse $\times1.5$  & 0 & 6.0 & 4.2 & 31.6 & 3.8 \\
        Gray world          & 0 & 2.4 & 3.4 & 1.1 & 5.2 \\
        White patch         & 0 & 0.8 & 0.6 & 0.6 & 17.2 \\
        Contrast stretch    & 0 & 41.0 & 47.4 & 32.4 & 11.5 \\
        \hline
    \end{tabular}
    \caption{\textbf{Distribution of selected candidates for the red cast (\%).} The exact inverse at multiplier $1.0$ is the highest-PSNR candidate in all evaluated instances. Columns show how frequently each scorer selects a candidate, not its average PSNR. Percentages may differ slightly from 100 due to rounding.}
    \label{tab:h7_red_choices}
\end{table*}

\begin{figure*}[!htb]
    \centering
    \includegraphics[width=0.90\textwidth]{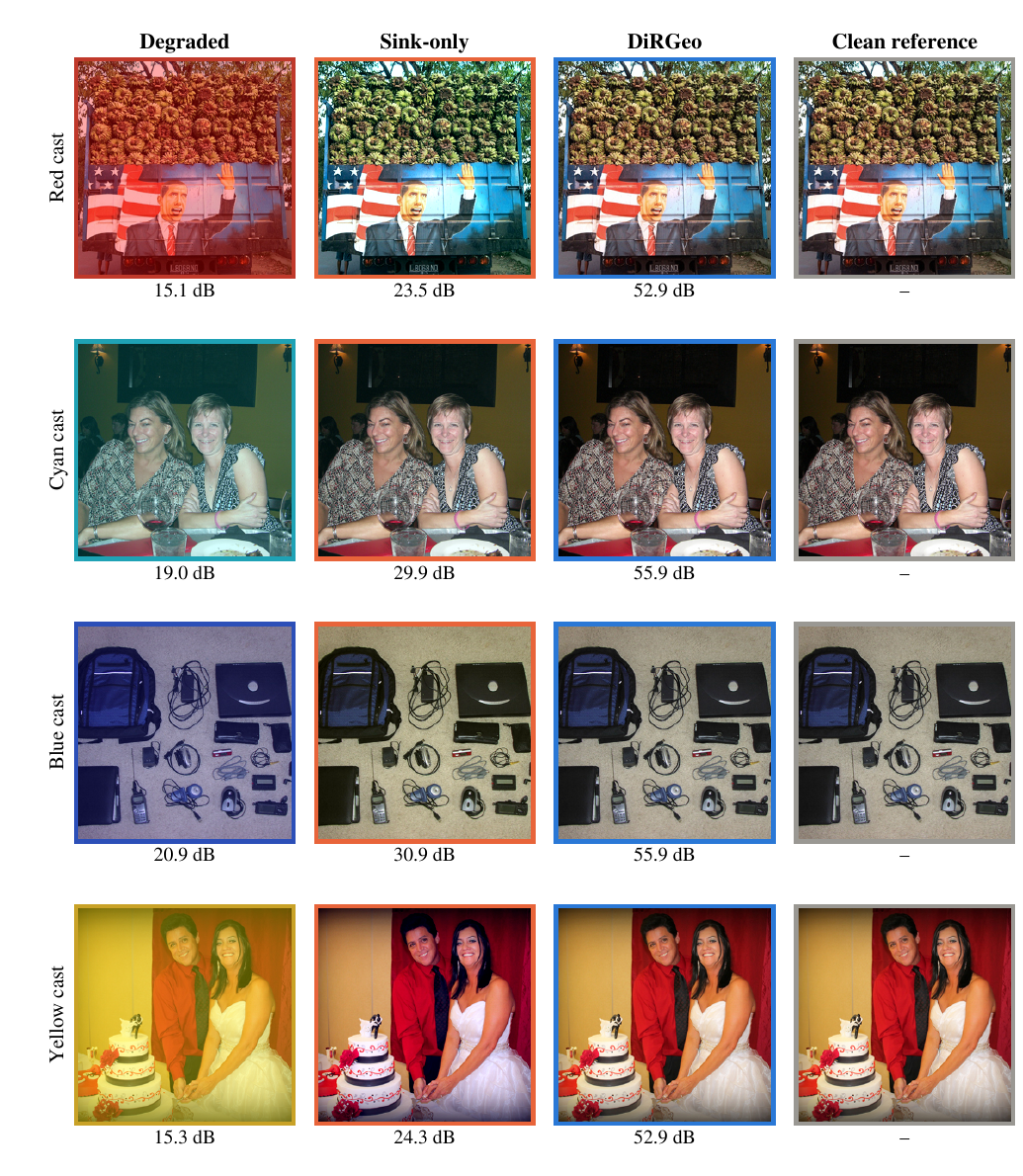}
    \caption{\textbf{Additional colour-correction examples across four degradation families.} Rows show red, cyan, blue, and yellow casts. Columns show the degraded image, correction selected by the fitted spherical sink with other-process penalties, correction selected by DiRGeo with the same penalties, and the clean reference. PSNR values are displayed beneath the first three images in each row. DiRGeo selects the more accurate correction in these examples, illustrating its ability to distinguish candidate paths beyond the fitted attractor. Clean references are used only for evaluation, not for candidate scoring.}
    \label{fig:h7_appendix_examples}
\end{figure*}

\paragraph{Qualitative examples across all four casts.}
Figure~\ref{fig:h7_appendix_examples} presents additional colour-correction examples for red, cyan, blue, and yellow casts. Each row shows the degraded image, the correction selected by the fitted sink, the correction selected by DiRGeo, and the clean reference. The PSNR values are shown beneath the corresponding images.

In the red-cast example, the degraded image has PSNR $15.1$\,dB. The fitted sink selects a correction with $23.5$\,dB, while DiRGeo selects one with $52.9$\,dB. The cyan example shows $19.0$, $29.9$, and $55.9$\,dB, respectively. For blue, the values are $20.9$, $30.9$, and $55.9$\,dB; for yellow, they are $15.3$, $24.3$, and $52.9$\,dB.

These examples illustrate cases where the learned geometry selects a correction close to the clean reference while the fitted sink selects a less accurate candidate. They are qualitative illustrations rather than estimates of average performance: the aggregate selection rates and paired uncertainty analyses provide the statistical evidence for the claimed improvement.

\paragraph{The stronger representation-level criterion is not satisfied.}
Although the pooled gain over the matched sink is positive and statistically significant, the predefined confirmation criterion requires the paired improvement to be significant in at least eight of the 10 representations.

For colour top-1 accuracy, only 6/10 representation-specific paired intervals exclude zero. The corresponding counts for selected-candidate percentile and Spearman correlation are also 6/10. Therefore, the final experiment supports a pooled advantage and positive gains across all four casts, but does not establish the stronger level of representation-wide consistency required by the confirmation criterion.

This distinction is important for interpreting the functional contribution. The learned geometry is useful for the tested colour-selection task, but the magnitude and statistical reliability of its advantage over the fitted sink depend on the chosen representation.

\subsection{Final Confirmation: Deblurring}
\label{app:h7_blur}

\paragraph{The fitted sink provides stronger overall blur ranking.}
The colour-correction findings do not automatically extend to other degradation families. We therefore evaluate the same scoring principles on the 12 deblurring candidates.

Table~\ref{tab:h7_blur_full} reports the final blur results. The learned-field score with other-process sink penalties achieves Spearman correlation $0.164$ and selected-candidate percentile $0.529$. The matched sink-only score performs better on both metrics, reaching $0.263$ and $0.559$, respectively.

The learned score has a slightly higher top-1 accuracy, $12.2\%$ versus $10.6\%$, but this improvement is statistically significant in only 3/10 representations. For Spearman correlation, the learned score does not significantly exceed the matched sink in any representation. These results do not support a general advantage of the learned field over the fitted sink for deblurring.

\begin{table*}[!htbp]
    \centering
    \small
    \setlength{\tabcolsep}{7pt}
    \begin{tabular}{lrrrrr}
        \hline
        Scorer & Top-1 (\%) & Percentile
        & Spearman [95\% CI] & Level 2 & Level 3 \\
        \hline
        Learned + sink penalties
        & 12.2 & .529 & $.164\ [.148,.178]$ & .127 & .201 \\
        Sink-only + sink penalties
        & 10.6 & .559 & $.263\ [.248,.276]$ & .259 & .266 \\
        Own learned path only
        & 5.0 & .339 & -.055 & -.095 & -.014 \\
        \hline
        NIQE
        & 5.5 & .496 & .108 & .421 & -.205 \\
        BRISQUE
        & 1.2 & .212 & -.023 & .159 & -.205 \\
        MUSIQ
        & 0.2 & .130 & -.397 & -.327 & -.467 \\
        Nearest to degraded
        & 0.0 & .617 & .351 & -- & -- \\
        Random
        & 8.3 & .500 & .000 & .000 & .000 \\
        \hline
    \end{tabular}
    \caption{\textbf{Final deblurring selection and ranking.} Twelve candidate restorations at severity levels 2 and 3. Top-1 measures exact best-candidate selection; percentile and Spearman measure ranking quality. The final two columns give severity-specific Spearman correlations. Dashes indicate values not reported separately in the experimental results.}
    \label{tab:h7_blur_full}
\end{table*}

\paragraph{Ranking quality and exact selection reveal different behavior.}
The nearest-to-degraded baseline has the highest pooled Spearman correlation, $0.351$, and selected-candidate percentile, $0.617$, among the listed Phase D scorers. However, it never selects the exact highest-PSNR candidate.

This illustrates a distinction between ranking candidates reasonably well and identifying the single optimum. A scorer can obtain positive rank correlation by favoring conservative changes, even when the best restoration requires a substantially different correction. Conversely, a method can improve top-1 accuracy without improving the full ordering.

For blur, the learned-field score performs better than BRISQUE and MUSIQ on the reported pooled ranking metrics, while comparisons with NIQE vary across severity levels. The fitted sink remains a stronger geometric reference for overall ranking in this setting.

\paragraph{Path integration is not universally preferable.}
We also compare integrated path scores with scores based on the local direction at the degraded image. The choice affects blur and colour differently.

For colour correction, replacing path integration with local direction scoring in the penalty construction produces top-1 accuracies of only $2.7\%$ for the learned field and $6.6\%$ for the sink. This suggests that accounting for the extent of correction along a candidate path is particularly important when distinguishing undercorrection, exact correction, and overcorrection.

For blur, however, local direction scoring achieves Spearman correlations of $0.416$ for the learned field and $0.407$ for the sink, exceeding their corresponding integrated-score results. Therefore, integration should not be viewed as uniformly superior across degradation processes. Its utility depends on the geometry of the candidate transformations and the type of correction being ranked.

\subsection{Earlier Tests and Exploratory Development}
\label{app:h7_development}

\paragraph{Purpose of the development experiments.}
The final scoring rule was informed by a sequence of exploratory experiments. We report these stages to make clear which observations motivated the method design and which results provide independent confirmation.

The initial experiments examined whether learned directional fields contained any useful restoration signal. Subsequent experiments compared local and integrated scoring, investigated the role of sink and residual components, and tested cross-process penalties. Phase D then evaluated the selected rule on a separate image cohort.

The exploratory results should not be interpreted as independent confirmations of the final score, particularly when later phases reuse previously computed candidate rankings.

\paragraph{Step 1: Pairwise restoration discrimination.}
The first experiment asks whether a learned field can distinguish a better restoration candidate from a worse one. Better candidates are required to improve the degraded image under PSNR, SSIM, and LPIPS, giving a controlled test of whether directional scores agree with multiple quality measures.

For blur, the learned direction score is above chance in all 10 representations across 3,171 candidate pairs. The same above-chance criterion holds for each colour cast. Table~\ref{tab:h7_pairwise} reports the blur results for the learned field, fitted sink, and sink-orthogonal residual.

The learned blur field also outperforms mismatched process fields in all 10 representations, suggesting that its directional preference is related to the degradation being corrected. However, it exceeds its own fitted sink in only three representations. This indicates that a useful restoration-direction signal exists, but much of that signal can also be captured by a simple attractor model.

\begin{table}[!htbp]
    \centering
    \small
    \setlength{\tabcolsep}{5pt}
    \begin{tabular}{lrrr}
        \hline
        Representation & Learned & Sink & Residual \\
        \hline
        CLIP L4  & 85.1 & 87.0 & 82.6 \\
        CLIP L8  & 82.4 & 87.3 & 74.8 \\
        CLIP L12 & 86.0 & 91.3 & 77.9 \\
        CLIP embedding & 85.6 & 89.8 & 75.1 \\
        DINOv2 L4  & 85.1 & 81.9 & 82.2 \\
        DINOv2 L8  & 67.0 & 68.3 & 62.0 \\
        DINOv2 L12 & 81.1 & 87.0 & 76.8 \\
        Supervised ViT L4  & 99.7 & 98.6 & 92.0 \\
        Supervised ViT L8  & 90.0 & 91.1 & 81.9 \\
        Supervised ViT L12 & 83.3 & 78.6 & 82.0 \\
        \hline
    \end{tabular}
    \caption{\textbf{Step 1 blur pairwise discrimination (\%).} Accuracy of learned-field, fitted-sink, and sink-orthogonal residual scores when comparing better and worse restoration candidates. All learned-field pointwise 95\% confidence intervals lie above the $50\%$ chance reference.}
    \label{tab:h7_pairwise}
\end{table}

\paragraph{Step 2 and Phase B: Ranking multiple candidates.}
Pairwise discrimination does not necessarily translate into reliable ranking of a larger candidate set. Step 2 therefore evaluates all 12 blur restorations per image.

The learned local direction score has positive Spearman correlation in all 10 representations, but its pooled correlation is below the fitted sink's, $0.302$ versus $0.351$. Phase B reproduces this pattern on a new evaluation cohort, yielding correlations of $0.297$ and $0.344$, respectively.

For colour correction, Phase B reveals a different opportunity. Integrating the learned field along the candidate path improves top-1 accuracy to $23.0\%$, compared with $2.5\%$ for the local direction score and $18.2\%$ for the fitted-sink path. This suggests that accumulated directional evidence is useful for distinguishing corrections of different strengths.

For blur, the unpenalized learned path score instead has negative Spearman correlation, $-0.021$. The predefined Phase B blur criteria are not met. This contrast motivates treating the restoration application as degradation-dependent rather than assuming that a single scoring construction should improve all restoration tasks.

\begin{table*}[!htbp]
    \centering
    \small
    \setlength{\tabcolsep}{9pt}
    \begin{tabular}{lrrr}
        \hline
        Scorer & Blur Spearman & Colour Spearman & Colour top-1 (\%) \\
        \hline
        Learned direction & .297 & .444 & 2.5 \\
        Sink direction    & .344 & .530 & 6.9 \\
        Learned path      & -.021 & .468 & 23.0 \\
        Sink path         & -.114 & .467 & 18.2 \\
        Nearest to degraded & .392 & -.055 & 0.0 \\
        BRISQUE & -.058 & .044 & 7.8 \\
        MUSIQ   & -.618 & .391 & 18.3 \\
        NIQE    & -.252 & .064 & 9.5 \\
        Random  & .000 & .000 & 12.5 \\
        \hline
    \end{tabular}
    \caption{\textbf{Phase B restoration ranking on an independent 200-image cohort.} Blur results use 12 candidates; colour results average four casts with eight candidates each. These measurements precede development of the other-process sink penalties and are exploratory rather than final confirmation results.}
    \label{tab:h7_phase_b}
\end{table*}

\paragraph{Phase A: Interpreting sink and residual contributions.}
Phase A reanalyzes previously saved results to understand the relative roles of common-attractor alignment and process-specific residual information.

Across colour casts, the fitted sink achieves $98.5\%$ pairwise accuracy on its own degradation process and $85.8\%$ on other casts. The residual achieves $84.5\%$ on its own process but only $39.8\%$ on other casts. For example, the red residual reaches $25.1\%$ on cyan pairs, whereas the red sink reaches $67.6\%$.

These observations suggest that fitted sinks capture a relatively broad directional tendency, while residual fields can exhibit more selective responses to particular degradation directions. They also indicate that opposing colour changes may produce competing directional preferences, motivating explicit constraints against moving toward other degradation processes.

However, the complementary-colour residual effect is not uniform across representations. It is absent in CLIP L12, the CLIP projected embedding, and DINOv2 L12. The effect therefore cannot be treated as a universal property of deep representation residuals.

\paragraph{Phase C: Developing cross-process penalties.}
Phase C reuses the Phase B scores to compare different penalty constructions. Table~\ref{tab:h7_phase_c} shows that penalties based on fitted sinks of other processes produce the strongest colour-selection results among the tested variants.

Adding other-process sink penalties increases learned-field top-1 accuracy from $23.0\%$ to $32.6\%$, while the analogous fitted-sink score reaches $28.0\%$. Penalties based on other learned fields or residual fields are less effective.

These results motivate using fitted sinks as relatively simple cross-process constraints while retaining the learned field for the own-process term. The penalty weight $\lambda=1$ and the scoring construction were then fixed for the separate Phase D confirmation.

Because Phase C operates on saved Phase B results, it is a development experiment and not independent evidence of final performance.

\begin{table}[!htbp]
    \centering
    \small
    \setlength{\tabcolsep}{4pt}
    \begin{tabular}{lrrr}
        \hline
        Own term / penalty source
        & Top-1 (\%) & Percentile & Spearman \\
        \hline
        Learned / none      & 23.0 & .607 & .468 \\
        Sink / none         & 18.2 & .580 & .467 \\
        Learned / learned   & 25.1 & .579 & .322 \\
        Learned / sinks     & 32.6 & .714 & .600 \\
        Sink / sinks        & 28.0 & .687 & .585 \\
        Learned / residuals & 21.4 & .523 & .204 \\
        \hline
    \end{tabular}
    \caption{\textbf{Exploratory Phase C colour-correction comparisons.} Alternative own-process scores and cross-process penalties evaluated using previously saved Phase B results. The penalty weight is $\lambda=1$. These results motivate the final scoring rule but are not independent confirmation measurements.}
    \label{tab:h7_phase_c}
\end{table}

\subsection{Confirmation Outcome and Interpretation}
\label{app:h7_conclusion}

\paragraph{The pooled colour improvement is supported, but the stronger confirmation criterion is not met.}
The final Phase D results support a positive pooled advantage of the learned directional field over the fitted sink for colour correction. The paired top-1 improvement is $3.98$ percentage points, and its 95\% confidence interval $[2.76,5.14]$ lies above zero. Positive pooled gains also occur across all four colour casts.

Nevertheless, only six of the 10 representation-specific paired intervals exclude zero. This falls short of the predefined eight-representation criterion. The results therefore support the more limited statement that the learned field improves pooled colour-correction selection, with benefits that vary across frozen representations.

By contrast, the benefit of adding other-process sink penalties to the learned own-process score is supported in all 10 representations. This indicates that the cross-process constraint is a more consistent contributor to colour-selection performance than the incremental advantage of the learned own-process field over its fitted-sink approximation.

Table~\ref{tab:h7_verdict} summarizes the final comparisons and their representation-level confirmation counts.

\begin{table*}[!htbp]
    \centering
    \small
    \setlength{\tabcolsep}{8pt}
    \begin{tabular}{llrr}
        \hline
        Setting and comparison & Metric & Pooled gain
        & Representations with CI $>0$ \\
        \hline
        Colour: learned vs.\ sink-only
        & Top-1 & $+3.98$ points & 6/10 \\
        & Percentile & $+0.024$ & 6/10 \\
        & Spearman & $+0.015$ & 6/10 \\
        Colour: penalties vs.\ own learned path
        & Top-1 & $+10.46$ points & 10/10 \\
        \hline
        Colour: learned vs.\ BRISQUE
        & Top-1 / Spearman & $+22.7$ / $+0.561$ & 10/10 / 10/10 \\
        Colour: learned vs.\ MUSIQ
        & Top-1 / Spearman & $+13.6$ / $+0.237$ & 8/10 / 10/10 \\
        Colour: learned vs.\ NIQE
        & Top-1 / Spearman & $+23.1$ / $+0.530$ & 10/10 / 10/10 \\
        \hline
        Blur: learned vs.\ sink-only
        & Top-1 & $+1.6$ points & 3/10 \\
        & Percentile & $-0.030$ & 2/10 \\
        & Spearman & $-0.099$ & 0/10 \\
        \hline
    \end{tabular}
    \caption{\textbf{Final restoration-selection confirmation outcomes.} A representation-level criterion is satisfied only when at least 8/10 paired confidence intervals lie above zero. The learned and fitted-sink scores use identical other-process penalties unless stated otherwise. Top-1 gains are percentage points; percentile and Spearman gains are differences in their respective metrics.}
    \label{tab:h7_verdict}
\end{table*}

\paragraph{Functional interpretation of directional geometry.}
The restoration experiments provide a controlled test of whether a learned directional ruler can support decisions beyond the geometric characterization tasks used to train and evaluate it. The learned fields are not optimized on restoration targets and receive no clean-image supervision during candidate scoring.

For colour correction, the field provides useful information about movement against a known degradation process. Integrating this directional information along candidate paths, while discouraging movement toward other degradation attractors, improves selection of the highest-PSNR correction. The qualitative examples in Fig.~\ref{fig:h7_appendix_examples} and the decision statistics in Fig.~\ref{fig:h7_appendix_stats} illustrate how this improvement manifests in individual corrections and aggregate candidate choices.

For blur, however, the fitted sink explains much of the useful ranking signal, and the learned field does not consistently improve over it. This shows that the value of additional state-dependent directional structure depends on the degradation and the candidate distribution.

The supported conclusion is therefore not that DiRGeo is a generally superior image restoration method, but that learned directional geometry can provide functional selection information beyond a fitted common attractor in a controlled colour-correction task.

\paragraph{Scope and limitations.}
The evaluation assumes that the degradation family is known and that candidate corrections have already been generated. Colour-correction candidates include the exact inverse of the synthetic degradation, while the blur candidates are restricted to a fixed set of classical methods and two moderate severity levels. These choices make the evaluation interpretable but do not constitute a general restoration benchmark.

The method also does not determine the degradation type, synthesize a corrected image, or guarantee that the highest-scored candidate has optimal perceptual quality. Its output is a geometric ranking among the provided alternatives.

Within this scope, the results establish initial functional evidence for the learned directional ruler. They also identify conditions under which the simpler fitted-attractor approximation remains competitive, providing a more precise understanding of when process-dependent directional geometry contributes useful information.



\section{Additional Analysis of Generalization and Object-Category Dependence}
\label{app:generalization}

The preceding experiments establish that DiRGeo learns reproducible directional preferences across visual processes and pretrained representation spaces. We now investigate how those preferences generalize beyond the data on which the directional fields were learned, and whether their dependence on the training distribution changes with representation depth.

Generalization has several distinct meanings in this setting. A field may generalize to new images undergoing the same process, but depend on the object categories represented during training. It may transfer poorly to another visual process, even when both processes manipulate the same image content. Its preferences may also change across early and late stages of a trajectory. These properties should not be conflated.

We examine four settings: unseen source images, replication across encoder families, exclusion of object categories during field training, and transfer across process families and trajectory stages. The primary new analysis uses five paired category splits. It measures whether removing particular object categories from training disproportionately reduces directional discrimination on those categories, while evaluating the compared fields on identical test images.

The results reveal a depth-dependent pattern. In deep representations, the \textsc{Insert}, \textsc{Grow}, and \textsc{Enter} fields exhibit substantially greater sensitivity to category exclusion than their matched controls. However, the predefined criterion based only on the sign of this effect does not distinguish object-level processes from controls. We therefore report both the planned criterion and the observed effect magnitudes, while distinguishing evidence of category-associated dependence from proof of exclusively semantic representations.

\subsection{Generalization to Unseen Images and Across Encoder Families}
\label{app:h5_images_encoders}

\paragraph{Generalization to unseen source images.}
All primary directional-discrimination experiments use source images excluded from the corresponding field-training sets. For global appearance transformations, the training and evaluation partitions contain 8,000 and 2,000 source images, respectively. For object-level processes and matched controls, the corresponding partitions contain 800 and 200 images.

All transformations derived from a source image remain within its assigned partition. Consequently, the held-out evaluations do not simply test whether a field recognizes another transformed version of an image encountered during training.

As shown in Appendix~\ref{app:directional_existence}, reverse and shuffled accuracy exceed chance in all 170 family--representation combinations, with every pointwise 95\% confidence interval above $50\%$. The corresponding whole-path asymmetry is also positive across all combinations. These findings establish generalization to unseen source images under the same specified visual processes.

This conclusion is intentionally narrower than transfer to arbitrary visual changes. The training and test trajectories are generated using the same process constructions, even though their source images are disjoint. We examine transfer to excluded process families separately below.

\paragraph{Replication across encoder families.}
DiRGeo is evaluated on three frozen encoders with different pretraining objectives: CLIP, trained using image--text contrastive supervision; DINOv2, trained through image-only self-supervision; and a supervised ViT trained for image classification. We extract features from transformer blocks 4, 8, and 12, together with CLIP's projected image embedding.

The directional findings replicate across these encoder families, indicating that orientation-dependent structure is not unique to one particular representation-learning objective. However, a separate wind network is trained for every encoder representation. The experiments therefore establish replication of the geometric phenomenon across representations, not transfer of a single trained field between different encoder spaces.

This distinction matters because the feature dimensions, coordinate systems, and learned representation organizations need not agree across encoders. Direct cross-encoder transfer would require an additional alignment or transport mechanism, which is not part of the present evaluation.

\subsection{Paired Object-Category Exclusion Protocol}
\label{app:h5_category_protocol}

\paragraph{Motivation and experimental question.}
Directional preferences may depend on the object content represented during training. For example, a field learned from progressive object insertion might capture visual changes that occur across many object categories, or it might rely partly on category-associated appearance and representation structure.

A simple comparison between seen-category and unseen-category accuracy is insufficient to distinguish these possibilities. Different images may have different intrinsic levels of directional-discrimination difficulty, even when evaluated using the same field. Lower accuracy on one category group does not by itself establish that excluding those categories from training caused the difference.

We therefore use a paired category-exclusion experiment. Its central question is whether training without a group of object categories causes a greater loss of directional discrimination on those categories than on the categories retained in training.

\paragraph{Five category splits and training conditions.}
Each of five splits partitions the 80 COCO categories into 40 seen and 40 unseen categories. Split 0 reproduces the category partition used in the initial single-split experiment.

For each of the nine object-dataset families, we compare two field-training conditions. The all-category field is trained using the original object-level training distribution, while the seen-category field is trained using only training images belonging to the retained categories for the corresponding split.

Both fields are evaluated on the same held-out images. Evaluation images are grouped according to whether their relevant category belongs to the retained or excluded category set. We repeat the comparison for the four object-level processes and five matched controls across all 10 encoder representations.

The all-category field is the saved reference model, while the category-restricted field is trained separately for each split. Thus, the comparison measures the effect of changing category coverage in the training data while holding the evaluation images fixed within each category group.

\paragraph{Difference-in-differences measurement.}
For category split $s$, let $A_{\mathrm{all},s}^{U}$ and $A_{\mathrm{seen},s}^{U}$ denote shuffled accuracy on unseen-category evaluation images for the all-category and seen-category-only fields, respectively. The superscript $U$ identifies evaluation images from excluded categories.

Similarly, let $A_{\mathrm{all},s}^{S}$ and $A_{\mathrm{seen},s}^{S}$ denote the corresponding accuracies on retained-category images, identified by superscript $S$.

We first measure the performance reduction caused by category-restricted training separately for the two evaluation groups:
\begin{align}
    g_s^{U}
    &=
    A_{\mathrm{all},s}^{U}
    -
    A_{\mathrm{seen},s}^{U},\\
    g_s^{S}
    &=
    A_{\mathrm{all},s}^{S}
    -
    A_{\mathrm{seen},s}^{S}.
\end{align}
We then define the category-exclusion effect as
\begin{equation}
    D_{\mathrm{cat},s}
    =
    g_s^{U}-g_s^{S}.
    \label{eq:h5_category_did}
\end{equation}
All accuracies are computed using sphere-tangent shuffled discrimination and expressed in percentage points.

A positive $D_{\mathrm{cat},s}$ means that restricting training to the retained categories produces a greater accuracy loss on excluded-category evaluation images than on retained-category images. For example, if category exclusion reduces unseen-category accuracy by 10 points but seen-category accuracy by 2 points, the resulting effect is $D_{\mathrm{cat},s}=8$ points.

This difference-in-differences construction is more informative than comparing raw seen- and unseen-category accuracies. Within each evaluation group, the all-category and restricted fields are tested on identical images, controlling for differences in test-image difficulty between those two models. Subtracting the seen-category loss further accounts for degradation shared by the two evaluation groups.

Nevertheless, this contrast is not a causal identification of semantic information in isolation. Category identity can correlate with object shape, texture, appearance, context, and other visual properties. The measurement therefore characterizes category-associated sensitivity of learned directional geometry, rather than separating semantic content from every accompanying visual difference.

\paragraph{Split aggregation and statistical criteria.}
For each family--representation combination, we report the mean $D_{\mathrm{cat},s}$ across the five splits, together with the minimum and maximum split values. These ranges describe variation across category partitions and are not confidence intervals.

We also report two separate counts: the number of splits with positive point estimates and the number of splits whose pointwise 95\% confidence intervals lie entirely above zero.

The predefined category-effect criterion requires both a positive mean effect across the five splits and a positive point estimate in at least four of the five splits. This rule evaluates consistency of the effect's direction; it does not require a particular effect magnitude or statistical significance in four splits.

Because the category partitions reuse the underlying source-image dataset, the five splits are not independent experimental datasets. We therefore use them to assess consistency across category partitions rather than interpreting them as independent replications.

\paragraph{Implementation consistency checks.}
The five-split analysis includes checks that the saved and retrained fields reproduce their corresponding earlier evaluations. Split 0 matches the original category partition. Loaded all-category fields reproduce the saved per-image accuracies exactly.

These checks support consistency between the initial single-split analysis and the expanded five-split experiment.

\subsection{Category Dependence Across Representation Depth}
\label{app:h5_depth}

\paragraph{Category-exclusion effects are small in shallow representations.}
Table~\ref{tab:h5_all_means} reports the mean category-exclusion effect for every object-dataset family and encoder representation.

At L4 and L8, the mean effects of all nine families lie between $-1.0$ and $+1.9$ percentage points. Thus, excluding half of the object categories from training generally produces little additional loss on the excluded categories relative to the retained categories in these shallower representations.

This does not mean that the learned fields are unaffected by category restriction. Both seen- and unseen-category accuracies may change. Rather, their changes are similar enough that the difference-in-differences remains small.

\paragraph{Deep representations show stronger category-associated sensitivity.}
The pattern changes in the deepest evaluated representations: CLIP L12, CLIP's projected embedding, DINOv2 L12, and supervised ViT L12.

For \textsc{Insert}, \textsc{Grow}, and \textsc{Enter}, the mean category-exclusion effects range from $5.0$ to $14.0$ percentage points. These values are substantially larger than the corresponding shallow-layer effects.

For example, the \textsc{Insert} field has effects of $6.4$ points at CLIP L12, $6.5$ in the CLIP embedding, $14.0$ at DINOv2 L12, and $12.2$ at supervised ViT L12. The pattern is therefore observed under all three encoder pretraining objectives, although its magnitude differs considerably.

The same deep representations show smaller effects for the five matched controls, whose means range from $-0.6$ to $4.2$ points. Every one of the 12 deep \textsc{Insert}, \textsc{Grow}, and \textsc{Enter} mean effects exceeds every matched-control mean across these four representations.

This is the strongest observed distinction between the real-object processes and their controls. It suggests that the directional fields for insertion, growth, and entry become more dependent on the categories represented during training as representation depth increases.

\begin{table*}[!htbp]
    \centering
    \small
    \setlength{\tabcolsep}{3.5pt}
    \begin{tabular}{lrrrrrrrrrr}
        \hline
        & \multicolumn{4}{c}{CLIP}
        & \multicolumn{3}{c}{DINOv2}
        & \multicolumn{3}{c}{Supervised ViT} \\
        Family & L4 & L8 & L12 & Emb.
               & L4 & L8 & L12 & L4 & L8 & L12 \\
        \hline
        \multicolumn{11}{l}{\textit{Object-level processes}} \\
        Insert
        & 1.5 & 0.3 & 6.4 & 6.5
        & 1.3 & 1.6 & 14.0 & 1.2 & 1.6 & 12.2 \\
        Grow
        & 0.6 & -1.0 & 7.3 & 5.9
        & 1.0 & 1.4 & 13.9 & 0.4 & 1.9 & 10.6 \\
        Enter
        & -0.3 & 0.1 & 5.2 & 5.0
        & -0.1 & 0.9 & 9.7 & -0.1 & 1.1 & 7.8 \\
        Colour
        & -0.2 & 0.3 & 2.4 & 2.7
        & 0.1 & -0.1 & 4.1 & -0.4 & 1.5 & 3.4 \\
        \hline
        \multicolumn{11}{l}{\textit{Matched controls}} \\
        Insert blob
        & 1.1 & 0.0 & 1.0 & 1.5
        & 0.4 & 0.6 & 2.2 & 0.3 & 0.4 & 2.9 \\
        Insert scramble
        & 0.2 & -0.8 & 1.4 & 1.2
        & 0.8 & 0.2 & 4.2 & 1.2 & 1.2 & 2.3 \\
        Grow blob
        & 0.0 & -0.1 & 0.2 & 0.5
        & -0.1 & 0.4 & 2.9 & 0.1 & 0.7 & 3.0 \\
        Enter blob
        & 1.1 & 0.0 & 1.6 & 1.2
        & 0.4 & 0.1 & 1.8 & 0.0 & 0.7 & 3.1 \\
        Insert blur
        & 0.2 & -0.4 & -0.6 & 0.2
        & 0.1 & 0.0 & 1.3 & 0.5 & -0.3 & 1.1 \\
        \hline
    \end{tabular}
    \caption{\textbf{Mean category-exclusion effect across representations.} Five-split mean difference-in-differences $D_{\mathrm{cat}}$ in sphere-tangent shuffled-accuracy percentage points. Positive values indicate greater accuracy loss on excluded categories after subtracting the loss on retained categories. All nine object-dataset families are included. Values shown as zero are rounded.}
    \label{tab:h5_all_means}
\end{table*}

\paragraph{The deep effects are consistent across category partitions.}
Table~\ref{tab:h5_split_ranges} reports the five-split means and observed split ranges for the four deepest representations. For \textsc{Insert}, \textsc{Grow}, and \textsc{Enter}, all 60 individual split effects are positive.

Moreover, 58 of their 60 pointwise 95\% confidence intervals lie above zero. The only unresolved cases are one \textsc{Enter} split at CLIP L12 and one at the CLIP embedding. These results show that the observed deep category dependence is not driven solely by one favorable category partition.

The evidence is weaker for \textsc{Colour}. All 20 deep split effects are positive, but only seven confidence intervals exclude zero. Its mean effects range from $2.4$ to $4.1$ points, overlapping the range observed among matched controls.

Thus, the category-associated dependence is strongest for the three processes involving object insertion, movement into the scene, and growth. We do not interpret it as equally strong for every object-level process.

\begin{table*}[!htbp]
    \centering
    \small
    \setlength{\tabcolsep}{7pt}
    \begin{tabular}{lrrrr}
        \hline
        Family & CLIP L12 & CLIP embedding
               & DINOv2 L12 & Supervised ViT L12 \\
        \hline
        Insert
        & $6.4\ [5.8,7.7]$
        & $6.5\ [5.8,7.3]$
        & $14.0\ [12.2,16.6]$
        & $12.2\ [8.9,16.6]$ \\
        Grow
        & $7.3\ [4.6,10.5]$
        & $5.9\ [3.3,10.2]$
        & $13.9\ [12.0,16.2]$
        & $10.6\ [7.6,15.5]$ \\
        Enter
        & $5.2\ [2.3,7.3]$
        & $5.0\ [2.1,6.6]$
        & $9.7\ [7.8,11.9]$
        & $7.8\ [5.5,10.1]$ \\
        Colour
        & $2.4\ [1.3,5.4]$
        & $2.7\ [1.8,3.2]$
        & $4.1\ [2.9,5.5]$
        & $3.4\ [1.0,6.9]$ \\
        \hline
        Insert blob
        & $1.0\ [-0.5,2.8]$
        & $1.5\ [-0.4,2.9]$
        & $2.2\ [0.7,3.0]$
        & $2.9\ [1.5,5.2]$ \\
        Insert scramble
        & $1.4\ [0.1,2.9]$
        & $1.2\ [-2.1,3.8]$
        & $4.2\ [2.4,8.2]$
        & $2.3\ [0.5,3.4]$ \\
        Grow blob
        & $0.2\ [-2.0,1.9]$
        & $0.5\ [-0.4,1.0]$
        & $2.9\ [0.4,4.9]$
        & $3.0\ [-0.6,6.7]$ \\
        Enter blob
        & $1.6\ [0.7,2.3]$
        & $1.2\ [0.2,2.7]$
        & $1.8\ [-0.2,4.1]$
        & $3.1\ [1.6,5.4]$ \\
        Insert blur
        & $-0.6\ [-2.0,1.1]$
        & $0.2\ [-1.9,1.5]$
        & $1.3\ [-1.8,5.5]$
        & $1.1\ [-2.7,3.9]$ \\
        \hline
    \end{tabular}
    \caption{\textbf{Category-exclusion effects in the deepest representations.} Each entry reports the five-split mean $D_{\mathrm{cat}}$ followed by the minimum and maximum split values, in shuffled-accuracy percentage points. The brackets describe observed variation across category partitions, not confidence intervals.}
    \label{tab:h5_split_ranges}
\end{table*}

\begin{table}[!htbp]
    \centering
    \small
    \setlength{\tabcolsep}{4pt}
    \begin{tabular}{lrrrr}
        \hline
        Family & CLIP L12 & Emb. & DINO L12 & ViT L12 \\
        \hline
        Insert & 5/5 & 5/5 & 5/5 & 5/5 \\
        Grow   & 5/5 & 5/5 & 5/5 & 5/5 \\
        Enter  & 5/4 & 5/4 & 5/5 & 5/5 \\
        Colour & 5/1 & 5/1 & 5/3 & 5/2 \\
        \hline
        Insert blob     & 4/1 & 4/1 & 5/0 & 5/1 \\
        Insert scramble & 5/1 & 3/1 & 5/3 & 5/2 \\
        Grow blob       & 3/0 & 4/0 & 5/2 & 4/2 \\
        Enter blob      & 5/0 & 5/1 & 4/1 & 5/3 \\
        Insert blur     & 1/0 & 4/0 & 3/1 & 4/1 \\
        \hline
    \end{tabular}
    \caption{\textbf{Consistency and statistical support across five category splits.} Each entry is $n_{+}/n_{\mathrm{CI}}$, where $n_{+}$ is the number of positive split effects and $n_{\mathrm{CI}}$ is the number of pointwise 95\% confidence intervals lying entirely above zero. Both counts are out of five; the slash separates two counts, rather than indicating a fraction.}
    \label{tab:h5_split_support}
\end{table}

\subsection{Interpretation of the Category-Dependence Evidence}
\label{app:h5_interpretation}

\paragraph{The predefined sign criterion does not separate object-level processes from controls.}
The predefined category-effect criterion requires a positive mean across five splits and positive effects in at least four splits. Table~\ref{tab:h5_sign_criterion} summarizes how often this criterion is satisfied.

In the four deepest representations, all 16 object-level family--representation combinations satisfy the criterion. However, 16 of the 20 matched-control combinations also satisfy it. Across all 10 representations, the corresponding counts are 26/40 for object-level processes and 25/50 for matched controls.

Thus, the predefined sign-based criterion does not establish a reliable separation between the two groups. A consistent positive category-exclusion effect can also occur for synthetic controls whose manipulated content does not preserve intact real-object appearance.

This result is important because it prevents us from interpreting any positive category effect as sufficient evidence of semantic dependence.

\begin{table}[!htbp]
    \centering
    \small
    \setlength{\tabcolsep}{5pt}
    \begin{tabular}{lrr}
        \hline
        Representations & Object-level & Controls \\
        \hline
        Deepest four & 16/16 & 16/20 \\
        L4           & 6/12  & 6/15 \\
        All ten      & 26/40 & 25/50 \\
        \hline
    \end{tabular}
    \caption{\textbf{Cases satisfying the predefined category-effect criterion.} A case passes when its five-split mean $D_{\mathrm{cat}}$ is positive and at least four of five split effects are positive. The criterion measures sign consistency, not effect magnitude or statistical significance. Matched controls also frequently satisfy it.}
    \label{tab:h5_sign_criterion}
\end{table}

\paragraph{Effect magnitude provides a stronger, descriptive distinction.}
Although the sign criterion does not distinguish the groups, the observed effect magnitudes reveal a substantially different pattern.

Across the four deepest representations, the 12 mean effects for \textsc{Insert}, \textsc{Grow}, and \textsc{Enter} range from $5.0$ to $14.0$ percentage points. The corresponding means for the five matched controls range from $-0.6$ to $4.2$ points.

Every deep mean for these three object-level processes exceeds every deep matched-control mean. Moreover, their split-specific confidence intervals exclude zero in 58/60 cases, compared with 21/100 for the controls.

This suggests that the learned fields for these three real-object processes depend more strongly on category-associated information in deep representations than the matched-control fields do.

However, this magnitude comparison was not the predefined category-effect criterion. We therefore present it as an observed descriptive separation rather than claiming that the predefined test established a semantic distinction.

\paragraph{Category dependence is not identical to semantic understanding.}
The stronger deep effects are compatible with deeper representations encoding object-category-related information that influences their directional preferences. Nevertheless, object-category identity is associated with multiple visual properties, and excluding categories changes the distribution of appearances available during training.

The experiment does not determine which underlying factors are responsible for the observed effect. In particular, it cannot establish that the field depends exclusively on semantic category identity independently of shape, texture, context, or other correlated image properties.

The supported conclusion is that deeper object-level directional fields exhibit stronger category-associated sensitivity under the tested training exclusions.

\paragraph{Path-level category effects provide complementary evidence.}
We additionally apply the same difference-in-differences construction to whole-path asymmetry $R/\rho$. Rather than measuring the proportion of correctly preferred shuffled directions, this analysis measures whether category exclusion disproportionately changes the accumulated forward--reverse cost difference on excluded-category trajectories.

Table~\ref{tab:h5_path_effects} reports the five-split mean path-level effects in the deepest representations. All reported means are positive.

For \textsc{Insert}, \textsc{Grow}, and \textsc{Enter}, the effects range from $0.0122$ to $0.0399$. For matched controls, they range from $0.0041$ to $0.0211$. Unlike shuffled accuracy, these ranges overlap, so whole-path asymmetry provides weaker separation between object-level and control families.

We therefore treat path-level category effects as complementary evidence rather than as the primary basis for claiming stronger category dependence. The main category finding rests on the paired sphere-tangent shuffled-accuracy analysis.

\begin{table}[!htbp]
    \centering
    \small
    \setlength{\tabcolsep}{3.5pt}
    \begin{tabular}{lrrrr}
        \hline
        Family & CLIP L12 & Emb. & DINO L12 & ViT L12 \\
        \hline
        Insert & .0328 & .0358 & .0399 & .0335 \\
        Grow   & .0226 & .0241 & .0350 & .0254 \\
        Enter  & .0152 & .0186 & .0211 & .0122 \\
        Colour & .0271 & .0315 & .0300 & .0253 \\
        \hline
        Insert blob     & .0099 & .0095 & .0088 & .0157 \\
        Insert scramble & .0154 & .0151 & .0191 & .0211 \\
        Grow blob       & .0080 & .0086 & .0066 & .0111 \\
        Enter blob      & .0084 & .0093 & .0044 & .0075 \\
        Insert blur     & .0044 & .0041 & .0043 & .0055 \\
        \hline
    \end{tabular}
    \caption{\textbf{Category-exclusion effects on whole-path asymmetry.} Five-split mean difference-in-differences for $R/\rho$ in the deepest representations. Positive values indicate a greater reduction in accumulated forward preference on excluded-category trajectories. These values are dimensionless asymmetry differences, not accuracy percentages.}
    \label{tab:h5_path_effects}
\end{table}

\subsection{Absolute Performance on Excluded Categories}
\label{app:h5_absolute}

\paragraph{Category sensitivity and absolute generalization answer different questions.}
The paired difference-in-differences measures the additional loss caused by category exclusion. It does not directly tell us how accurately the category-restricted field performs on excluded-category images.

For example, a large positive category effect may coexist with above-chance discrimination on excluded categories. Conversely, a small category effect does not guarantee strong absolute discrimination, because both all-category and restricted fields may perform poorly.

We therefore report the absolute unseen-category accuracies from the original single-split experiment as complementary evidence.

\paragraph{Most deep category-restricted fields retain above-chance discrimination.}
Table~\ref{tab:h5_unseen_accuracy} reports sphere-tangent shuffled accuracy on excluded-category test images after training only on retained categories.

For \textsc{Insert}, the four deep representation accuracies range from $56.0\%$ to $61.2\%$, with every reported pointwise 95\% confidence interval above chance. The corresponding \textsc{Grow} accuracies range from $53.8\%$ to $58.5\%$, also with intervals above chance.

For \textsc{Enter}, the results are less uniform. CLIP L12 and its projected embedding remain above chance, while DINOv2 L12 reaches $51.1\%$ with a 95\% confidence interval of $[48.6,53.5]$, which includes chance. Supervised ViT L12 reaches $52.6\%$ with interval $[50.4,54.7]$.

The \textsc{Colour} field retains relatively strong unseen-category discrimination, ranging from $60.5\%$ to $71.6\%$ across these representations, despite its smaller paired category-exclusion effects.

Together, these results indicate that category restriction often reduces directional discrimination without eliminating it. Generalization to excluded categories remains measurable, but is not uniformly significant for every process and representation.

\begin{table*}[!htbp]
    \centering
    \small
    \setlength{\tabcolsep}{8pt}
    \begin{tabular}{lrrrr}
        \hline
        Family & CLIP L12 & CLIP embedding
               & DINOv2 L12 & Supervised ViT L12 \\
        \hline
        Insert
        & $61.2\ [57.5,64.6]$
        & $59.5\ [55.4,63.4]$
        & $56.5\ [52.9,60.2]$
        & $56.0\ [52.7,58.9]$ \\
        Grow
        & $57.6\ [54.7,60.5]$
        & $58.5\ [55.3,61.6]$
        & $53.8\ [50.9,56.8]$
        & $54.1\ [51.2,57.0]$ \\
        Enter
        & $57.2\ [54.6,60.0]$
        & $59.2\ [56.2,62.1]$
        & $51.1\ [48.6,53.5]$
        & $52.6\ [50.4,54.7]$ \\
        Colour
        & $69.1\ [65.4,73.1]$
        & $68.2\ [64.7,71.7]$
        & $60.5\ [57.2,64.0]$
        & $71.6\ [67.8,75.7]$ \\
        \hline
    \end{tabular}
    \caption{\textbf{Absolute directional discrimination on excluded categories.} Sphere-tangent shuffled accuracy (\%) of fields trained only on retained categories in the original single-split experiment. Brackets are pointwise 95\% confidence intervals, unlike the split ranges in Table~\ref{tab:h5_split_ranges}.}
    \label{tab:h5_unseen_accuracy}
\end{table*}

\paragraph{Relation to the five-split paired analysis.}
The original single-split evaluation compares different source-image groups to measure seen- and unseen-category performance. Its absolute accuracies provide useful context, but a raw accuracy difference between those groups may reflect their different visual difficulty.

The five-split paired analysis provides stronger evidence about the effect of category exclusion because all-category and category-restricted fields are evaluated on identical images within each group. For \textsc{Insert}, \textsc{Grow}, and \textsc{Enter}, the paired deep effects are at least as large as the corresponding earlier unpaired gaps in all 12 deep configurations.

For example, the DINOv2 L12 \textsc{Insert} field has a paired category-exclusion effect of $14.0$ percentage points, compared with an earlier unpaired seen--unseen gap of $12.9$ points. This reinforces the conclusion that the deep effect is not explained solely by comparing intrinsically different test-image groups.

\subsection{Transfer to Unseen Process Families and Trajectory Stages}
\label{app:h5_additional_transfer}

\paragraph{Transfer across process families.}
Generalization to new images under a known process is different from transfer to an unobserved process. To evaluate the latter, we train directional fields after excluding one process family and test them on trajectories from that family.

For global appearance transformations, the held-out-family field is trained using the other seven transformation families. For object-level processes, the corresponding field is trained using the other three object-level process families.

We compare this field with the field trained specifically for the target process, using identical test trajectories and sphere-tangent shuffled discrimination. The reported transfer gap is
\begin{equation}
    \Delta_{\mathrm{process}}
    =
    \mathrm{Acc}_{\mathrm{target}}
    -
    \mathrm{Acc}_{\mathrm{excluded}},
    \label{eq:h5_process_transfer}
\end{equation}
where $\mathrm{Acc}_{\mathrm{target}}$ denotes accuracy of the target-family-trained field and $\mathrm{Acc}_{\mathrm{excluded}}$ denotes accuracy of the field trained without the target family. Positive values indicate that access to target-process training improves discrimination.

Table~\ref{tab:h5_transfer_ranges} reports the minimum and maximum gaps across the 10 representations.

\paragraph{Transfer varies substantially across visual processes.}
For \textsc{Insert}, \textsc{Grow}, and \textsc{Enter}, the held-out-family gaps range from $-0.4$ to $7.9$ percentage points across representations. These relatively modest gaps indicate that fields learned from the remaining object-level processes can retain some ability to discriminate motion from an excluded process.

In contrast, excluding \textsc{Colour} produces gaps from $17.1$ to $29.3$ points. This is consistent with the field-similarity results in Appendix~\ref{app:process_specificity}, where the \textsc{Colour} field is nearly orthogonal to the \textsc{Insert} field even in shallow representations.

For global appearance transformations, excluding the target family also produces substantial, representation-dependent losses. The reported gap ranges extend as high as $45.9$ points for noise and $43.1$ points for red tint.

These results show that transfer across processes is not uniform. A field learned from other processes may partially capture some shared directional changes, but certain transformation families require substantially more process-specific supervision.

\paragraph{Transfer across trajectory stages.}
The final experiment asks whether a field learned from earlier stages of a visual transformation generalizes to later stages of the same process.

We compare a field trained using frames 0--4 with one trained using frames 5--9. Both are evaluated on later trajectory stages. We define the stage-transfer gap as
\begin{equation}
    \Delta_{\mathrm{stage}}
    =
    \mathrm{Acc}_{\mathrm{late-trained}}
    -
    \mathrm{Acc}_{\mathrm{early-trained}},
    \label{eq:h5_stage_transfer}
\end{equation}
where positive values indicate an advantage for training directly on the evaluated later stages.

This comparison tests whether directional preferences learned from one region of the process trajectory remain predictive in another. Importantly, trajectory stage has different meanings across processes: it may correspond to blur severity, colour strength, object visibility, object size, or object position.

\paragraph{Later-stage transfer is process-dependent.}
For global appearance transformations, all reported stage-transfer gaps are positive, ranging from $3.0$ to $46.5$ percentage points. Thus, fields trained on later transformation states consistently outperform fields trained on earlier states when evaluated on the later stages.

For object-level processes, the pattern is more varied. The \textsc{Insert} gaps range from $8.2$ to $27.4$ points, while \textsc{Colour} ranges from $14.7$ to $37.5$ points. These processes show substantial advantages from training on the evaluated stages.

By contrast, \textsc{Grow} gaps range from $-0.3$ to $7.2$ points, indicating substantially better transfer from early to later stages. \textsc{Enter} shows intermediate gaps from $1.4$ to $10.5$ points.

The results support state- and process-dependent directional geometry: directional preferences can change along a transformation trajectory, and the magnitude of this change varies with the visual process.

They do not, however, establish that all later states are intrinsically more difficult. The experiment measures a training-distribution mismatch between earlier and later trajectory stages, not a universal ordering of stage difficulty.

\begin{table}[!htbp]
    \centering
    \small
    \setlength{\tabcolsep}{5pt}
    \begin{tabular}{lrr}
        \hline
        Family & Held-out family gap & Later-stage gap \\
        \hline
        Blur        & $[9.7,39.1]$  & $[9.2,46.5]$ \\
        Noise       & $[9.4,45.9]$  & $[3.0,25.9]$ \\
        JPEG        & $[12.2,34.0]$ & $[12.4,36.1]$ \\
        Brightness  & $[5.8,35.6]$  & $[10.1,36.0]$ \\
        Red tint    & $[2.6,43.1]$  & $[6.1,25.8]$ \\
        Cyan tint   & $[2.4,35.6]$  & $[10.3,37.4]$ \\
        Blue tint   & $[2.5,36.4]$  & $[8.6,32.0]$ \\
        Yellow tint & $[2.5,39.2]$  & $[6.9,29.1]$ \\
        \hline
        Insert & $[1.8,7.9]$   & $[8.2,27.4]$ \\
        Grow   & $[-0.4,4.5]$  & $[-0.3,7.2]$ \\
        Enter  & $[1.0,4.0]$   & $[1.4,10.5]$ \\
        Colour & $[17.1,29.3]$ & $[14.7,37.5]$ \\
        \hline
    \end{tabular}
    \caption{\textbf{Transfer across process families and trajectory stages.} Shuffled-accuracy gaps in percentage points. The held-out-family gap compares a target-family-trained field with a field trained without that family. The later-stage gap compares late-stage-trained and early-stage-trained fields on later-stage evaluation data. Brackets report the minimum and maximum across 10 representations, not confidence intervals. Positive values favor training on the target process or target trajectory stage.}
    \label{tab:h5_transfer_ranges}
\end{table}

\subsection{Overall Interpretation}
\label{app:h5_conclusion}

\paragraph{Directional geometry generalizes, but its dependence on training data is structured.}
The experiments establish several complementary properties of DiRGeo.

First, directional preferences generalize to unseen source images under the same visual process. Their reproducibility across CLIP, DINOv2, and supervised ViT further indicates that this phenomenon is not confined to a particular encoder pretraining objective.

Second, the paired category-exclusion experiments reveal stronger category-associated sensitivity in deep representations, particularly for \textsc{Insert}, \textsc{Grow}, and \textsc{Enter}. Their mean effects of $5.0$--$14.0$ percentage points exceed the matched-control range of $-0.6$--$4.2$ points, with positive effects consistently observed across five category partitions.

The predefined sign criterion alone does not separate object-level processes from controls, so the stronger magnitude distinction is reported as descriptive evidence. The results support category-associated dependence, not an exclusive semantic interpretation of the learned fields.

Third, transfer across process families and trajectory stages is limited and process-dependent. Some object-level processes share directional information, while \textsc{Colour} and several global appearance transformations exhibit substantial losses when their process-specific training data are excluded. Early-stage fields also vary in their ability to predict later-stage motion.

Together, these findings refine the interpretation of learned directional geometry. DiRGeo does not merely memorize individual image trajectories, but neither does it recover a universal direction field that transfers unchanged across arbitrary processes, categories, or representation spaces. Instead, its directional preferences reflect the interaction between the pretrained representation, the visual process, and the distribution of trajectory states used for learning.



\section{Additional Geometric Validation: Same Endpoints, Different Journeys}
\label{app:same_endpoint_journeys}
\label{app:same_endpoints}

This section examines whether the directional geometry learned by DiRGeo distinguishes different journeys between identical visual endpoints. This experiment provides a direct test of the path-dependent nature of the proposed ruler. If two visual trajectories begin and end at the same images, their endpoint displacement is identical. Nevertheless, the trajectories may follow different intermediate states, and a state-dependent directional field can assign different accumulated preferences to those journeys.

We construct two types of controlled object trajectories: progressive insertion, in which an object appears at a fixed location, and entry, in which the object moves into the image. Both start from the same object-removed background and end at the same image containing the complete object. Their intermediate images differ because the object becomes visible through different visual processes.

The central question is whether the separately learned \textsc{Insert} and \textsc{Enter} fields distinguish these journeys in a process-consistent manner. Specifically, does the \textsc{Insert} field favor the insertion journey, while the \textsc{Enter} field favors the entry journey, despite their identical endpoints?

We first define a directional path-preference metric and evaluate the original trajectory constructions on an independent confirmation set. 
We then test previously unseen journey constructions, compare the learned field with a standalone fitted sink, and examine whether the preference persists in the sink-orthogonal residual. 
Finally, we assess numerical robustness and distinguish directional alignment from the complete Randers path cost.

\subsection{Directional Path Preferences Between Identical Endpoints}
\label{app:h6_metric}

\paragraph{Directional contribution of a path.}
For a representation trajectory $\gamma:[0,1]\rightarrow\mathcal{M}$ and a process-trained directional field $\beta_p$, we define the accumulated wind alignment as
\begin{equation}
    W_p(\gamma)
    =
    \int_{\gamma}\beta_p(z)^\top dz.
    \label{eq:h6_directional_integral}
\end{equation}
This integral measures the accumulated alignment between the local displacement and the direction preferred by field $p$. Positive contributions indicate motion aligned with the field, while negative contributions indicate motion against it.

For the Euclidean-base Randers ruler introduced in Section~\ref{sec:characterization}, the complete path length decomposes as
\begin{equation}
    L_{F_p}(\gamma)
    =
    L_0(\gamma)-W_p(\gamma),
    \label{eq:h6_cost_decomposition}
\end{equation}
where $L_0(\gamma)$ denotes the symmetric Euclidean contribution accumulated along the path.

The two terms measure different properties. $L_0$ measures the total extent of motion through representation space without distinguishing forward and reverse directions. In contrast, $W_p$ measures how the path follows the learned directional field. A larger $W_p$ provides a greater directional reduction in the Randers cost, but does not necessarily produce a smaller complete path length when the symmetric lengths of the compared journeys differ.

The present experiment therefore focuses on $W_p$, which directly measures the process-dependent directional contribution. We examine the implications for the full Randers cost separately in Appendix~\ref{app:h6_full_cost}.

\paragraph{Discrete evaluation on visual trajectories.}
Each image trajectory contains sampled states $\gamma=(z_0,\ldots,z_T)$ extracted using a frozen visual encoder. The directional integral is evaluated using the corresponding finite trajectory displacements,
\begin{equation}
    \widehat{W}_p(\gamma)
    =
    \sum_{t=0}^{T-1}
    {\tfrac{1}{2}\bigl(\beta_p(z_t)+\beta_p(z_{t+1})\bigr)^\top}(z_{t+1}-z_t).
    \label{eq:h6_discrete_integral}
\end{equation}
This is the trapezoid rule: the field is averaged over the two states of each step.

This approximation is sensitive to the intermediate states of the trajectory rather than only its endpoints. Its numerical value can change under temporal refinement because the learned field generally varies across representation states. We therefore evaluate denser path sampling as a robustness check.

The primary reported calculations use the saved directional fields without an additional sphere-tangent projection at each state. We subsequently repeat the unseen-path experiment using the tangent-projected fields to verify that the result is not attributable to radial components introduced by feature normalization.

\paragraph{Reciprocal process preference.}
Let $\gamma_I$ denote an insert-type journey and $\gamma_E$ an enter-type journey. Both begin at the same object-removed image and end at the same fully composed image:
\begin{equation}
    \gamma_I(0)=\gamma_E(0),
    \qquad
    \gamma_I(1)=\gamma_E(1).
    \label{eq:h6_identical_endpoints}
\end{equation}
Although their endpoints coincide, their intermediate states and local displacements can differ.

We evaluate both journeys under both process-trained fields. For the \textsc{Insert} field, we define
\begin{equation}
    \ell_I
    =
    W_{\mathrm{Insert}}(\gamma_I)
    -
    W_{\mathrm{Insert}}(\gamma_E).
    \label{eq:h6_insert_margin}
\end{equation}
A positive $\ell_I$ indicates that the \textsc{Insert} field provides greater accumulated alignment along the insertion journey than along the entry journey.

For the \textsc{Enter} field, we reverse the subtraction order:
\begin{equation}
    \ell_E
    =
    W_{\mathrm{Enter}}(\gamma_E)
    -
    W_{\mathrm{Enter}}(\gamma_I).
    \label{eq:h6_enter_margin}
\end{equation}
A positive $\ell_E$ indicates that the \textsc{Enter} field favors the entry journey.

We define reciprocal preference as the simultaneous satisfaction of both conditions:
\begin{equation}
    \mathcal{I}_{\mathrm{recip}}
    =
    \mathbb{I}
    \left[
        \ell_I>0
        \;\wedge\;
        \ell_E>0
    \right].
    \label{eq:h6_reciprocal_indicator}
\end{equation}
Thus, reciprocal preference requires each process-trained field to favor its corresponding journey over the alternative.

This is more demanding than showing that a field assigns different scores to two paths. A field could distinguish the journeys while consistently preferring the same journey regardless of which process it was trained to represent. Reciprocal preference instead tests whether the two fields exhibit complementary process-specific preferences.

\paragraph{Why matching endpoints is a useful control.}
An endpoint-only directional measurement cannot distinguish two paths with identical endpoints. In particular, for a constant ambient field $\beta_0$,
\begin{equation}
    W_0(\gamma)
    =
    \beta_0^\top(z_T-z_0).
    \label{eq:h6_constant_field}
\end{equation}
This expression depends only on the endpoint displacement, so its value is identical for $\gamma_I$ and $\gamma_E$ when their endpoints coincide.

More generally, the continuous line integral of a conservative field depends only on the endpoints. Therefore, path-dependent differences in the directional integral require a contribution that is not explained by an exact endpoint potential under the evaluated geometry.

For numerical evaluation, however, finite-step integration can produce small nonzero differences even for fields whose continuous integrals are path-independent. We do not assume that fitted sink contributions vanish exactly. Instead, we evaluate the standalone fitted sink, learned field, and sink-orthogonal residual using the same discrete trajectory calculations.

\paragraph{Relationship to the sink--residual decomposition.}
As defined in Appendix~\ref{app:local_directionality}, the learned tangent field can be decomposed into a component aligned with a fitted spherical sink and an orthogonal residual. The corresponding directional path contributions can be evaluated separately.

If reciprocal preferences arise primarily from convergence toward a common attractor, the sink-aligned component should explain much of the observed path preference. Conversely, if the residual retains the learned field's reciprocal behavior, the effect cannot be attributed solely to that chosen sink axis.

The experiment therefore compares four fields on identical image paths: the learned field (unprojected), the sink-orthogonal residual $\delta\beta$ (computed from the tangent-projected field), the standalone fitted spherical sink (the separate baseline of Appendix~\ref{app:local_directionality}, with magnitude $\rho$ at every state), and a constant field. Because the learned and residual fields use different projections, and the fitted sink is a separate field rather than the projected component $\beta_{\mathrm{sink}}(z)$, the reported rows are not additive.

\paragraph{Aggregation and confidence intervals.}
For every source image, we evaluate the path-preference margins separately for each encoder representation and training seed. We report mean margins pooled over 10 representations and three seeds, as well as representation-specific measurements.

The reciprocal rate is computed from individual paired comparisons before aggregation. Thus, it measures the fraction of image--representation--seed evaluations for which both margins are positive, rather than the fraction of images whose representation-averaged margins are positive.

For the unseen-path experiment, the primary reciprocal rate additionally pools the 12 predefined pairs of new journey constructions. All comparisons associated with the same source image remain grouped during uncertainty estimation.

We compute 95\% confidence intervals using source-image bootstrap resampling over the 200 confirmation images. This preserves dependence among representations, training seeds, and trajectory constructions derived from the same image.

We also evaluate a convention that assigns half credit to an individual comparison whose margin is tied at zero. Under this convention, a constant field with two zero margins can obtain a reciprocal score of $25\%$, since each tied preference contributes $0.5$. This value is a consequence of the scoring convention, not a demonstrated directional preference or a meaningful chance-level baseline.

For the main conclusions, we use strict reciprocal preference, requiring both margins to be positive. The learned, residual, and sink results agree between the strict and half-credit conventions for the original paths, while the constant-field strict reciprocal rate is zero.

\subsection{Independent Confirmation on New Source Images}
\label{app:h6_confirmation}

\paragraph{Exploratory observation and predefined confirmation criteria.}
The original experiment was conducted on the 200 test images from the object-level dataset. It produced a reciprocal preference rate of $56.5\%$ with a 95\% confidence interval of $[53.7,59.5]$ for the learned fields, compared with $3.5\%$ for the standalone fitted sink.

Because this result was initially exploratory, we fixed the confirmation criteria before evaluating a separate image cohort.

The first criterion requires the learned field's reciprocal preference rate to exceed that of the standalone fitted sink. The second requires both pooled mean margins, $\ell_I$ and $\ell_E$, to be positive. The third requires both mean margins to be positive in at least eight of the 10 encoder representations.

These criteria jointly test whether the learned fields exhibit complementary process preferences, whether the result exceeds the simple attractor explanation, and whether it is reproducible across representation choices.

\paragraph{Confirmation dataset and trajectory construction.}
The confirmation cohort contains 200 new COCO train2017 source images, excluding all 1,000 images from the original object-level dataset and all 10,000 source images from the global appearance dataset.

We generate the new image sequences using the original object-trajectory builder with seed 2027 instead of the original seed 42. No directional field is retrained or adapted to the confirmation images.

Each source image produces two 10-frame trajectories. The insertion journey progressively reveals an object at its final location, while the entry journey moves the object into the scene. Both trajectories use exactly the same initial background and final object-composited image.

The median object area is $7.1\%$ of the image in the confirmation cohort, compared with $7.4\%$ in the original test cohort. The mean pixel change per step is $0.61$ versus $0.63$ for insertion and $3.19$ versus $3.11$ for entry, respectively. These measurements indicate that the confirmation data preserve broadly comparable object sizes and transformation magnitudes.

\paragraph{Data-generation and encoder checks.}
Several checks verify that the confirmation experiment evaluates the intended visual changes.

First, the dataset builder reproduces 40 previously saved test trajectories pixel-identically, with maximum difference of zero gray levels. Second, every newly generated insertion--entry pair has identical first and final frames.

Third, the frozen encoders reproduce saved representation features with cosine similarity at least $0.99999$. These checks reduce the possibility that the observed path differences arise from inconsistent image generation, endpoint mismatch, or feature-extraction changes between experimental runs.

\paragraph{Reciprocal preferences are confirmed on new images.}
Table~\ref{tab:h6_margins} reports the confirmation results.

For the original fade-in and slide-in constructions, the learned \textsc{Insert} margin is
$\ell_I=0.01177$ with 95\% confidence interval $[0.00903,0.01470]$.
The learned \textsc{Enter} margin is
$\ell_E=0.00362$ with interval $[0.00277,0.00451]$.
Both means are positive with confidence intervals above zero.

The learned fields exhibit reciprocal preferences in $56.6\%$ of individual evaluations, with interval $[54.0,59.1]$. 
The standalone fitted sink yields only $2.2\%$ $[1.6,2.8]$.

The paired reciprocal-rate improvement is $54.4$ percentage points with confidence interval $[51.6,57.0]$. Both mean margins are also positive in all 10 representations.

Consequently, all three predefined confirmation criteria are satisfied.

\paragraph{The residual preserves the preference.}
The sink-orthogonal residual yields mean margins $\ell_I=0.01064$ and $\ell_E=0.00336$, with both confidence intervals above zero. Its reciprocal preference rate is $56.0\%$ $[53.2,58.6]$, nearly identical to the complete learned field's $56.6\%$.

The standalone fitted sink gives much smaller margins: $\ell_I=0.00019$ and $\ell_E=-0.00012$. The opposite signs indicate that the fitted sink does not provide the complementary preferences required for reciprocal recognition.

These observations identify the primary source of the effect. The learned field's ability to distinguish the two journeys is largely retained after removing its fitted sink-aligned component. The result therefore reflects directional structure beyond the chosen common-attractor axis, rather than merely convergence toward the shared final representation.

\begin{table*}[!htbp]
    \centering
    \small
    \setlength{\tabcolsep}{6pt}
    \begin{tabular}{llrrr}
        \hline
        Paths & Field
        & Mean $\ell_I$ [95\% CI]
        & Mean $\ell_E$ [95\% CI]
        & Reciprocal (\%) [95\% CI] \\
        \hline
        Original & Learned
        & $.01177\ [.00903,.01470]$
        & $.00362\ [.00277,.00451]$
        & $56.6\ [54.0,59.1]$ \\
        & Residual
        & $.01064\ [.00824,.01325]$
        & $.00336\ [.00254,.00424]$
        & $56.0\ [53.2,58.6]$ \\
        & Fitted sink
        & $.00019\ [.00012,.00026]$
        & $-.00012\ [-.00019,-.00005]$
        & $2.2\ [1.6,2.8]$ \\
        \hline
        Unseen & Learned
        & $.00658\ [.00506,.00823]$
        & $.00310\ [.00236,.00390]$
        & $38.7\ [36.7,40.8]$ \\
        & Residual
        & $.00592\ [.00456,.00736]$
        & $.00281\ [.00211,.00357]$
        & $38.7\ [36.7,40.7]$ \\
        & Fitted sink
        & $.00011\ [.00007,.00016]$
        & $-.00011\ [-.00015,-.00007]$
        & $2.8\ [2.5,3.1]$ \\
        \hline
    \end{tabular}
    \caption{\textbf{Directional journey preferences on 200 independent confirmation images.} Original paths compare the fade-in and slide-in constructions; unseen paths pool 12 new insert-type/enter-type construction pairs. $\ell_I$ and $\ell_E$ measure the directional contribution favoring each field's corresponding journey. Reciprocal preference requires both margins to be positive in an individual comparison. Confidence intervals resample source images,    retaining their representations, seeds, and journey pairs. Learned is the unprojected field, Residual is the tangent field after removing its sink-axis component, and Fitted sink is the standalone sink baseline; the rows are therefore not additive.} 
    \label{tab:h6_margins}
\end{table*}

\paragraph{The effect varies across representation spaces.}
Table~\ref{tab:h6_by_representation} reports the learned margins and reciprocal rates separately for each representation.

For the original paths, reciprocal rates range from $34.7\%$ in supervised ViT L12 to $81.7\%$ in CLIP L4. Nevertheless, both mean margins are positive in every representation.

A positive average margin does not imply reciprocal preference for most individual images. The average can remain positive when some images exhibit large process-consistent preferences while others exhibit weak or inconsistent preferences. Therefore, the confirmation criteria do not require every representation to exceed a $50\%$ reciprocal rate.

The representation-level results support the presence of complementary preferences across all evaluated feature spaces, while also showing substantial variation in how frequently they occur.

\begin{table*}[!htbp]
    \centering
    \small
    \setlength{\tabcolsep}{4pt}
    \begin{tabular}{lrrrrrrrr}
        \hline
        & \multicolumn{4}{c}{Original paths}
        & \multicolumn{4}{c}{Unseen paths} \\
        Representation
        & $\ell_I$ & $\ell_E$ & Learned (\%) & Fitted sink (\%)
        & $\ell_I$ & $\ell_E$ & Learned (\%) & Fitted sink (\%) \\
        \hline
        CLIP L4
        & .00706 & .00206 & 81.7 & 0.0
        & .00220 & .00188 & 47.3 & 1.0 \\
        CLIP L8
        & .01792 & .00977 & 73.8 & 2.2
        & .00815 & .01049 & 58.7 & 1.4 \\
        CLIP L12
        & .02065 & .00580 & 42.2 & 0.5
        & .01777 & .00427 & 35.5 & 3.2 \\
        CLIP embedding
        & .01647 & .00489 & 41.2 & 4.2
        & .01604 & .00342 & 34.4 & 4.2 \\
        \hline
        DINOv2 L4
        & .00168 & .00056 & 52.0 & 1.0
        & .00003 & .00041 & 25.7 & 0.8 \\
        DINOv2 L8
        & .00831 & .00297 & 73.0 & 1.2
        & .00245 & .00316 & 44.9 & 2.7 \\
        DINOv2 L12
        & .01643 & .00425 & 36.2 & 5.5
        & .00830 & .00248 & 30.2 & 5.4 \\
        \hline
        Supervised ViT L4
        & .00334 & .00076 & 73.3 & 0.3
        & .00091 & .00056 & 43.3 & 0.2 \\
        Supervised ViT L8
        & .01563 & .00289 & 57.8 & 2.0
        & .00526 & .00270 & 40.5 & 1.7 \\
        Supervised ViT L12
        & .01019 & .00219 & 34.7 & 5.2
        & .00464 & .00158 & 26.6 & 7.5 \\
        \hline
    \end{tabular}
    \caption{\textbf{Representation-specific directional preference margins and reciprocal rates.} Mean learned-field margins are reported alongside reciprocal preference rates for the learned field and the fitted sink. All 10 representations have positive mean margins under both the original and unseen journey constructions, although reciprocal rates vary substantially across representations.}
    \label{tab:h6_by_representation}
\end{table*}

\begin{figure*}[!htb]
    \centering
    \includegraphics[width=\textwidth]{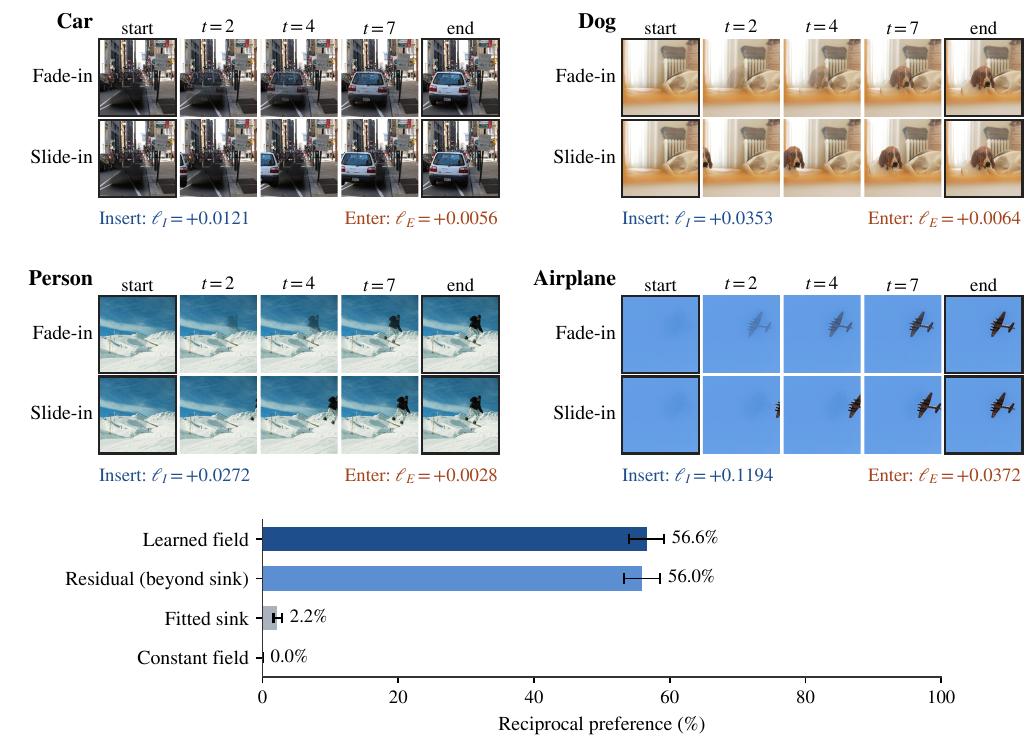}
    \caption{\textbf{Identical endpoints, different visual journeys.} Four confirmation images illustrate fade-in and slide-in trajectories with identical first and last frames but different intermediate states. Frames $t=0,2,4,7,9$ are shown for each journey. Annotated values are the directional margins of the \textsc{Insert} and \textsc{Enter} fields, pooled over 10 representations and three seeds. The summary panel reports strict reciprocal preference rates over all 200 confirmation images.}
    \label{fig:h6_journey_examples}
\end{figure*}

\paragraph{Qualitative examples.}
Figure~\ref{fig:h6_journey_examples} shows four confirmation images containing a car, dog, person, and airplane, with the original fade-in and slide-in journeys at $t=0,2,4,7,9$. The first and last frames are pixel-identical across the two rows, while the intermediate frames differ. Below each example are the mean margins of the \textsc{Insert} and \textsc{Enter} fields, averaged over the 10 representations and three training seeds (Table~\ref{tab:h6_figure_examples}).

\begin{table}[!htbp]
    \centering
    \small
    \setlength{\tabcolsep}{6pt}
    \begin{tabular}{lrr}
        \hline
        Example & $\ell_I$ & $\ell_E$ \\
        \hline
        Car & +.0121 & +.0056 \\
        Dog & +.0353 & +.0064 \\
        Person & +.0272 & +.0028 \\
        Airplane & +.1194 & +.0372 \\
        \hline
    \end{tabular}
    \caption{\textbf{Directional margins for the qualitative examples.} All values are pooled over 10 representations and three wind seeds. Positive values indicate that each field favors its corresponding journey. The selected examples illustrate the effect; the full 200-image confirmation provides the statistical evidence.}
    \label{tab:h6_figure_examples}
\end{table}

\subsection{Generalization to Previously Unseen Journey Constructions}
\label{app:h6_unseen_paths}

\paragraph{Why new journey constructions are necessary.}
The confirmation experiment establishes that learned fields distinguish the original fade-in and slide-in trajectories on new source images. However, both constructions are generated by the same procedures used during training.

The observed preferences could therefore depend on specific interpolation patterns rather than on the broader difference between object insertion and entry.

To examine this possibility, we construct new trajectories that preserve the same initial and final images while changing the intermediate journey. The previously trained fields are applied without retraining, so this experiment tests generalization to unseen trajectory constructions rather than adaptation to new supervision.

\paragraph{Four new insert-type journeys.}
We construct four alternative insertion trajectories.

The radial-reveal trajectory introduces the object progressively from its center outward, using a soft moving boundary. The vertical-wipe trajectory reveals the object from top to bottom. The pixel-dissolve trajectory reveals different object pixels at independently assigned times using a fixed random ordering for each image. The blur-to-sharp fade progressively increases opacity while sharpening an initially blurred object.

All four trajectories begin with the object-removed background and end with the complete object. They modify how the object becomes visible rather than introducing motion from outside the frame.

For these variants, construction checks verify that pixels outside the manipulated object region remain unchanged.

\paragraph{Three new enter-type journeys.}
We also construct three alternative object-entry trajectories.

The far-edge trajectory moves the object into the image from the edge farthest from its final location. The corner-entry trajectory moves the object diagonally from outside two image boundaries. The curved-entry trajectory follows a quadratic B\'ezier path that bends toward the image center before reaching the final object location.

These constructions change the spatial route by which the object enters the image while preserving its final appearance and position.

For all enter-type variants, visible object area is required to be nondecreasing along the trajectory. The construction checks confirm that this property holds and that no trajectory unintentionally reuses the original nearest-edge entry route in the far-edge condition.

\paragraph{Primary evaluation with 12 path pairs.}
The primary unseen-journey evaluation pairs every new insert-type path with every new enter-type path, producing $4\times3=12$ comparisons for each of the 200 confirmation images.

All pairs share the same initial background and final object-composited image. Thus, differences in directional contribution arise from their intermediate representation paths rather than their endpoints.

The evaluation criteria were fixed before generating or scoring the new paths. They match the three criteria used in the independent confirmation: a learned reciprocal rate above the sink's, positive pooled mean margins, and positive mean margins in at least eight of 10 representations.

The implementation verifies that the original paths are reproduced pixel-identically, every new construction preserves its endpoints, and the corresponding endpoint features agree to within $7.45\times10^{-7}$. Encoder reproductions also retain cosine similarity of at least $0.99999$ with the saved features.

\paragraph{The learned fields generalize to unseen journeys.}
The primary unseen-path results appear in Table~\ref{tab:h6_margins}.

Across the 12 new pairs, the learned fields achieve a reciprocal preference rate of $38.7\%$ with a 95\% confidence interval of $[36.7,40.8]$. The standalone fitted sink achieves only $2.8\%$ $[2.5,3.1]$.

The paired improvement is $35.9$ percentage points, with confidence interval $[33.9,38.0]$.

The learned mean margins are $\ell_I=0.00658$ and $\ell_E=0.00310$, both with confidence intervals above zero. Both mean margins remain positive in all 10 representations.

Therefore, all three predefined criteria are satisfied for the new journey constructions.

The reciprocal rate is lower than the $56.6\%$ obtained for the original paths, indicating that generalization is not complete. Nevertheless, the positive mean margins and substantial advantage over the sink demonstrate that the learned path preference persists beyond the original fade-in and slide-in interpolation procedures.

\paragraph{The residual continues to explain the effect.}
The sink-orthogonal residual yields mean margins $\ell_I=0.00592$ and $\ell_E=0.00281$ and the same reported reciprocal rate of $38.7\%$.

The corresponding standalone fitted sink has a very small positive Insert margin and a negative Enter margin, resulting in only $2.8\%$ reciprocal preference.
Under a common tangent projection, the decomposition is exactly additive. The tangent-projected learned margins are $\ell_I=0.00673$ and $\ell_E=0.00303$, so the learned field's own sink-aligned component contributes only $0.00081$ and $0.00022$, about 7 and 13 times less than the residual.

Thus, the unseen-journey result retains the same geometric interpretation as the original confirmation: the complementary process preference is largely preserved outside the fitted common-attractor direction.

\paragraph{Results for all 12 construction pairs.}
Table~\ref{tab:h6_new_pair_matrix} reports the reciprocal preference rate for each of the 12 new insert--enter construction pairs.

The learned rates range from $33.8\%$ to $47.1\%$, while sink rates range from $2.3\%$ to $3.5\%$.

Among the tested combinations, blur-to-sharp fade paired with corner entry produces the highest learned reciprocal rate, $47.1\%$. Radial reveal paired with far-edge entry yields $33.9\%$, while vertical wipe paired with far-edge entry yields $33.8\%$.

The substantial variation across pairs indicates that the particular path construction affects how well the learned field recognizes the process. Nevertheless, every tested pair retains a much larger reciprocal rate for the learned field than for the fitted sink.
\begin{table}[!htbp]
    \centering
    \small
    \setlength{\tabcolsep}{3pt}
    \begin{tabular}{lrrr}
        \hline
        Insert-type path
        & \shortstack{Far-edge\\entry} & \shortstack{Corner\\entry} & \shortstack{Curved\\entry} \\
        \hline
        Radial reveal
        & 33.9 / 3.1 & 38.6 / 2.7 & 36.2 / 2.9 \\
        Vertical wipe
        & 33.8 / 2.6 & 39.3 / 2.8 & 37.5 / 2.3 \\
        Pixel dissolve
        & 35.1 / 2.8 & 39.5 / 2.8 & 36.2 / 2.3 \\
        Blur-to-sharp fade
        & 42.2 / 3.0 & 47.1 / 2.8 & 45.1 / 3.5 \\
        \hline
    \end{tabular}
    \caption{\textbf{Reciprocal preference rates for 12 unseen journey pairs (\%).} Each entry reports learned-field / fitted-sink reciprocal rates, pooled over source images, representations, and training seeds. All path pairs have identical endpoints. The learned field exceeds the fitted sink in every construction pair.}
    \label{tab:h6_new_pair_matrix}
\end{table}

\paragraph{Transfer depends on which journey is modified.}
We additionally evaluate each new path against the original path of the other process type, rather than requiring both journeys to be new.

Table~\ref{tab:h6_path_variants} reports these secondary comparisons.

When the original fade-in is retained and only the entry journey changes, learned reciprocal rates range from $50.8\%$ to $56.8\%$, close to the original $56.6\%$ result.

When the original slide-in is retained and only the insertion journey changes, the rates are lower, ranging from $35.6\%$ to $45.4\%$.

This asymmetry suggests that the learned \textsc{Insert} field is more sensitive to the specific uniform-opacity progression used during its training than the \textsc{Enter} field is to the original straight-line entry construction.

The new insertion variants nevertheless retain positive pooled mean margins. For pixel dissolve, both mean margins are positive in nine of the 10 representations, rather than all 10. The corresponding unresolved representation is DINOv2 L4, where the mean Insert margin is $-0.00073$.

Thus, the new paths satisfy the predefined eight-representation requirement without implying uniformly positive margins in every representation for every variant.

\begin{table*}[!htbp]
    \centering
    \small
    \setlength{\tabcolsep}{6pt}
    \begin{tabular}{llrrrr}
        \hline
        New journey & Compared with
        & Learned (\%) [95\% CI] & Fitted sink (\%)
        & Mean $\ell_I$ & Mean $\ell_E$ \\
        \hline
        Radial reveal & Original slide-in
        & $35.6\ [33.2,38.0]$ & 3.2 & .00344 & .00208 \\
        Vertical wipe & Original slide-in
        & $37.6\ [35.3,40.0]$ & 2.5 & .00278 & .00234 \\
        Pixel dissolve & Original slide-in
        & $36.5\ [34.2,38.8]$ & 3.1 & .01065 & .00325 \\
        Blur-to-sharp fade & Original slide-in
        & $45.4\ [42.8,48.0]$ & 3.5 & .00973 & .00343 \\
        \hline
        Far-edge entry & Original fade-in
        & $50.8\ [48.0,53.5]$ & 2.5 & .01138 & .00374 \\
        Corner entry & Original fade-in
        & $56.8\ [54.0,59.5]$ & 2.5 & .01209 & .00436 \\
        Curved entry & Original fade-in
        & $55.6\ [53.0,58.2]$ & 2.2 & .01163 & .00372 \\
        \hline
        Fade-in with $t^2$ timing & Original slide-in
        & $56.4\ [53.7,59.0]$ & 2.6 & .01160 & .00366 \\
        Slide-in with $t^2$ timing & Original fade-in
        & $57.3\ [54.7,59.8]$ & 1.5 & .01171 & .00384 \\
        \hline
    \end{tabular}
    \caption{\textbf{Secondary journey-construction and timing comparisons.} Reciprocal preference rates for a new journey paired with the original path of the other process type. The first seven rows change the intermediate journey construction. The final two modify only the timing of the original geometric path. Confidence intervals resample the 200 source images.}
    \label{tab:h6_path_variants}
\end{table*}

\subsection{Process Specificity and Numerical Robustness}
\label{app:h6_robustness}

\paragraph{The preference follows the corresponding trained process.}
The reciprocal experiment uses \textsc{Insert} and \textsc{Enter} fields, which are trained on the two relevant visual processes. To determine whether the path difference is simply preferred by arbitrary directional fields, we also evaluate the same path pairs using the \textsc{Grow} and \textsc{Colour} fields.

For this comparison, we adopt a common subtraction order for every field:
\begin{equation}
    \Delta W_p
    =
    W_p(\gamma_I)-W_p(\gamma_E).
    \label{eq:h6_common_contrast}
\end{equation}
A positive value favors the insertion journey, whereas a negative value favors the entry journey.

Under this convention, the \textsc{Insert} field has a positive mean and the \textsc{Enter} field a negative mean, both with confidence intervals excluding zero. By contrast, the \textsc{Grow} and \textsc{Colour} field differences are near zero, with confidence intervals containing zero.

Table~\ref{tab:h6_specificity} reports the results for both original and unseen paths.

This comparison provides additional evidence that the path preference is process-dependent. It is not a uniform property shared by all object-level directional fields.

\begin{table*}[!htbp]
    \centering
    \small
    \setlength{\tabcolsep}{10pt}
    \begin{tabular}{lrr}
        \hline
        Field
        & Original paths: mean difference [95\% CI]
        & Unseen paths: mean difference [95\% CI] \\
        \hline
        Insert
        & $.01177\ [.00903,.01470]$
        & $.00658\ [.00506,.00823]$ \\
        Enter
        & $-.00362\ [-.00451,-.00277]$
        & $-.00310\ [-.00390,-.00236]$ \\
        Grow
        & $-.00041\ [-.00085,.00002]$
        & $-.00027\ [-.00073,.00012]$ \\
        Colour
        & $-.00032\ [-.00076,.00012]$
        & $-.00019\ [-.00056,.00021]$ \\
        \hline
    \end{tabular}
    \caption{\textbf{Specificity of the same-endpoint directional contrast.} Mean $W_p(\gamma_I)-W_p(\gamma_E)$ under four process-trained fields. Positive values favor insertion, and negative values favor entry. Only the fields trained for the corresponding two processes show confidence intervals excluding zero under both the original and unseen constructions. The Enter row uses the common subtraction order and is therefore the negative of $\ell_E$.}
    \label{tab:h6_specificity}
\end{table*}

\paragraph{Temporal sampling refinement.}
The directional contribution is evaluated as a finite sum, so we test whether the result depends strongly on the number of sampled trajectory states.

We refine the original 10-frame paths to 19 frames while preserving the original states at every second sampling position. The denser trajectories therefore follow the same intended image transitions, but use more intermediate steps to approximate the directional integral.

Table~\ref{tab:h6_refinement} reports the results.

The learned reciprocal rate changes only from $56.6\%$ to $57.5\%$, while the residual changes from $56.0\%$ to $57.2\%$. Both mean margins remain positive in all 10 representations.

The average absolute directional margin changes modestly under refinement, the ratio is $1.05$ between the original and refined calculations. Margin signs agree in $91.6\%$ of the evaluated comparisons.

These results indicate that the observed preference is reasonably stable under this tested refinement. They do not establish exact discretization invariance or convergence of the numerical integral in the limit of arbitrarily dense sampling.

\begin{table*}[!htbp]
    \centering
    \small
    \setlength{\tabcolsep}{7pt}
    \begin{tabular}{lrrr}
        \hline
        Field
        & Mean $\ell_I$ [95\% CI]
        & Mean $\ell_E$ [95\% CI]
        & Reciprocal (\%) [95\% CI] \\
        \hline
        Learned
        & $.01244\ [.00956,.01558]$
        & $.00386\ [.00296,.00479]$
        & $57.5\ [54.9,60.1]$ \\
        Residual
        & $.01131\ [.00871,.01406]$
        & $.00354\ [.00266,.00447]$
        & $57.2\ [54.5,59.9]$ \\
        Fitted sink
        & $.00022\ [.00017,.00028]$
        & $-.00019\ [-.00024,-.00014]$
        & $1.8\ [1.3,2.3]$ \\
        \hline
    \end{tabular}
    \caption{\textbf{Temporal refinement of original same-endpoint journeys.} Results using 19-frame trajectories that preserve the original 10-frame states. Learned and residual reciprocal preference rates remain close to the original values, while fitted-sink preferences remain small. Confidence intervals resample the 200 confirmation images.}
    \label{tab:h6_refinement}
\end{table*}

\paragraph{Changes in traversal timing.}
We also modify the timing of the original journeys without changing their intended spatial construction.

For insertion, the opacity progression is changed from linear timing to a quadratic schedule $t^2$. For entry, the fraction of completed object motion is similarly changed to $t^2$ along the original straight-line route.

The resulting reciprocal rates are $56.4\%$ for the retimed fade-in and $57.3\%$ for the retimed slide-in, compared with $56.6\%$ for the original pair.

These results suggest that the observed preferences are not critically dependent on the exact linear timing of the original transformations. However, the test covers only the specified quadratic reparameterizations, not arbitrary temporal warps.

\paragraph{Sphere-tangent field projection.}
The primary calculations use the directional-integral implementation without applying an additional field projection at every state. Because all encoder features are unit-normalized, we repeat the primary unseen-path experiment using the sphere-tangent field projection
\(
\beta_p^{\mathrm{tan}}(z)
=
(I-zz^\top)\beta_p(z).
\)

The projected calculation yields mean margins $\ell_I=0.00673$ and $\ell_E=0.00303$. Its reciprocal preference rate is $38.6\%$ with a 95\% confidence interval of $[36.6,40.7]$, compared with $38.7\%$ $[36.7,40.8]$ in the original calculation. The standalone fitted sink remains at $2.8\%$.

Thus, removing the radial field component has negligible effect on the central unseen-journey conclusion. This supports the interpretation that the observed reciprocal preferences are associated with directional differences along the tested paths rather than primarily with normalization-induced radial contributions.

\paragraph{Constant-field and fitted-sink references.}
The constant-field baseline provides an exact conceptual reference for endpoint-only directional information. Its continuous directional integral depends only on endpoint displacement, which is identical for the compared journeys.

In the original discrete confirmation, the constant-field margins are effectively zero to five decimal places. Under strict reciprocal scoring, its rate is $0.0\%$. Under the alternative half-credit convention, ties produce an apparent score of $25.0\%$ for original paths and $24.6\%$ for unseen paths.

These half-credit values should not be interpreted as evidence that a constant field distinguishes the journeys.

The standalone fitted sink produces small but nonzero discrete margins. Its reciprocal rates of $2.2\%$ and $2.8\%$ for the original and unseen paths indicate that the discrete fitted-sink calculation is not treated as an exact endpoint-only null.

The learned and residual fields, however, produce substantially larger mean margins and reciprocal rates under the same evaluation procedure. The relevant empirical result is therefore the paired advantage over these simpler references, not an assumption that every sink integral must vanish exactly.

\subsection{Directional Contribution Versus Complete Randers Path Cost}
\label{app:h6_full_cost}

\paragraph{Identical endpoints do not imply identical symmetric path lengths.}
The preceding results establish process-consistent preferences in the directional contribution $W_p(\gamma)$. However, a larger wind alignment does not necessarily imply a smaller complete Randers length.

From Eq.~\ref{eq:h6_cost_decomposition}, comparing two journeys under the same field gives
\begin{equation}
\begin{aligned}
    L_{F_p}(\gamma_I)-L_{F_p}(\gamma_E)
    &=
    L_0(\gamma_I)-L_0(\gamma_E)\\
    &\quad-
    \left[
        W_p(\gamma_I)-W_p(\gamma_E)
    \right].
\end{aligned}
\label{eq:h6_full_cost_difference}
\end{equation}
Even when the endpoints are identical, the symmetric lengths $L_0(\gamma_I)$ and $L_0(\gamma_E)$ can differ because the trajectories pass through different intermediate representation states.

For example, a fade-in journey may follow a relatively short route through feature space, while a slide-in journey visits a longer sequence of intermediate states. The directional contribution can favor the slide-in under the \textsc{Enter} field, but its greater symmetric length may still make its full Randers cost larger.

\paragraph{The shorter journey often dominates the complete cost.}
In the confirmation experiment, the original fade-in journey has lower symmetric path length than the slide-in journey in $99.5\%$ of source images.

For the new journey constructions, insert-type paths are shorter than enter-type paths in $98.5\%$ of the source images.

Consequently, the symmetric component strongly favors insertion-type journeys. Under the complete normalized path-cost comparison, the \textsc{Enter} field prefers the slide-in journey only $16.4\%$ of the time, compared with $7.7\%$ for the sink and $5.1\%$ for the constant-field baseline.

Thus, the directional contribution and the complete path cost answer different questions. The former characterizes how a journey aligns with a learned process direction; the latter combines this alignment with the total amount of movement through representation space.

\paragraph{What the experiment establishes.}
The experiment demonstrates that learned directional fields retain information about the intermediate journey even when the initial and final images are fixed.

This information is process-dependent: the \textsc{Insert} field favors insertion-type journeys, while the \textsc{Enter} field favors entry-type journeys. The effect is confirmed on new source images, persists for unseen journey constructions, survives removal of the fitted sink component, and remains stable under the tested temporal and sphere-tangent corrections.

However, the results do not establish that the complete Randers cost always selects the journey corresponding to the trained process. Nor do they establish a general method for choosing better image-generation or video-generation paths.

The confirmed contribution is narrower and more precise: the learned directional component of the ruler captures process-dependent properties of visual journeys that are not determined by their endpoints alone.

\subsection{Overall Interpretation}
\label{app:h6_conclusion}

The same-endpoint experiments provide an additional geometric validation of DiRGeo beyond local directional discrimination and whole-path forward--reverse asymmetry.

The existence evidence established that observed trajectories exhibit learnable directional preferences. The sink--residual analysis showed that these preferences are not fully explained by a selected common attractor. The process-specificity experiments demonstrated that different visual processes induce different local fields.

The present experiment adds a complementary result: two trajectories may share exactly the same starting and ending images, yet different process-trained fields assign them distinct and reciprocal accumulated directional contributions.

The independent confirmation shows a $56.6\%$ reciprocal preference rate for the original journeys, compared with $2.2\%$ for the standalone fitted sink. On 12 new journey pairs, the learned rate remains $38.7\%$, compared with $2.8\%$ for the fitted sink. The corresponding residual fields retain nearly identical preferences.

These results support the view that the geometry of visual representations cannot be characterized solely by endpoints or a common destination. The intermediate path matters, and learned directional fields can encode how the representation moves through that path.

At the same time, the distinction between directional alignment and full Randers cost remains essential. The experiment establishes path-dependent directional information, not universal superiority in complete trajectory-length comparisons.



\section{Scope of the Proposed DiRGeo Geometry and Limitations}
\label{app:scope_limitations}

\paragraph{Scientific scope.}
DiRGeo investigates whether ordered visual processes induce learnable, orientation-dependent geometry in frozen visual representations. Our objective is to characterize how representations move along visual trajectories, rather than to reconstruct the complete representation manifold or learn a new visual encoder. We focus on the asymmetric component of trajectory geometry, modeled through a bounded directional field under a fixed Euclidean base metric. This formulation isolates directional preferences from symmetric displacement costs, allowing their existence, structure, and functional consequences to be examined independently.

\paragraph{Ordered visual processes.}
The proposed geometry is learned from ordered image trajectories in which the transformation specifies a meaningful progression. Our experiments include global appearance transformations, object-level processes, and matched controls, with additional evaluation on natural-video events. These processes provide directional supervision without requiring semantic labels for individual trajectory steps. However, an arbitrary pair of images does not necessarily define a unique or meaningful direction. The learned field should therefore be interpreted relative to the process distribution used for training, rather than as a universal orientation assigned to all possible image changes.

\paragraph{Geometric assumptions and interpretation.}
We adopt a Randers-type ruler with a fixed Euclidean symmetric component and a learned, norm-bounded directional field. The sphere-tangent correction removes radial components associated with normalized representations, but does not recover the tangent space of an unknown representation manifold. Similarly, the sink--residual decomposition tests whether directional information remains beyond a specified common-attractor axis; it does not exclude every possible global geometric explanation. Our findings establish that the learned ruler captures reproducible directional structure under the tested conditions, rather than proving the existence or uniqueness of an intrinsic Finsler geometry in pretrained representations. The fixed symmetric component is a deliberate modeling choice that enables the directional contribution to be studied without introducing additional metric-learning assumptions.

\paragraph{Empirical generalization.}
Directional preferences are observed across 17 visual processes, 10 representations from three encoder families, and held-out source images. Nevertheless, separate fields are trained for each process--representation combination. The experiments therefore demonstrate reproducibility across different representations and visual transformations, rather than transfer of a single universal field between encoders or processes. Additional category-exclusion experiments reveal stronger category-associated dependence in deep object-level representations, while transfer across unseen processes and trajectory stages varies substantially. Natural-video experiments provide evidence of event-conditioned directional preferences, but their limited number of recordings and annotated events does not support a universal temporal arrow. Together, these findings characterize both the generality and the process dependence of the observed geometry.

\paragraph{Functional validation.}
We examine the functional consequences of directional geometry through two complementary settings. First, same-endpoint journey experiments demonstrate that process-trained fields can assign distinct directional contributions to trajectories connecting identical initial and final images. These preferences persist on unseen images and alternative journey constructions, supporting the importance of intermediate paths beyond endpoint displacement alone. However, differences in directional contribution do not necessarily imply the same ordering under the complete Randers cost, which also includes symmetric path length. Second, restoration-selection experiments test whether the learned fields can rank predefined correction candidates under a known degradation process, without accessing clean references during scoring. The learned geometry improves pooled colour-correction selection over a fitted-sink reference, although the benefit varies across representations and does not consistently extend to deblurring. These experiments establish complementary geometric and task-level utility within controlled settings, rather than a general method for trajectory generation or image restoration.

\paragraph{Extensions of the framework.}
The present formulation provides a starting point for investigating richer representation geometry. One direction is to jointly learn the symmetric metric and directional field, allowing both displacement sensitivity and orientation preference to vary with representation state. Another is to model interactions among visual processes through shared or process-conditioned fields, rather than independently trained networks. Extending directional geometry to continuous natural-video trajectories, multimodal representations, and trajectory planning may further reveal how learned representations organize evolving visual content. These extensions build on the current findings while introducing additional questions concerning geometric identifiability, process composition, and the relationship between learned trajectory preferences and downstream behavior.


%% file: main.bib
@article{fefferman2016testing,
  title = {Testing the manifold hypothesis},
  author = {Fefferman, Charles and Mitter, Sanjoy and Narayanan, Hariharan},
  journal = {Journal of the American Mathematical Society},
  volume = {29},
  number = {4},
  pages = {983--1049},
  year = {2016},
  doi = {10.1090/jams/852}
}

@article{bao2004zermelo,
  title = {{Zermelo} navigation on {Riemannian} manifolds},
  author = {Bao, David and Robles, Colleen and Shen, Zhongmin},
  journal = {Journal of Differential Geometry},
  volume = {66},
  number = {3},
  pages = {377--435},
  year = {2004},
  doi = {10.4310/jdg/1098137838}
}

@inproceedings{arvanitidis2018latent,
  title = {Latent space oddity: On the curvature of deep generative models},
  author = {Arvanitidis, Georgios and Hansen, Lars Kai and Hauberg, S{\o}ren},
  booktitle = {International Conference on Learning Representations},
  year = {2018},
  url = {https://openreview.net/forum?id=SJzRZ-WCZ}
}

@article{henaff2019straightening,
  title = {Perceptual straightening of natural videos},
  author = {H{\'e}naff, Olivier J. and Goris, Robbe L. T. and Simoncelli, Eero P.},
  journal = {Nature Neuroscience},
  volume = {22},
  number = {6},
  pages = {984--991},
  year = {2019},
  doi = {10.1038/s41593-019-0377-4}
}

@inproceedings{wang2022quasimetrics,
  title = {On the learning and learnability of quasimetrics},
  author = {Wang, Tongzhou and Isola, Phillip},
  booktitle = {International Conference on Learning Representations},
  year = {2022},
  url = {https://openreview.net/forum?id=y0VvIg25yk}
}

@inproceedings{wang2022interval,
  title = {Improved representation of asymmetrical distances with interval quasimetric embeddings},
  author = {Wang, Tongzhou and Isola, Phillip},
  booktitle = {NeurIPS Workshop on Symmetry and Geometry in Neural Representations},
  year = {2022},
  url = {https://openreview.net/forum?id=KRiST_rzkGl}
}

@inproceedings{dages2025finslermds,
  title = {{Finsler} multi-dimensional scaling: Manifold learning for asymmetric dimensionality reduction and embedding},
  author = {Dag{\`e}s, Thomas and Weber, Simon and Lin, Ya-Wei Eileen and Talmon, Ronen and Cremers, Daniel and Lindenbaum, Michael and Bruckstein, Alfred M. and Kimmel, Ron},
  booktitle = {Proceedings of the IEEE/CVF Conference on Computer Vision and Pattern Recognition},
  pages = {25842--25853},
  year = {2025}
}

@misc{debaussart2026directed,
  title = {A {Finslerian} approach for embedding directed data},
  author = {Debaussart-Joniec, Gwendal and Blanchard, Th{\'e}au and Kalogeratos, Argyris},
  howpublished = {arXiv preprint arXiv:2609.37649},
  year = {2026},
  eprint = {2609.37649},
  archivePrefix = {arXiv},
  url = {https://arxiv.org/abs/2609.37649}
}

@inproceedings{zweig2026lineage,
  title = {Learning lineage-guided geodesics with {Finsler} geometry},
  author = {Zweig, Aaron and Zhang, Mingxuan and Knowles, David A. and Azizi, Elham},
  booktitle = {Proceedings of the 42nd Conference on Uncertainty in Artificial Intelligence},
  series = {Proceedings of Machine Learning Research},
  volume = {337},
  pages = {8375--8386},
  year = {2026},
  publisher = {PMLR},
  url = {https://proceedings.mlr.press/v337/zweig26b.html}
}

@article{gahtan2026eikonal,
  title = {Differentiable {Randers-Finsler} eikonal solvers},
  author = {Gahtan, Barak and Shpund, Jacob and Bronstein, Alex M.},
  journal = {Computer Graphics Forum},
  pages = {e70489},
  year = {2026},
  doi = {10.1111/cgf.70489}
}

@inproceedings{misra2016shuffle,
  title = {Shuffle and learn: Unsupervised learning using temporal order verification},
  author = {Misra, Ishan and Zitnick, C. Lawrence and Hebert, Martial},
  booktitle = {Proceedings of the European Conference on Computer Vision},
  pages = {527--544},
  year = {2016},
  doi = {10.1007/978-3-319-46448-0_32}
}

@inproceedings{pickup2014seeing,
  title = {Seeing the arrow of time},
  author = {Pickup, Lyndsey C. and Pan, Zheng and Wei, Donglai and Shih, YiChang and Zhang, Changshui and Zisserman, Andrew and Sch{\"o}lkopf, Bernhard and Freeman, William T.},
  booktitle = {Proceedings of the IEEE Conference on Computer Vision and Pattern Recognition},
  pages = {2035--2042},
  year = {2014}
}

@inproceedings{wei2018arrow,
  title = {Learning and using the arrow of time},
  author = {Wei, Donglai and Lim, Joseph J. and Zisserman, Andrew and Freeman, William T.},
  booktitle = {Proceedings of the IEEE Conference on Computer Vision and Pattern Recognition},
  pages = {8052--8060},
  year = {2018}
}

@inproceedings{dasgupta2026inv,
  title = {Invariant features in language models: Geometric characterization and model attribution},
  author = {Dasgupta, Agnibh and Tanvir, Abdullah and Zhong, Xin},
  booktitle = {Advances in Neural Information Processing Systems},
  year = {2026},
  note = {To appear},
  url = {https://arxiv.org/abs/2605.06458}
}

@inproceedings{shao2018riemannian,
  title = {The {Riemannian} geometry of deep generative models},
  author = {Shao, Hang and Kumar, Abhishek and Fletcher, P. Thomas},
  booktitle = {Proceedings of the IEEE Conference on Computer Vision and Pattern Recognition Workshops},
  pages = {315--323},
  year = {2018},
  doi = {10.1109/CVPRW.2018.00071}
}

@inproceedings{ke2021musiq,
  title = {{MUSIQ}: Multi-scale image quality transformer},
  author = {Ke, Junjie and Wang, Qifei and Wang, Yilin and Milanfar, Peyman and Yang, Feng},
  booktitle = {Proceedings of the IEEE/CVF International Conference on Computer Vision},
  pages = {5148--5157},
  year = {2021}
}

@article{mittal2012brisque,
  title = {No-reference image quality assessment in the spatial domain},
  author = {Mittal, Anish and Moorthy, Anush Krishna and Bovik, Alan Conrad},
  journal = {IEEE Transactions on Image Processing},
  volume = {21},
  number = {12},
  pages = {4695--4708},
  year = {2012},
  doi = {10.1109/TIP.2012.2214050}
}

@article{mittal2013niqe,
  title = {Making a ``completely blind'' image quality analyzer},
  author = {Mittal, Anish and Soundararajan, Rajiv and Bovik, Alan Conrad},
  journal = {IEEE Signal Processing Letters},
  volume = {20},
  number = {3},
  pages = {209--212},
  year = {2013},
  doi = {10.1109/LSP.2012.2227726}
}

@inproceedings{radford2021clip,
  title = {Learning transferable visual models from natural language supervision},
  author = {Radford, Alec and Kim, Jong Wook and Hallacy, Chris and Ramesh, Aditya and Goh, Gabriel and Agarwal, Sandhini and Sastry, Girish and Askell, Amanda and Mishkin, Pamela and Clark, Jack and Krueger, Gretchen and Sutskever, Ilya},
  booktitle = {Proceedings of the 38th International Conference on Machine Learning},
  series = {Proceedings of Machine Learning Research},
  volume = {139},
  pages = {8748--8763},
  year = {2021},
  publisher = {PMLR},
  url = {https://proceedings.mlr.press/v139/radford21a.html}
}

@article{oquab2024dinov2,
  title = {{DINOv2}: Learning robust visual features without supervision},
  author = {Oquab, Maxime and Darcet, Timoth{\'e}e and Moutakanni, Th{\'e}o and Vo, Huy V. and Szafraniec, Marc and Khalidov, Vasil and Fernandez, Pierre and Haziza, Daniel and Massa, Francisco and El-Nouby, Alaaeldin and Assran, Mahmoud and Ballas, Nicolas and Galuba, Wojciech and Howes, Russell and Huang, Po-Yao and Li, Shang-Wen and Misra, Ishan and Rabbat, Michael and Sharma, Vasu and Synnaeve, Gabriel and Xu, Hu and J{\'e}gou, Herv{\'e} and Mairal, Julien and Labatut, Patrick and Joulin, Armand and Bojanowski, Piotr},
  journal = {Transactions on Machine Learning Research},
  year = {2024},
  url = {https://openreview.net/forum?id=a68SUt6zFt}
}

@inproceedings{dosovitskiy2021vit,
  title = {An image is worth 16x16 words: Transformers for image recognition at scale},
  author = {Dosovitskiy, Alexey and Beyer, Lucas and Kolesnikov, Alexander and Weissenborn, Dirk and Zhai, Xiaohua and Unterthiner, Thomas and Dehghani, Mostafa and Minderer, Matthias and Heigold, Georg and Gelly, Sylvain and Uszkoreit, Jakob and Houlsby, Neil},
  booktitle = {International Conference on Learning Representations},
  year = {2021},
  url = {https://openreview.net/forum?id=YicbFdNTTy}
}

@inproceedings{lin2014coco,
  title = {{Microsoft COCO}: Common objects in context},
  author = {Lin, Tsung-Yi and Maire, Michael and Belongie, Serge and Hays, James and Perona, Pietro and Ramanan, Deva and Doll{\'a}r, Piotr and Zitnick, C. Lawrence},
  booktitle = {Proceedings of the European Conference on Computer Vision},
  pages = {740--755},
  year = {2014},
  doi = {10.1007/978-3-319-10602-1_48}
}
